\documentclass[12pt]{article}

\usepackage{arxiv}            

\usepackage{xcolor}
\definecolor{darkblue}{rgb}{0, 0.12, 0.55}
\definecolor{darkred}{rgb}{0.6,0,0}
\definecolor{darkgreen}{rgb}{0,0.6,0}
\definecolor{Gray}{gray}{0.9}
\definecolor{mygreen}{rgb}{0,0.6,0}
\definecolor{mymauve}{rgb}{0.58,0,0.82}

\usepackage[
  breaklinks = true,
  colorlinks = true,
  linkcolor  = black,
  urlcolor   = darkblue,
  citecolor  = blue,
  bookmarks  = false
]{hyperref}

\usepackage[
  backend=biber,
  style=numeric,      
  sorting=none,
  maxbibnames=6,
  minbibnames=6,
  doi=false,
  isbn=false,
  url=false
]{biblatex}
\DeclareCiteCommand{\supercite}[\mkbibsuperscript]
  {} 
  {\printtext[bibhyperref]{\printfield{labelnumber}}} 
  {\addcomma} 
  {} 

\let\cite\supercite

\DeclareNameAlias{default}{family-given}
\DeclareNameAlias{sortname}{family-given}

\DeclareFieldFormat{citehyperref}{\bibhyperref{#1}}
\savebibmacro{cite}
\renewbibmacro*{cite}{%
  \printtext[citehyperref]{%
    \restorebibmacro{cite}%
    \usebibmacro{cite}%
  }%
}
\savebibmacro{textcite}
\renewbibmacro*{textcite}{%
  \printtext[citehyperref]{%
    \restorebibmacro{textcite}%
    \usebibmacro{textcite}%
  }%
}

\usepackage{multirow} 
\usepackage{amsmath,amsfonts,bm}

\def\eqref#1{equation~\ref{#1}}
\def\1{\bm{1}}

\DeclareMathAlphabet{\mathsfit}{\encodingdefault}{\sfdefault}{m}{sl}
\SetMathAlphabet{\mathsfit}{bold}{\encodingdefault}{\sfdefault}{bx}{n}

\usepackage{amsthm,amsmath,amssymb}

\usepackage{bm}
\usepackage{bbm}

\usepackage[ruled,vlined]{algorithm2e}

\usepackage{wrapfig}
\usepackage{subcaption}   % loads 'caption' internally
\usepackage{colortbl}
\usepackage{adjustbox}
\usepackage{makecell}
\usepackage{pifont}

\usepackage{caption}
\usepackage{listings}
\usepackage[normalem]{ulem}  % \sout without redefining \emph

\usepackage[normalem]{ulem} % put in preamble
\renewenvironment{abstract}
{
  \centerline{\large \bfseries \scshape Abstract}
  \vspace{0.5em}
  \noindent
}
{
  \par\vspace{1em}
}

\newlength{\mysize}

\allowdisplaybreaks

\theoremstyle{definition}

\usepackage{setspace}
\usepackage{lineno}
\title{Identifying Trustworthiness Challenges in Deep Learning Models for Continental-Scale Water Quality Prediction
}
\author{
Xiaobo Xia$^{1,2}$ \ \ \ \ \
Xiaofeng Liu$^{3,4}$ \ \ \ \ \ 
Jiale Liu$^{5}$ \ \ \ \ \
Kuai Fang$^{6}$ \ \ \ \ \ \\
\textbf{Lu Lu}$^{2}$ \ \ \ \ \
\textbf{Samet Oymak}$^{7}$ \ \ \ \ \
\textbf{William S. Currie}$^{3,4}$ \ \ \ \ \
\textbf{Tongliang Liu}$^{1\dagger}$ \\
\small$^1$School of Computer Science, University of Sydney, Sydney, NSW 2008, Australia \\
\small$^2$Department of Statistics and Data Science, Yale University, New Haven, CT 06511, USA \\
\small$^3$Michigan Institute for Data and AI in Society, University of Michigan, Ann Arbor, MI 48109, USA \\
\small$^4$School for Environment and Sustainability, University of Michigan, Ann Arbor, MI 48109, USA\\
\small$^5$College of Information Science and Technology, Penn State University, University Park, PA 16802, USA\\
\small$^6$Department of Earth System Science, Stanford University, Stanford, CA 94305, USA\\
\small$^7$Department of Electrical Engineering and Computer Science, University of Michigan, Ann Arbor, MI 48109, USA\\
}

\begin{document}
\maketitle
\def\thefootnote{$\dagger$}\footnotetext{Corresponding author~(tongliang.liu@sydney.edu.au).}
\begin{abstract}
Water quality is foundational to environmental sustainability, ecosystem resilience, and public health. Deep learning offers transformative potential for large-scale water quality prediction and scientific insights generation. However, their widespread adoption in high-stakes operational decision-making, such as pollution mitigation and equitable resource allocation, is prevented by unresolved trustworthiness challenges, including performance disparity, robustness, uncertainty, interpretability, generalizability, and reproducibility. In this work, we present a multi-dimensional, quantitative evaluation of trustworthiness benchmarking three state-of-the-art deep learning architectures: recurrent (LSTM), operator-learning (DeepONet), and transformer-based (Informer), trained on 37 years of data from 482 U.S. basins to predict 20 water quality variables. Our investigation reveals systematic performance disparities tied to process complexity, data availability, and basin heterogeneity. Management-critical variables remain the least predictable and most uncertain. Robustness tests reveal pronounced sensitivity to outliers and corrupted targets; notably, the architecture with the strongest baseline performance (LSTM) proves most vulnerable under data corruption. Attribution analyses align for simple variables but diverge for nutrients, underscoring the need for multi-method interpretability. Spatial generalization to ungauged basins remains poor across all models. This work serves as a timely call to action for advancing trustworthy data-driven methods for water resources management and provides a pathway to offering critical insights for researchers, decision-makers, and practitioners seeking to leverage artificial intelligence (AI) responsibly in environmental management.
\end{abstract}

\section*{Introduction}
\vspace{-0.3cm}
Water quality is essential for both environmental sustainability and public health, and clean water supports aquatic biodiversity, ecosystem services, and safe drinking water access~\cite{dudgeon2006freshwater,mckeown2015impact}. However, freshwater contaminants increasingly threaten ecological integrity and human well-being by causing habitat degradation, species loss, and waterborne diseases~\cite{du2022persistent}. In response, substantial efforts have been made, notably with the U.S. investing more than \$1.9 trillion since 1960 to reduce pollution in rivers, lakes, and other surface waters, a commitment exceeding most other national environmental initiatives~\cite{keiser2019low}. Despite these investments, persistent challenges remain. Water quality monitoring and sampling are costly and labor-intensive, producing sparse and heterogeneous datasets that limit their value for informed decision-making~\cite{liu2021chlorophyll,babatunde2024study}. Traditional modeling approaches further compound these challenges. Process-based models, although grounded in physical and biogeochemical principles, require in-depth domain knowledge, intensive parameterization, and substantial computational resources, often making them difficult to scale or generalize across diverse hydrological and climatic regimes. Empirical and statistical methods, however, often oversimplify the inherent non-linearity and complex interactions among climatic, hydrological, and anthropogenic drivers of water quality, limiting their predictive power.

Artificial Intelligence (AI) has provided a new set of powerful tools to learn directly from large, heterogeneous environmental datasets and uncover patterns not easily captured by process-based and statistical models. Long short-term memory (LSTM) models, renowned for their ability to capture long temporal dependencies in time series data, have become the most widely applied machine learning methods in hydrological modeling, demonstrating strong performance for streamflow, sediments, dissolved oxygen, and nutrients prediction~\cite{kratzert2019toward,shen2018transdisciplinary,zhi2024deep,zhi2024increasing}. Beyond recurrent networks, operator-learning frameworks such as DeepONet show promise for improving model transferability across watersheds and leveraging ensemble simulations~\cite{sun2024bridging}. Attention-based transformer architectures~\cite{vaswani2017attention}, originally developed for Natural Language Processing, have demonstrated superior performance across many scientific domains and are now being also applied to rainfall-runoff and water quality prediction~\cite{liu2025rnns,liuself}. Together, these recurrent-, operator-, and attention-based models represent the major families of deep learning architectures currently advancing hydrology and water quality modeling.

Despite their predictive power, AI models face a significant ``trustworthiness gap'' that limits their adoption in environmental management. For instance, a model that underestimates spikes in nutrient loads during extreme rainfall events could delay warnings for harmful algal blooms, directly risking public health. Similarly, as highlighted by the 2021 cyberattack on the Oldsmar, Florida water treatment plant demonstrated, vulnerabilities in any part of our water infrastructure, including AI-enabled tools, can have immediate and dangerous consequences. Beyond these risks, a lack of explainability can mislead management efforts. For example, if a model incorrectly attributes water quality drivers to static watershed properties rather than climate extremes, managers may invest in costly long-term land-use changes instead of more effective real-time flow regulation. To be truly useful, AI models must therefore be not only accurate but also trustworthy.

Trustworthiness in AI broadly encompasses multiple critical aspects, including but not limited to their fairness in both modeling processes and outcomes, robustness against noise and adversarial disruption, uncertainty quantification, explainability of model decisions, generalizability to unseen conditions, and reproducibility of research~\cite{li2023trustworthy,liang2022advances,eshete2021making,kurakin2018adversarial,tomsett2020rapid}. While building trustworthy data-driven methods has long been a vision and key development goal in many fields of science and engineering~\cite{zhu2023reliable}, such as healthcare, autonomous driving, sentiment analysis, and climate science~\cite{markus2021role,fernandez2021trustworthy,wang2024sentiment,mcgovern2024value}, their adoption in environmental research lags considerably behind. Most current deep learning research in this domain has largely focused on improving performance metrics (e.g., predictive accuracy), often overlooking these essential trustworthiness considerations that ultimately determine whether AI-enabled systems can be reliably and responsibly deployed in operational decision-making contexts. 

To address these concerns, in this work, we present a multi-dimensional, quantitative evaluation of trustworthiness in deep learning models for large-scale water quality prediction. We benchmark LSTM, DeepONet, and Informer models across six key dimensions: (1) performance across variables and basin types, (2) robustness to outliers, random noise, and adversarial perturbations, (3) model- and data-based uncertainties, (4) consistency among feature importance methods in identifying key drivers, (5) generalizability to unseen basins, and (6) reproducibility. Using 37 years of water quality observations, hydroclimate forcings, and static basin attributes across 482 U.S. basins, we train and evaluate these models to predict 20 water quality variables representing physical/chemical, geochemical weathering, and nutrient cycling processes. Our main contribution is the development of a reproducible trust benchmarking protocol for deep learning in water quality prediction. Beyond identifying technical limitations, we link model behaviors to underlying watershed processes, data characteristics, and hydrological complexity, providing actionable insights for researchers, regulators, and policymakers. In addition, the protocol is readily adaptable to other architectures or application domains, providing a foundation for advancing trustworthy AI in environmental sciences.

Developing trustworthy deep learning models for water quality prediction is also a critical step toward achieving global sustainability goals. This work directly supports SDG 6 (Clean Water and Sanitation) by improving the capacity to monitor pollutants, anticipate risks, and inform equitable water resource allocation. By emphasizing transparency, reproducibility, and collaborative development, it also advances SDG 17 (Partnerships for the Goals).

\section*{Results and Discussion}
\vspace{-0.3cm}
\section*{Challenges in predicting management-critical water quality variables}
\begin{figure}[]
    \centering
    \includegraphics[width=0.95\linewidth]{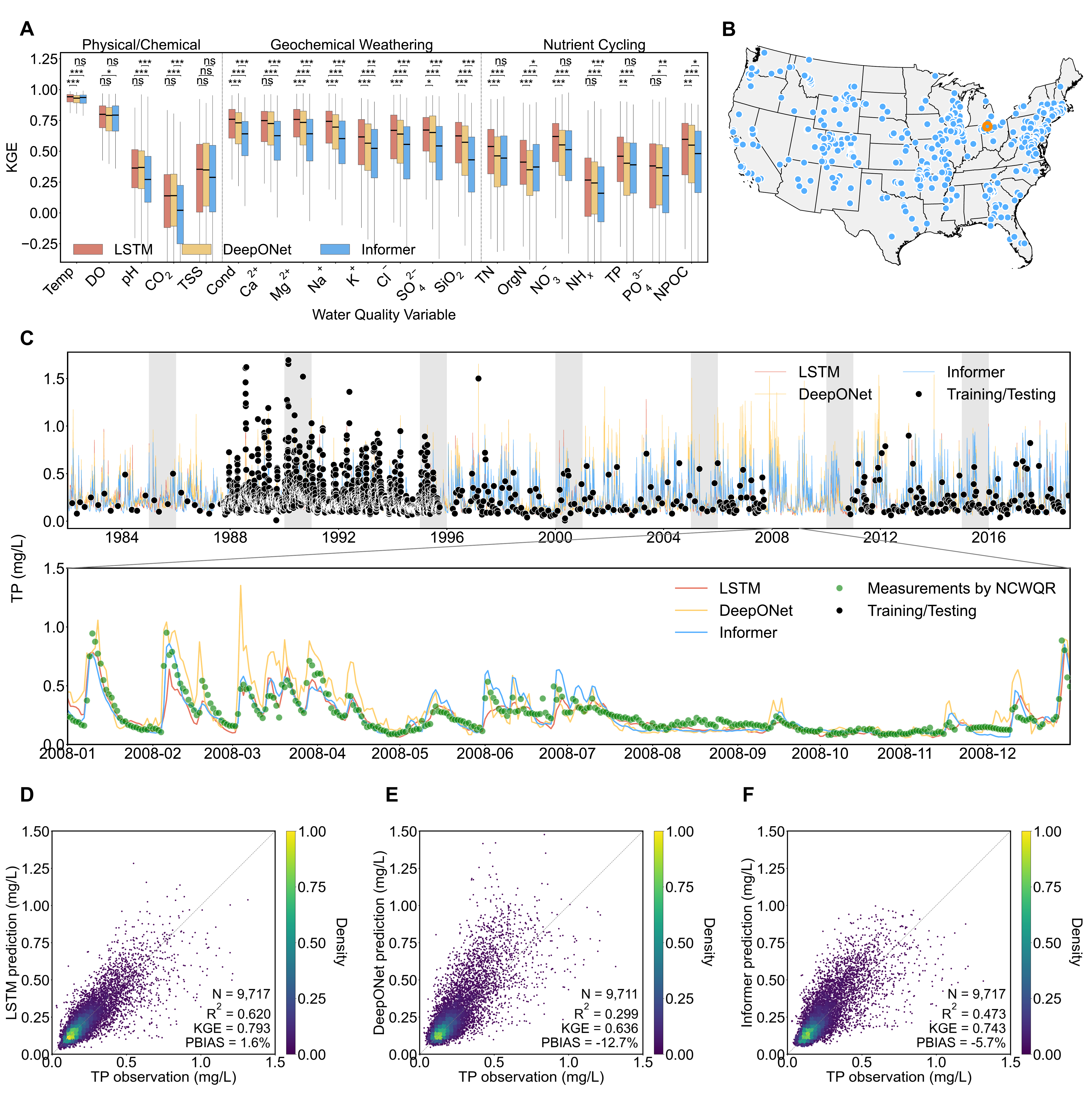}
    \captionsetup{format=plain}
    \caption{\setlength{\baselineskip}{1.5\baselineskip}\textbf{Multi-task deep learning models performance for water quality predictions across the continental United States~(CONUS).} (\textbf{A}) Boxplot of Kling-Gupta Efficiency (KGE) for the testing periods (1985, 1990, 1995, 2000, 2005, 2010, and 2015) across 20 predicted water quality variables associated with physical/chemical properties, geochemical weathering processes, and nutrient cycling. Boxes show the median (central line), the interquartile range (IQR; Q1-Q3), and whiskers extending to $\text{Q1}-1.5\times\text{IQR}$ and $\text{Q3} + 1.5\times\text{IQR}$. Wilcoxon signed-rank p-values ($^{***}p < 0.001$, $^{**}p < 0.01$, $^{*}p < 0.05$, and ``ns'' $p\geq 0.05$) were adjusted using Benjamini-Hochberg false discovery rate (FDR). (\textbf{B}) Locations of 482 riverine monitoring sites used in this study. (\textbf{C}) Example time series of total phosphorus (TP) showing model predictions, training/testing samples, and additional daily observations (not used in training or testing) collected by the National Center for Water Quality Research (NCWQR) at the Maumee River in Waterville, OH (orange circle in panel (\textbf{B})) during 2008. (\textbf{D-F}) Scatter plots comparing predicted TP concentrations from three deep learning models with independent NCWQR observations, with PBIAS indicating percentage bias (observation minus simulation).}
    \label{fig:1}
\end{figure}

\vspace{-0.3cm}
Our continental-scale, multi-task deep learning models exhibit a wide range of predictive performance (as measured by Kling-Gupta Efficiency, KGE, see Methods) across different water quality variables. Overall, all three models achieve high median KGE values for temperature (0.94, 0.93, 0.94 for LSTM, DeepONet, and Informer, respectively) and DO (dissolved oxygen, 0.80, 0.79, 0.79) (Fig.~\ref{fig:1}A). They also demonstrate moderate predictive accuracy for variables associated with geochemical weathering processes, including Cond (conductivity), $\text{Mg}^{2+}$, $\text{K}^{+}$, and $\text{SiO}_2$~(median KGE: 0.62-0.76 for LSTM, 0.57-0.73 for DeepONet, and 0.43-0.64 for Informer). However, all models fail to capture the dynamic variability of $\text{CO}_2$, pH, and TSS~(total suspended sediments) and underperform in nutrient-related variables, particularly $\text{NH}_\text{x}$ and $\text{PO}_4^{3-}$ (median KGE $<$ 0.4).

To further investigate model behavior, we evaluated TP predictions at an agriculturally intensive basin where additional daily observations (not in either the training or testing datasets) are available (Fig.~\ref{fig:1}B, C). At this site, DeepONet achieves the highest KGE (0.64) on the standard testing set, followed by Informer (0.57) and LSTM (0.55). However, when evaluated using the independent high-frequency observations, LSTM exhibits the most consistent performance, while DeepONet shows poor generalizability, significantly overestimating TP concentrations (PBIAS = -12.7\%; Fig.~\ref{fig:1}D, E, F). These results highlight a broader trustworthiness concern: models trained and validated on national datasets may not generalize reliably across all observational contexts. For large-scale models to support operational decisions such as nutrient regulation and harmful algal blooms (HABs) risk assessments, it will likely require integrating watershed process knowledge, high-frequency monitoring, and ensemble strategies to reconcile continental-scale accuracy with local-scale reliability.

Across model types, LSTM achieves the best overall performance, followed by DeepONet, while Informer generally underperforms. These patterns reflect both the unique characteristics of the hydrological process and model architectures. LSTM leverages recurrent memory to capture seasonal cycles and hysteresis effects in flow-water quality relationships, which dominate many variables~\cite{agrawal2025improving}. DeepONet, while effective at learning broad functional relationships, is prone to overfitting the continental-scale training distribution, which limits its ability to generalize at local scales. Transformer-based models such as Informer are typically data-hungry and require large training sets to demonstrate their superiority~\cite{liu2024probing}, which however is difficult to realize with the sparse and irregular water quality data. Although most recent foundation models are transformer-based and demonstrate state-of-the-art performance in many domains~\cite{zhou2024comprehensive}, hydrological applications have been shown as an exception where LSTM can outperform transformer-based models, especially for regression tasks~\cite{liu2025rnns}. In terms of computational efficiency, DeepONet is the most time-efficient since it bypasses explicit sequential recurrence or attention, directly mapping temporal windows and static attributes through multilayer perceptrons. LSTM requires recurrent processing over the 365-day input sequence, which increases training time but remains more lightweight in parameter size compared to Informer. Informer also processes the 365-day sequence, but its encoder-decoder attention blocks and feed-forward layers introduce the highest computational overhead, leading to longer training times and larger memory usage despite parallelization.

Nutrient variables (e.g., various forms of nitrogen and phosphorus) are among the most challenging variables for both deep learning and hydrological benchmark models (e.g., WRTDS)~\cite{fang2024modeling}. These difficulties arise likely because nutrient concentrations often spike or fluctuate in response to episodic events (e.g., fertilizer applications, storm runoff) and involve reactive processes in both soils and streams. Phosphorus concentrations are also sensitive to long-term legacies of P accumulation in soils. Deep learning models, which primarily learn from historical input-output patterns, struggle to represent these mechanistic and legacy-driven dynamics. Additionally, routine water quality monitoring for nutrients is typically infrequent (e.g., weekly or monthly grab samples), causing responses to episodic events and associated rapid dynamics to be under-sampled and thus underrepresented in training datasets. Consequently, both the sparsity and the process complexity of nutrient data undermine the models' predictive ability where however accurate estimates are often most critical for water quality management. 

\begin{figure}[]
    \centering
    \includegraphics[width=0.95\linewidth]{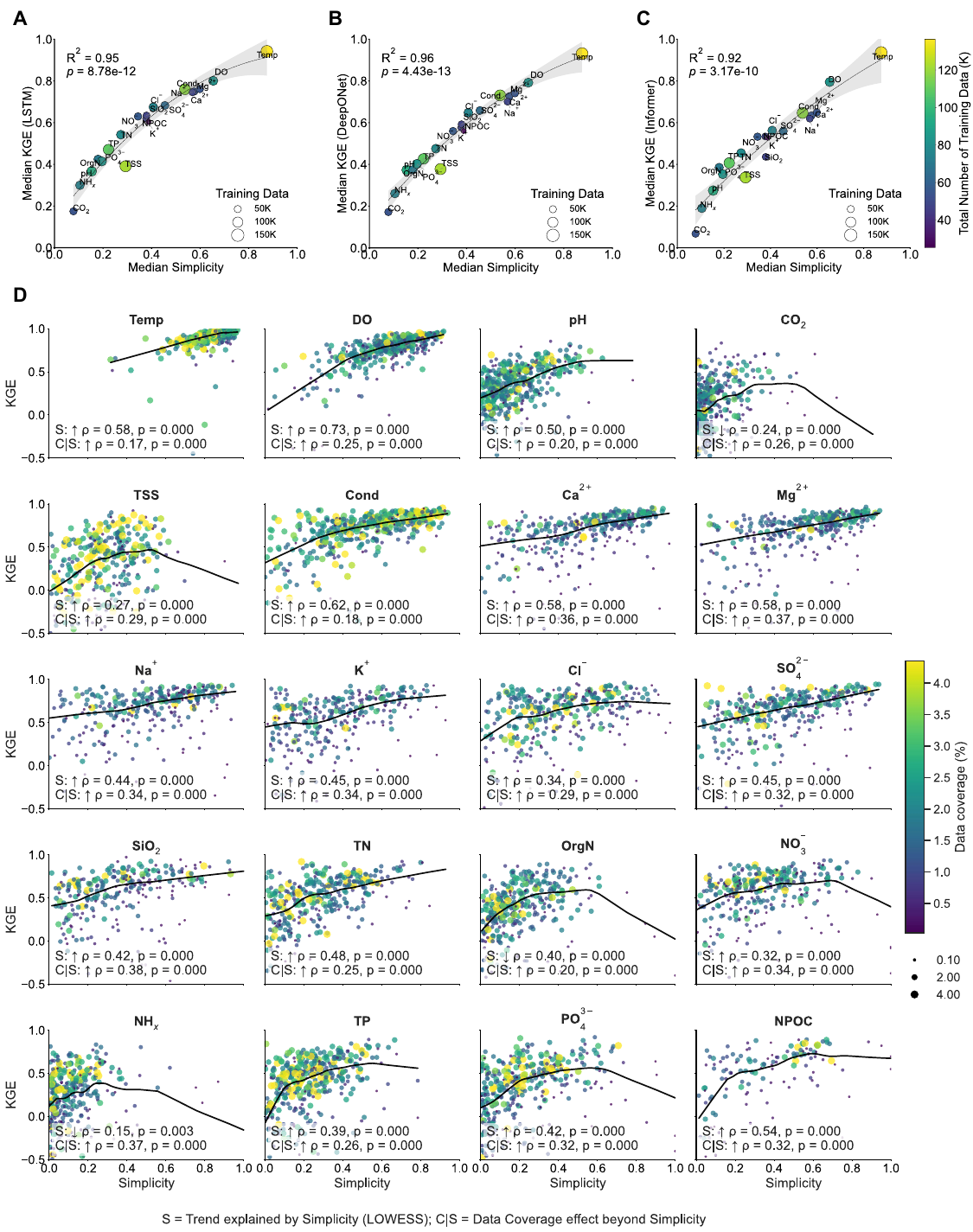}
    \captionsetup{format=plain}
    \caption{\setlength{\baselineskip}{1.5\baselineskip}\textbf{Relationships between model performance, simplicity, and training sample size across 20 water quality variables for LSTM (A), DeepONet (B), and Informer (C), and basin-level relations for LSTM (D).} In panels (\textbf{A-B}), each dot represents one variable. Model performance is represented by the median KGE across CONUS, while the simplicity index measures the proportion of variance in water quality explained by linear relationships with runoff and annual cycles \cite{fang2024modeling}. Both the size and color of each dot indicate the number of training samples, with larger, yellow}
    \label{fig:2}
\end{figure}
\begin{figure}[]
    \ContinuedFloat
    \caption*{\setlength{\baselineskip}{1.5\baselineskip}dots representing more data. The grey shade represents the 95\% confidence interval of the polynomial regression line. In panel (\textbf{D}), each dot represents a basin and both the dot's color and size encode data coverage. A locally weighted scatterplot smoothing (LOWESS) curve summarizes the relationship between model performance (KGE) and simplicity. The arrow marks the LOWESS slope at the highest simplicity, indicating whether performance tends to increase or decrease with simplicity. Each panel reports Spearman's correlation coefficient ($\rho$) and p-value for: (1) KGE vs. simplicity, and (2) data coverage vs. LOWESS residuals (i.e., the data coverage effect conditional on simplicity), where the residual is computed as the observed KGE minus the LOWESS predicted KGE at the same simplicity. A positive value indicates that, at fixed simplicity, higher data coverage is associated with higher-than-expected performance (KGE). Analogous figures for DeepONet and Informer are provided in Fig.~\ref{fig:s3} and \ref{fig:s4}.}
    \label{fig:part2}
\end{figure}

\section*{Performance disparities due to variable simplicity, data availability, and basin heterogeneity}
\vspace{-0.3cm}
In water quality prediction, achieving consistent predictive performance across diverse geographic, environmental, and socio-economic contexts is desirable to avoid systematically disadvantaging particular regions. However, performance disparities often reflect differences in data availability and the inherent complexity of the system being modeled rather than model unfairness.

Comparing across variables, the amount of training data alone does not explain observed disparities (Fig.~\ref{fig:2}A-C). Most of the weathering variables achieve moderate accuracy (median KGE $>$ 0.5) even with fewer training samples, whereas nutrient variables exhibit low performance even when trained with more data than the former. Instead, a ``simplicity index"~\cite{fang2024modeling}, which quantifies the proportion of variance in water quality dynamics explained by linear relationships with runoff and annual cycles, shows a strong correlation with model performance across variables~($\text{R}^2$ = 0.92-0.96, $p<0.001$) in all three models. This suggests that model performance disparities largely depend on the inherent predictability of water quality variables, driven by hydrological or seasonal patterns. 

Building on these cross-variable patterns, basin-level characteristics further shape the model performance of individual variables. Basins with greater temporal data coverage achieve higher realized KGE after controlling for simplicity. Locally weighted scatterplot smoothing (LOWESS) curves are upward-sloping for more than 16 variables (Fig.~\ref{fig:2}D, Figs.~\ref{fig:s3},~\ref{fig:s4}). Temperature and dissolved oxygen show the strongest association (Spearman's $\rho>0.50$, $p<0.001$), and all weathering variables exhibit consistently strong positive trends (mean $\rho=0.48$ across three models, $p<0.001$). In contrast, event- and source-driven variables, such as TSS, OrgN, $\text{NH}_\text{x}$, and $\text{PO}_4^{3-}$, show weaker or more curved relationships, with flattening or slight downturns at high simplicity, indicating performance ceilings and limited data at very high simplicity. Low-coverage sites tend to fall below the LOWESS line at the same simplicity. After removing the simplicity trend, data coverage still explains additional variation in KGE for most variables (mean $\rho=0.28$, $p<0.001$). In other words, even among basins of similar simplicity, higher temporal coverage is associated with higher realized model performance.

These variable- and basin-level patterns manifest as systematic differences across land use types (Figs.~\ref{fig:s5}, \ref{fig:s6}, \ref{fig:s7}). Nutrient variables, such as TN, $\text{NO}_{3}^-$, TP, and $\text{PO}_4^{3-}$ are more reliable in agricultural, urban, and mixed basins, where fertilizer applications and human activities (e.g., discharge of wastewater treatment plants) yield more consistent and predictable concentration-runoff (C-Q) relationships. In contrast, in undeveloped basins, nutrient predictions remain problematic due to compounding challenges: limited data coverage (Fig.~\ref{fig:s8}), inherently low simplicity (Fig.~\ref{fig:s9}), and possible signal masking by higher concentrations from the urban, mixed, and agricultural basins. Nevertheless, NPOC is an exception, which achieves relatively better performance in undeveloped basins (though not statistically significant), a pattern consistent with terrestrial carbon export mechanisms rather than anthropogenic point sources. For weathering variables, performances are higher in undeveloped basins, except for $\text{SiO}_2$, $\text{K}^+$, and $\text{Cl}^-$. Temperature predictions show significantly reduced performance in undeveloped western mountain basins, suggesting an inherent bias likely due to complex snowmelt-driven thermal variability that disproportionately impacts these regions.

These results highlight two key insights. First, the strong positive simplicity-performance relationship indicates that the simplicity index is a useful tool for prioritizing basins and variables: basins and variables with higher simplicity are reliably easier to model, and many approach a performance ceiling. Second, monitoring intensity has impacts independently of simplicity: insufficient coverage reduces KGE beyond what would be expected from ``inherent" predictability alone, particularly for nutrients where more frequent sampling is necessary. It is noted that while residual-based analysis controls for simplicity, unmeasured confounders, such as data quality, may still mediate performance.

Addressing these disparities requires both data-centric strategies (expanded monitoring in low-coverage basins, integration of auxiliary drivers such as fertilizer timing, livestock, and industrial discharges if available) and model-centric advances (such as hybrid process-ML approaches). These steps are critical to ensure that AI-driven predictions are equitable across regions, supporting fair and effective water quality management.
\vspace{-0.3cm}

\begin{figure}[]
    \centering
    \includegraphics[width=1.0\linewidth]{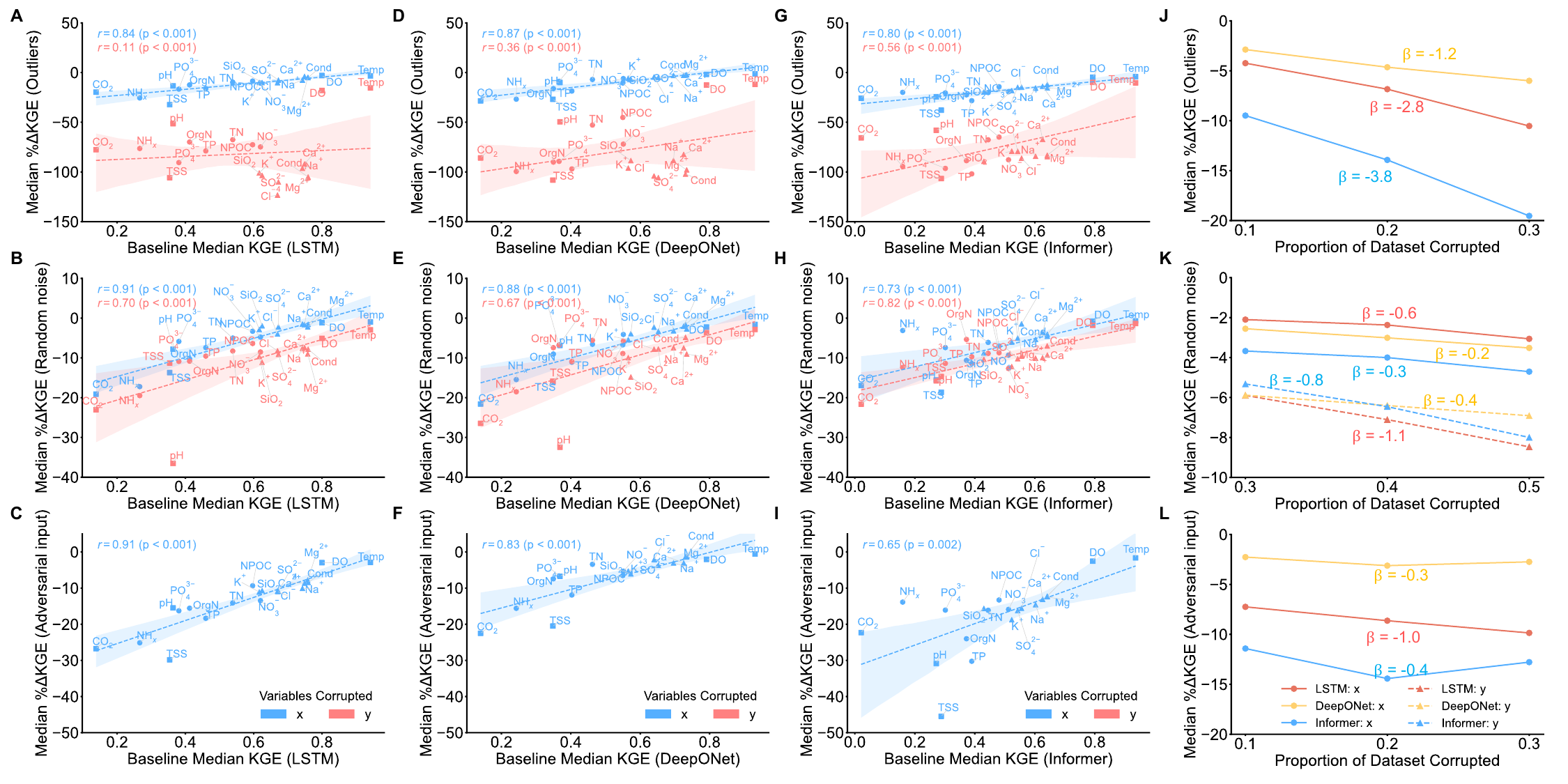}
    \caption{\setlength{\baselineskip}{1.5\baselineskip}\textcolor{black}{\textbf{Robustness of three deep learning models under dataset corruptions.} (\textbf{A}-\textbf{I}) Scatterplots of the median percent change in KGE relative to the uncorrupted baseline for each model (columns) and data corruption types (rows). Blue dots represent corruptions applied to input features (x) and red dots represent corruptions applied on targets (y). The fitted line shows the Pearson correlation between baseline median KGE and percent change (shaded 95\% CI), reflecting how model vulnerability relates to baseline performance. (\textbf{J}-\textbf{L}) Aggregate robustness curves plotting the median percent change in KGE (across all basins and variables) versus the proportion of the dataset corrupted. The Theil-Sen median station-level slope $\beta$ is used to quantify the model performance degradation rate and is interpreted as the expected percent change in KGE per 0.1 (10\% of the dataset) increase in corruption.}}
    \vspace{-5pt}
    \label{fig:3}
\end{figure}

\section*{Limited robustness to outliers, random noise, and adversarial disruptions}
\vspace{-0.3cm}
\textbf{Robustness against outliers.} All three models exhibit pronounced sensitivity to outliers, a critical limitation given the prevalence of extreme values in environmental datasets (e.g., storm-driven pollutant spikes, sensor malfunctions, and measurement errors) (Fig.~\ref{fig:3}A, D, G). Introducing 10-30\% synthetic outliers (see Methods) into input features led to median performance declines of up to 28.3\% across nutrients, 19.6\% across weathering variables, and less than 5\% for DO and temperature. The most affected variables are $\text{CO}_2$, TSS, and $\text{NH}_\text{x}$. Across variables, higher baseline predictive performance is strongly correlated with lower sensitivity to input outliers ($r$ = 0.80-0.87, $p < 0.001$). Across model types, DeepONet exhibits the greatest robustness to increasing data corruption (Theil-Sen slope $\beta$ = -1.2 per 10\% increment; Methods), while Informer declines most rapidly ($\beta$ = -3.8) (Fig.~\ref{fig:3}J).

However, outliers injected directly into water quality measurements cause disproportionately greater impacts. Specifically, for 18 out of 20 variables, more than half of the basins show KGE reductions exceeding 50\% except for temperature and DO. Surprisingly, variables associated with geochemical weathering demonstrate a greater overall performance decline than nutrient variables in LSTM and DeepONet, while Informer shows little distinction between groups. This pronounced impact on weathering-related variables also disrupts the previously observed correlation between the baseline median KGE and the percent change in KGE ($r$ = 0.11-0.36, Fig.~\ref{fig:3}A, D, G). 

\textbf{Robustness against random noise.} Compared with outliers, perturbing input features or targets with random Gaussian noise (see Methods) causes relatively smaller but still meaningful performance declines across all three models (Fig.~\ref{fig:3}B, E, H). As with outliers, noise applied to targets produces larger impacts than noise in input features, demonstrating the importance of accurate water quality measurements. Compared with the irregular impacts of outliers, performance degradations caused by random noise in targets are more predictable: correlations between baseline KGE and the percent change in KGE are strong ($r$ = 0.70 for LSTM, 0.67 for DeepONet, and 0.82 for Informer). Across models, DeepONet shows the highest robustness, with only modest declines as the proportion of corrupted data increased ($\beta$ = -0.2 for features and -0.4 for targets), while LSTM is most sensitive to random noise (-0.6 and -1.1 for features and targets, respectively) (Fig.~\ref{fig:3}K).

\textbf{Robustness against adversarial disruptions.} Unlike outliers or random noise, adversarial perturbations are systematically optimized to maximize prediction errors, making them a critical stress test for model robustness. When 10-30\% adversarial noise (see Methods) was introduced to corrupt input features, geochemical weathering variables exhibit greater robustness to adversarial noise compared to nutrient variables. Specifically, the average median KGE decline in weathering group is 9.81\%, 3.23\%, and 14.8\% in LSTM, DeepONet, and Informer, respectively, compared with 16.03\%, 8.1\%, and 18.5\% for nutrient variables (Fig.~\ref{fig:3}C, F, I). Similarly, strong positive correlations between the median baseline KGE and the percentage reduction of KGE further reveal that variables with initially lower performance would be more sensitive to input adversarial noise. Across models, DeepONet again demonstrates the greatest robustness ($\beta$ = -0.3), while LSTM is the least robust ($\beta$ = -1.0). These results demonstrate that adversarial vulnerabilities, commonly highlighted in Computer Vision and Natural Language Processing tasks, are equally relevant to hydrological prediction. Importantly, such vulnerabilities are not limited to academic experiments: the 2021 cyberattack on the Oldsmar, Florida water treatment plant, where hackers attempted to alter sodium hydroxide concentrations in the public supply, highlights the real-world risks of adversarial disruptions to water infrastructure. As AI-based water quality prediction models become increasingly integrated into monitoring and management frameworks, ensuring their robustness to both accidental and intentional perturbations will be critical for safeguarding public health and environmental decision-making.

\begin{figure}[!tp]
    \centering
    \includegraphics[width=0.9\linewidth]{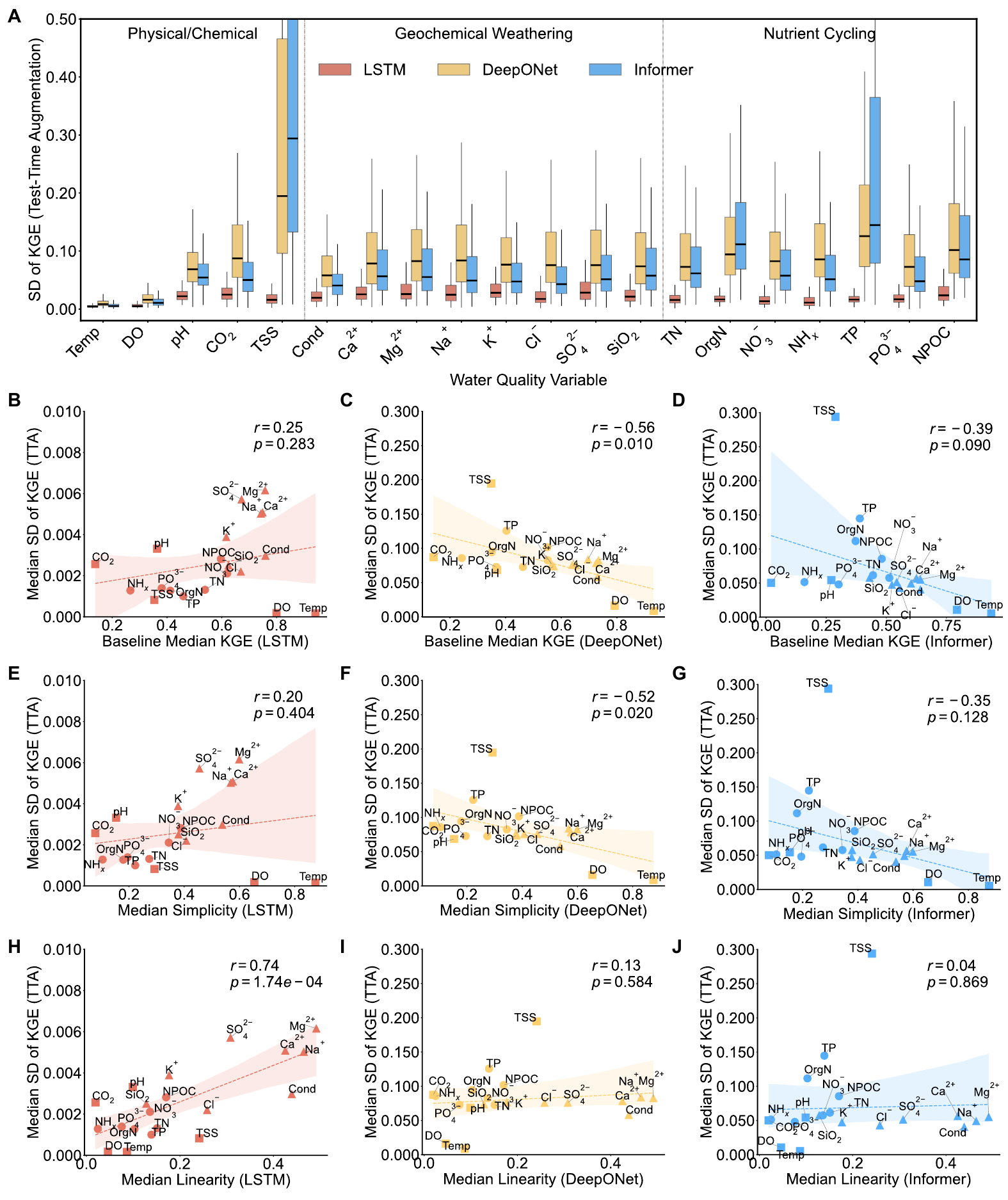}
    \caption{\setlength{\baselineskip}{1.5\baselineskip}\textcolor{black}{\textbf{Model prediction uncertainty across water quality variables and its relationship with baseline performance, variable simplicity, and linearity.} (\textbf{A}) For each water quality variable and deep learning model, prediction uncertainty is quantified as the standard deviation (SD) of the Kling-Gupta Efficiency (KGE) over 50 test-time augmentation (TTA) runs (see Methods). Boxplots show the median (central line), interquartile range (IQR, represented by the boxes spanning the first (Q1) to the third quartile (Q3)), and whiskers extending to $\text{Q1}-1.5\times\text{IQR}$ and $\text{Q3} + 1.5\times\text{IQR}$. (\textbf{B-D}) For each model (column), per-variable median uncertainty across all basins versus the baseline median KGE. (\textbf{E-G}) As in (\textbf{B-D}), but versus per-variable median simplicity. (\textbf{H-J}) As in (\textbf{B-D}), but versus per-variable median linearity. In (\textbf{B-J}), dashed lines are least-squares fits with 95\% CI; Pearson’s r and corresponding p-values are reported in each panel.}}
    \label{fig:4}
\end{figure}

A key finding from the robustness evaluation is the contrasting behavior between models with high predictive skill and those with greater resilience to data corruption. LSTM, which achieves the highest baseline performance under clean data conditions, exhibits the greatest vulnerability when input features or target variables are perturbed. In contrast, DeepONet, despite slightly lower baseline accuracy, maintains higher robustness across corruption types, especially for adversarial disruption. This trade-off likely arises from differences in model architectures and learning mechanisms: the LSTM’s recurrent structure enables strong temporal pattern fitting but also amplifies the propagation of small errors through its hidden states, making it highly sensitive to data corruptions. DeepONet, however, learns generalized functional mappings between input and output spaces through operator learning, which imposes an intrinsic smoothness and regularization on its response surface, therefore stabilizing its predictions under imperfect data. This trade-off has important implications for practical water quality modeling. In controlled research settings with high-quality and dense datasets, LSTM may achieve better predictive accuracy. However, in real-world monitoring networks characterized by sparse, noisy, and irregular sampling, DeepONet’s architectural resilience ensures more robust performance. For operational deployment, model selection should therefore balance predictive fidelity with robustness, ensuring that predictions can inform regulatory enforcement, environmental risk assessment, and adaptive water quality management under real-world data conditions. 

For environmental management, improving robustness requires an integrated data-centric and model-centric approach. On the data side, implementing strict quality assurance and control (QA/QC) protocols for sensor/sample data is necessary to reduce noise and errors in training datasets. Leveraging domain knowledge to identify variables most susceptible to corruption can further improve model performance by allowing targeted preprocessing efforts. On the model side, incorporating robustness-oriented learning strategies, such as adversarial training~\cite{bai2021recent} and distributionally robust optimization~\cite{lin2022distributionally}, can improve resilience against both random perturbations and targeted attacks. 

Beyond robustness to synthetic data corruption, it is also important to recognize that many extreme high and low values in hydrological datasets represent real system-critical processes. Events such as storm-driven sediment pulses, fertilizer runoff peaks, and drought-induced concentration spikes carry valuable information about watershed responses to environmental change. The pronounced sensitivity of all three models to these extremes suggests that current architectures, optimized primarily for average conditions, may underrepresent the nonlinear dynamics governing these events. Improving model capacity to capture such behaviors may require targeted training strategies, such as event-aware sampling~\cite{frame2022deep}, imbalance-corrected loss functions~\cite{wang2024self,ding2019modeling}, or explicit inclusion of mechanistic process constraints~\cite{radfar2025integrating, song2025high}, to prevent models from over-smoothing rare but hydrologically important events.

\section*{Higher prediction uncertainty in management-critical water quality variables}
\vspace{-0.5cm}
Accurate and reliable uncertainty quantification is essential for deploying deep learning models for water quality management. Our test-time augmentation (TTA) analysis (see Methods) shows that introducing Gaussian noise ($\sigma=0.1$) to streamflow inputs only leads to the highest predictive uncertainty (quantified as the standard deviation (SD) of the KGE) in DeepONet (average = 0.081), intermediate in Informer (0.069), and the lowest in LSTM (0.003) (Fig.~\ref{fig:4}A). However, when the noise was added to all dynamic input features, LSTM uncertainty increases from 0.003 to 0.018, suggesting that measurement error in meteorological or other hydrological drivers could amplify prediction uncertainty (Fig.~\ref{fig:s10}). 

Patterns of predictive uncertainty vary systematically across both water quality variables and model architectures. Across variables, temperature and DO are consistently the most stable, while TP and TSS show the greatest uncertainty in both DeepONet and Informer (0.195 and 0.126 in DeepONet; 0.294 and 0.145 in Informer), but not in LSTM (Fig.~\ref{fig:4}A). LSTM exhibits less uncertainty for nutrients relative to weathering variables, while DeepONet and Informer show the opposite, with the highest instability in nutrients. Moreover, the relationship between model uncertainty and hydrologic characteristics is architecture-dependent (Fig.~\ref{fig:4}E-J). DeepONet demonstrates significant negative correlations between uncertainty and both baseline KGE ($r=-0.56$, $p=0.01$; Fig.~\ref{fig:4}C) and simplicity ($r=-0.52$, $p=0.02$; Fig.~\ref{fig:4}F), suggesting that simpler, runoff-seasonality-driven variables are both more accurate and more stable. In contrast, LSTM uncertainty increases with linearity (quantified by the proportion of variance explained by linear relationships with runoff) ($r=0.74$, $p<0.001$; Fig.~\ref{fig:4}H), indicating that when water quality variables are highly correlated with runoff, small perturbations in discharge propagate and amplify through the recurrent states. Informer shows no significant relationships, implying less interpretable but more uniformly distributed uncertainty. These divergent patterns reveal that predictive performance and predictive stability are not always aligned and that uncertainty-simplicity relationships can serve as diagnostics of model trustworthiness.

Monte Carlo (MC) dropout analysis (see Methods) of LSTM demonstrates higher absolute uncertainties compared to TTA. Nutrient variables, critical for ecosystem health, such as $\text{NH}_\text{x}$, $\text{PO}_4^{3-}$, and TSS exhibit the highest uncertainty (median SD of $\text{KGE} > 0.08$; Fig.~\ref{fig:s11}A),  followed by variables associated with the weathering process (median SD of KGE: 0.06-0.08), while physical/chemical variables like temperature and DO show the least uncertainty (median SD of $\text{KGE} < 0.02$). The strong negative relationship between predictive performance and uncertainty ($r = -0.62$, $p < 0.001$; Fig.~\ref{fig:s11}B) further indicates that the most management-critical variables are also the least reliably predicted. Importantly, the discrepancy between MC dropout and TTA suggests that LSTM predictions may appear stable under input perturbations but remain structurally unstable when model uncertainty is accounted for. Thus, TTA provides a lower bound on predictive instability, while MC dropout captures deeper epistemic uncertainty in the model, especially for nutrient variables.

These results highlight three challenges. First, nutrient variables and sediment, critical to Total Maximum Daily Load (TMDL) assessments, algal bloom mitigation, and watershed restoration, however, carry the greatest predictive uncertainty, significantly undermining their operational utility. Second, while LSTM shows lower overall TTA uncertainty, its instability in runoff-dominated regimes suggests a need for more advanced strategies. Finally, sparse monitoring data for nutrient and weathering variables exacerbate uncertainty, particularly in undeveloped basins, highlighting the need for expanded observation networks. These findings suggest that predictive uncertainty must be explicitly quantified and incorporated into management workflows, and that architectural selection, data curation, and uncertainty-aware training~\cite{hu2021uncertainty,einbinder2022training,li2021learning,chua2023tackling} are essential steps toward trustworthy deployment of AI in water quality management.

\begin{figure}[!tp]
    \centering
    \includegraphics[width=1.0\linewidth]{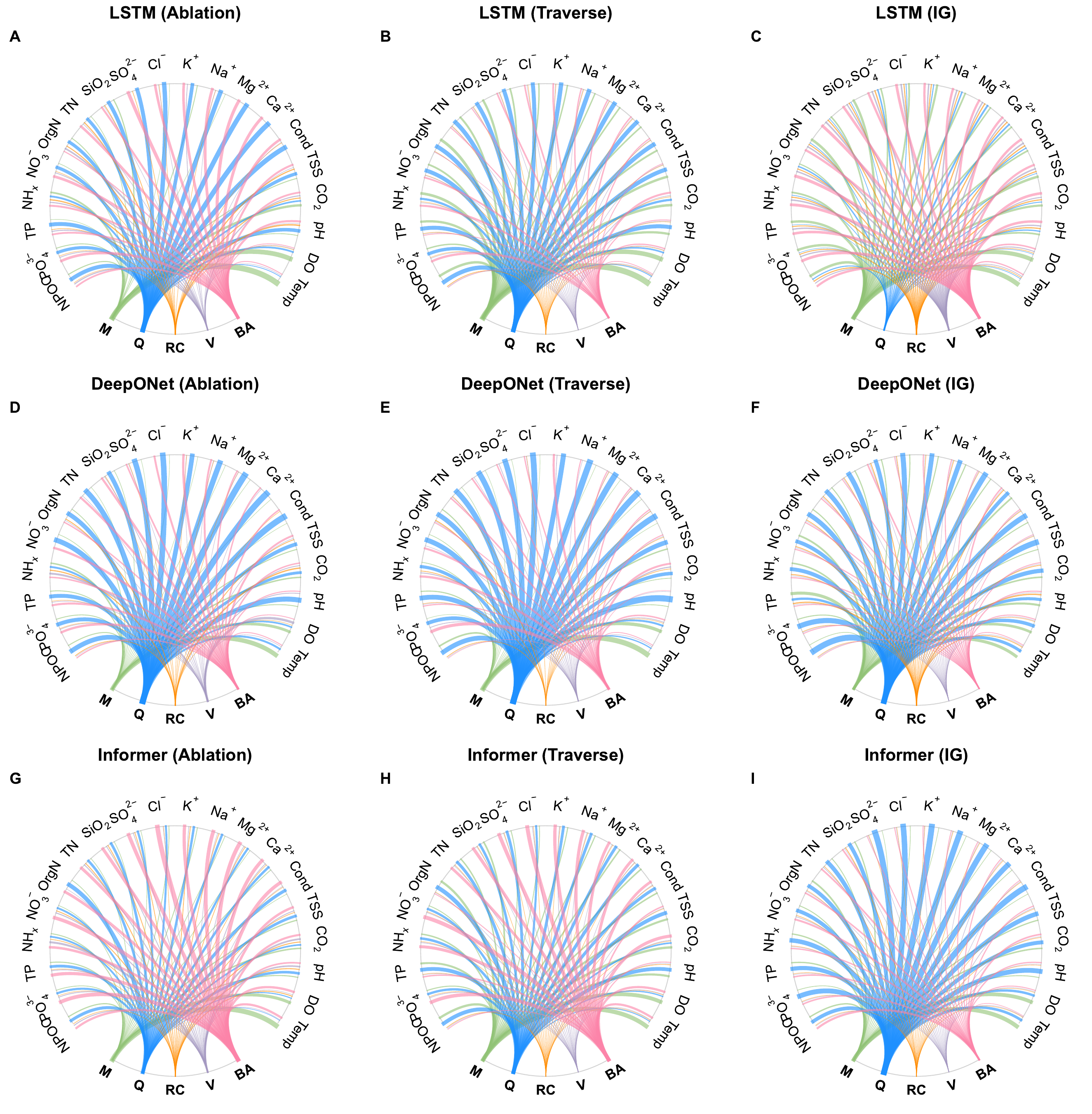}
    \caption{\setlength{\baselineskip}{1.5\baselineskip}\textcolor{black}{\textbf{Group-level feature importance across water quality variables, deep learning models, and attribution methods.} In each panel, the top and bottom arcs list 20 water quality variables and feature groups: meteorological forcings (M), runoff (Q), rainfall chemistry (RC), vegetation indices (V), and basin attributes (BA) (full group composition in Methods). The ribbon width is normalized for each variable so widths to all groups sum to 1, representing the variable’s fractional attribution (comparable across groups for a given variable, but not across different variables). For Ablation (\textbf{A, D, G}), group importance is the percent decrease in Kling-Gupta Efficiency (KGE) when that group is removed from the full model. For Traverse (\textbf{B, E, H}), group importance is the average percent KGE reduction across all model variants with and without the target group (approximating its marginal contribution across subsets). For IG (Integrated Gradients, \textbf{C, F, I}), attributions are computed for each sample; the feature importance is the mean absolute IG over samples, and the group importance is the mean of feature-level $|$IG$|$ within that group.}}
    \label{fig:5}
\end{figure}

\section*{Inconsistent feature importances challenge model interpretability}

Feature importance analysis (see Methods) reveals consistent patterns for some variables but pronounced divergences for others. For temperature and dissolved oxygen, all three models and both performance-based approaches identify meteorological forcings (M) as the dominant feature group, with Integrated Gradients (IG) confirming this in LSTM and Informer (Fig.~\ref{fig:5}). This also aligns with previous findings that highlight the strong influence of air temperature on oxygen dynamics governed by solubility, photosynthesis, and other biological activities~\cite{zhi2023widespread}. Similarly, runoff (Q) consistently emerges as the key driver for pH and TSS, across all attribution methods.

In contrast, attribution results for nutrients and many weathering-related variables diverge across methods and models. In LSTM, surprisingly, IG often emphasizes basin attributes (BA), while Ablation and Traverse highlight Q (Figs.~\ref{fig:5}A-C). This discrepancy likely arises from (1) hydrologic memory and collinearity, where lagged M encodes much of Q’s variability, reducing Q’s local gradients despite its global importance; and (2) gate saturation, where Q inputs pass through saturating gates that suppress marginal gradients, which however needs further investigation. These findings differ from a previous study~\cite{zhi2024increasing}, which reported that LSTM predictions of TP were primarily driven by discharge according to IG. However, direct comparison remains challenging given differences in data sources, model architectures, IG implementations, and the absence of open-source code. For Informer, IG consistently highlights Q, while Ablation and Traverse frequently point to BA, especially for dissolved nutrients (i.e., $\text{NO}_3^-$, $\text{NH}_\text{x}$, and $\text{PO}_4^{3-}$) (Figs.~\ref{fig:5}G-I). This is likely because in our implementation, the attention-based architecture concatenates BA to each token, acting as a persistent identity signal under the same-site temporal splits and improving generalization. While removing BA causes large performance drops (captured by Ablation/Traverse), IG assigns weak local gradients to BA, leading to underestimation. Moreover, our additional experiments show that IG itself can yield inconsistent results depending on model setup: task formulation (single vs multi-task), input representation (raw vs area-normalized streamflow), or features missing-value handling (0 or -1) (Fig.~\ref{fig:s12}).

Performance-based methods also produce inconsistent estimates of meteorological importance. In both LSTM and Informer, Ablation approach underestimates M’s importance, showing no significant KGE reduction in LSTM ($p > 0.05$; Fig.~\ref{fig:s13}) and assigning relatively lower importance in Informer (Fig.~\ref{fig:s14}) compared with the Traverse approach. Further analysis indicates that specifically when the runoff was included, meteorological information appeared redundant (Fig.~\ref{fig:s16}, ~\ref{fig:s17}), which was also demonstrated in a previous study~\cite{fang2024modeling}. A similar pattern emerges for rainfall chemistry (RC) and vegetation indices (V) in LSTM: Ablation demonstrates nonsignificant importance for most variables (except for $\text{NO}_3^-$), whereas Traverse method reveals significant median KGE declines for many weathering parameters ($p < 0.001$; Fig.~\ref{fig:s13}). These inconsistencies align with previous findings that in scenarios with substantial feature overlap, a model may favor one subset of variables without fully reflecting real-world dependencies, causing Ablation to underestimate the value of certain input features~\cite{hooker2019benchmark,catav2021marginal}. Traverse method evaluates a feature group’s contribution across all subsets (e.g., M alone, M + Q, M + BA, etc), isolating standalone and interactive effects. In contrast, Ablation measures only the marginal loss from removing a group from the full model, where overlapping signals (e.g., precipitation information embedded in runoff through the runoff-generating process) mask true importance. However, DeepONet exhibits smaller gaps between Ablation and Traverse (Fig.~\ref{fig:s15}) and shows less sensitivity to whether Q or M are included in feature subsets (Fig.~\ref{fig:s18}), likely because it uses same-day forcings to predict same-day outputs, so inputs do not carry long-lagged information and the overlap between Q and meteorological drivers is inherently limited.

Each attribution method captures a different aspect of model reasoning: Integrated Gradients reflects local sensitivity to small perturbations, and performance-based approaches quantify global and interactive effects among features or feature groups. Rather than treating these differences as contradictions, they should be considered as complementary sources of insight. In practice, whenever possible, interpretability results should be grounded in a multi-method consensus framework, where agreement across methods identifies robust, confident drivers. However, when feature importance rankings diverge, developing a hierarchy of evidence for interpretability, analogous to frameworks in empirical sciences, can provide a transparent basis for decision-making: consistently important features highlight reliable intervention targets (e.g., runoff regulation), while inconsistent results indicate processes that require additional monitoring, process-based modeling, or field validation. Recognizing such differences is critical as attribution outcomes have direct implications for water quality management. For instance, if ablation analyses underestimate meteorological contributions due to redundancy with runoff, managers may overlook the importance of precipitation extremes that intensify sediment-nutrient coupling and eutrophication risks. In addition, if IG overemphasizes BA while underestimating hydrologic controls, interventions could be misdirected toward static watershed properties rather than more responsive flow regulation or climate adaptation measures.

Finally, although our analyses are conducted at the continental scale, effective management depends on identifying local and regional drivers, which may differ from national patterns. Therefore, adopting a multi-method and multi-scale interpretability framework can enhance the credibility and practical relevance of machine learning insights, ensuring that climate, hydrologic, and watershed attributes are all appropriately considered when designing effective strategies to protect and restore freshwater quality under a changing climate and land use.

\section*{Generalization challenges and emerging solutions}

A long-standing challenge in water quality prediction lies in achieving spatial generalization: the ability of models trained in data-rich basins to perform reliably in data-scarce or ungauged basins. However, our spatial training-testing split results (Fig.~\ref{fig:s19}) indicate that all three models exhibit poor generalizability for most water quality variables. The only exceptions are temperature and dissolved oxygen, which achieve relatively high median KGE values: 0.89, 0.88, and 0.91 for temperature, and 0.72, 0.72, and 0.74 for dissolved oxygen with LSTM, DeepONet, and Informer, respectively. In contrast, the median KGE for all other variables falls below 0.45, highlighting the challenge of learning spatially transferable patterns, especially for those geochemical and nutrient-related variables.

Another major generalization challenge arises from the limited representation of extreme conditions in observational datasets. Most routine water quality samples are collected under fair-weather conditions, leading to training datasets that inadequately capture the hydrological and biogeochemical dynamics during extreme events such as floods and droughts~\cite{li2024research}. As the frequency and severity of such extremes continuously increase~\cite{calvin2023ipcc}, models trained on such incomplete data often underestimate concentration spikes or fail to reproduce nonlinear responses during events. Addressing this limitation is critical for developing trustworthy and climate-resilient prediction systems capable of operating under both typical and extreme conditions.

Knowledge-Guided Machine Learning (KGML)~\cite{karpatne2017theory} and large-scale pretrained foundation models offer promising solutions to both generalization challenges. KGML incorporates domain knowledge, such as mass conservation and transport dynamics, directly into the model’s architectures or loss functions. By embedding physical constraints, KGML supplements limited observations with process-based information, reducing the risk of overfitting and ensuring physically consistent behavior in both ungauged basins and under unobserved conditions. For example, Agrawal et al.~\cite{agrawal2025improving} integrated a physical ``flow-gate'' mechanism into an LSTM model to explicitly model hysteresis between discharge and solute dynamics, which improved predictions of nine stream solutes (RMSE reduced by 1-32\% compared with standard LSTM). Similarly, hybrid models combining process-based simulations with physics-informed objectives have enhanced generalizability and scientific consistency of results in lake temperature modeling~\cite{daw2022physics}. While these studies illustrate the potential of KGML, its application in large-scale water quality prediction remains limited, compared to its broader use in hydrology and other scientific fields~\cite{bhasme2022enhancing,daneker2024transfer,wang2021learning}.

Pretrained foundation models provide a data-driven complement to KGML by leveraging information from vast datasets. These models can be pretrained not only on cross-domain environmental data (e.g., climate reanalysis, Earth system simulations, and remote sensing), but also on synthetic datasets purposely designed to simulate rare or extreme events. Exposure to such heterogeneous datasets allows the models to learn the dynamics of both normal and extreme conditions. Once pretrained, they can be adapted to unseen tasks via few-shot (minimal samples) or zero-shot (no samples) learning~\cite{brown2020language}. For example, a model pretrained on extensive, diverse datasets, even those unrelated specifically to water quality, can effectively generalize to predict water quality variables in data-scarce basins using limited local measurements~\cite{liuself}. By combining cross-domain pretraining with efficient adaptation, foundation models have the potential to improve both the spatial generalization and generalization to climate extremes.

\section*{Reproducibility}
\vspace{-0.3cm}
Reproducibility in AI research is essential for verifying findings~\cite{li2023trustworthy}, and it has increasingly become a requirement for publication within AI communities~\cite{gundersen2018reproducible}. While AI models show significant promise for water quality research, progress in this domain is hindered by limited openness and transparency. A relatively small number of studies in this field provide full public access to their original data, models, and code, creating barriers to reproducibility, benchmarking, and collaboration. The commitment to reproducibility is more than just a verification of research~\cite{li2023trustworthy}; it is a critical step to building trust in adopting AI tools in this domain.

\section*{Limitations}
\vspace{-0.3cm}
Although this work provides a comprehensive evaluation of deep learning trustworthiness for continental-scale water quality prediction, it is important to acknowledge that our analysis focused on basins with more than 200 observations for at least one water quality variable. This threshold was selected to ensure that each basin had sufficient temporal coverage to capture seasonal and interannual variability, approximately equivalent to weekly sampling over four years, while maintaining broad spatial representation across the United States. While this approach provides a balanced compromise between data quality and availability, it may also bias the findings toward data-rich basins. Therefore, the analyses and conclusions drawn across all trustworthiness dimensions should be interpreted with caution when extrapolated to more sparsely monitored regions, where higher measurement uncertainty and irregular sampling may lead to distinct model behaviors that require further investigation in future work.

\section*{Conclusion}
This study presents a multi-dimensional, quantitative assessment of deep learning trustworthiness for continental-scale water quality prediction. By benchmarking three representative architectures: recurrent (LSTM), operator-learning (DeepONet), and transformer-based (Informer), across six dimensions of trustworthiness, we reveal key challenges that limit the reliable and responsible application of AI in water quality management.

Our results show that model performance disparities highly correlate with inherent predictability and data coverage. Variables driven by strong hydrological and seasonal patterns are well predicted, whereas nutrient-related variables remain challenging due to their event- and source-driven dynamics and sparse observations. A trade-off exists between predictive accuracy and robustness, with LSTM achieving the highest baseline performance but showing the greatest vulnerability to data corruption, while DeepONet maintains greater stability. Enhancing robustness through improved data quality control and robustness-oriented learning will be critical for real-world deployment. Predictive uncertainty is highest for management-critical variables such as nutrients and sediments, underscoring the need for explicit uncertainty quantification, improved monitoring, and uncertainty-aware training. Feature importance results are inconsistent across models and methods, highlighting the need for multi-method and multi-scale frameworks to identify reliable drivers and guide transparent, science-based decisions. Generalization remains a central challenge: current models perform poorly across basins and under extreme hydrological conditions. Integrating physical knowledge through Knowledge-Guided Machine Learning (KGML) and leveraging cross-domain pretraining in foundation models offer complementary pathways to enhance both spatial and extreme-event generalization. Finally, ensuring reproducibility through open access to data, models, and code is essential for transparency, verification, and sustained community progress.

These findings highlight that advancing deep learning for water quality prediction requires more than simply improving predictive accuracy. As AI becomes increasingly integrated into operational decision-making, it is essential to incorporate trustworthiness principles into every stage of model development. Building trust among practitioners and decision-makers is critical to ensuring that AI-driven insights are socially and environmentally responsible.
\section*{Methods}
\vspace{-0.3cm}
\subsection*{Water quality data and basin selection}
\vspace{-0.3cm}
In this work, we study 20 water quality variables regularly measured by the U.S. Geological Survey (USGS). These variables are extracted from the USGS National Water Information System (NWIS) database~\cite{USGS_WaterData} and represent various aspects of stream water quality dynamics, including physical and chemical processes, geochemical weathering, and nutrient cycling. Variables related to stream physical/chemical processes include temperature (Temp, °C), dissolved oxygen (DO, mg/L), pH, total dissolved $\text{CO}_2$ (mg/L), and total suspended sediment concentration (TSS, mg/L). Variables associated with geochemical weathering include conductivity (Cond, uS/cm at 25°C), dissolved silica ($\text{SiO}_2$, mg/L), calcium ($\text{Ca}^{2+}$, mg/L), sodium ($\text{Na}^{+}$, mg/L), potassium ($\text{K}^{+}$, mg/L), magnesium ($\text{Mg}^{2+}$, mg/L), sulfate ($\text{SO}_{4}^{2-}$, mg/L), and chloride ($\text{Cl}^{-}$, mg/L). Variables related to nutrient cycling include total nitrogen (TN, mg/L), organic nitrogen (OrgN, mg/L as N), nitrate ($\text{NO}_3^-$, mg/L as N), ammonia and ammonium ($\text{NH}_{\text{x}}$, mg/L as  $\text{NH}_{\text{4}}^+$), total phosphorus (TP, mg/L as P), orthophosphate ($\text{PO}_{4}^{3-}$, mg/L as $\text{PO}_{4}^{3-}$), and non-particulate organic carbon (NPOC, mg/L). 
Water quality data are from samples collected on a daily basis over a 37-year period, from January 1, 1982, to December 31, 2018, across 482 basins in the continental United States (CONUS)~(Fig.~\ref{fig:1}B). These basins were selected based on relatively complete water quality records using a sequential screening process as follows: (1) Basins are included in the Geospatial Attributes of Gages for Evaluating Streamflow version II (GAGES-II)~\cite{falcone2011gages}, a comprehensive dataset maintained by the USGS that provides geospatial data and classifications for over 9,000 stream gages, including basin boundaries. (2) Basins with records in which at least one water quality variable was measured for more than 200 days were retained, while basins with records not meeting this criterion were discarded. (3) We further excluded basins that measured only water temperature and specific conductance. Following this selection process, 482 basins remained for model training and evaluation. Table~\ref{tab:wq_statistics} summarizes the statistics of the selected water quality variables and the average number of observations per site over the study period. 

\subsection*{Model input}
\vspace{-0.3cm}
In addition to the target water quality variables, input features include both time-series forcings and static basin attributes. The time-series forcings are categorized into four groups: runoff, meteorological variables, vegetation indices, and rainfall chemistry. Runoff was derived as streamflow measured by the USGS divided by the basin area. Meteorological variables, including precipitation, maximum and minimum temperature, solar radiation, specific humidity, and reference evapotranspiration (grass and alfalfa, calculated using the ASCE Penman-Montieth method), were from the gridMET dataset~\cite{abatzoglou2013development}. These data were spatially aggregated for each basin using basin boundaries from the GAGES-II database. Vegetation indices, including leaf area index (LAI), net primary production (NPP), and fraction of absorbed photosynthetically active radiation (FAPAR), were obtained from the Global Land Surface Satellite (GLASS) dataset~\cite{liang2013long}. The GLASS dataset provides 8-day estimates with a spatial resolution of 0.05°. To align with the daily modeling time step, these data were interpolated to daily values using cubic splines and spatially aggregated by basin boundaries. Rainfall chemistry data were extracted from the National Atmospheric Deposition Program/National Trends Network (NADP/NTN)~\cite{NADP_2005}, which reports weekly measurements of sulfate, nitrate, chloride, ammonium, potassium, sodium, calcium, and magnesium, pH, and specific conductivity. To construct a daily time series, weekly concentrations at each NTN station were assumed to be constant over each week. Rainfall chemistry for each basin was assigned using the nearest NTN station, with the distance between the basin center and the corresponding NTN station included as an additional input feature. To capture temporal and cyclical patterns in the data, we also incorporated three time-related variables: datenum (T), the sine of the time variable (sinT), and the cosine of the time variable (cosT). The datenum (T) represents the number of days relative to January 1, 2000, with negative values for dates before this reference point and positive values for dates after. 

Based on domain knowledge and insights from previous modeling studies~\cite{zhi2023widespread,zhi2024increasing,fang2024modeling}, we identified 49 static basin attributes from the GAGES-II database as additional features. These static basin attributes encompass a wide range of phenomena and basin characteristics including topographic characteristics, the average percentage of total precipitation occurring as snow in the basin, stream hydrologic characteristics, dam information, land cover percentages, soil properties, geological features, nutrient application rates (nitrogen and phosphorus) from agriculture in the basin, and ecological classifications, as detailed in Table~\ref{tab:inputs1}.

 The selected 482 basins span multiple geographic regions and exhibit a wide range of hydrologic characteristics, hydroclimatic conditions, and land use patterns, reflecting the broad geographical diversity and regional representativeness of the water systems included in the study. These basins include 126 headwater basins (26\%) with 1st to 3rd stream orders, 280 medium-sized basins (58\%) with 4th to 6th stream orders, and 76 larger basins (16\%) with the 7th stream order or higher. The mean (median) drainage areas are 89.03 (108.54) $\text{km}^2$ for headwater basins, 3,224.07 (5,474.93) $\text{km}^2$ for medium basins, and 20,520.4 (13,062.6) $\text{km}^2$ for larger basins. Hydroclimatic conditions vary significantly across the basins. Mean annual precipitation ranges from 213.5 to 2,748.4 mm/year, with an overall mean (median) of 976.8 (985.9) mm/year. Mean annual temperatures range from -1.3 to 22.9°C, with a mean (median) of 10.5 (10.1) °C. Mean annual runoff values range from 1.6 to 2,181.5 mm/year, with a mean (median) of 348.6 (318.2) mm/year. In addition to hydrologic and climatic variability, the basins exhibit diverse land use patterns. According to classification criteria established by the USGS~\cite{spahr2010nitrate}, agricultural basins (AG) were defined as those with more than 50\% agricultural land (PLANTNLCD06 in the GAGES-II database) and less than or equal to 5\% urban land (DEVNLCD06 in the GAGES-II database). Undeveloped basins (UD) were identified as having less than or equal to 5\% urban land and less than or equal to 25\% agricultural land. Urban basins (UR) were classified as those with more than 25\% urban land and less than or equal to 25\% agricultural land, while mixed basins (MX) included all other combinations of urban, agricultural, and undeveloped land. Among the selected basins, 3.1\% were classified as AG, 11.2\% as UR, 35.1\% as UD, and 50.6\% as MX, respectively (Fig.~\ref{fig:s1}A). However, to provide a more balanced representation of basin types for subsequent analysis while maintaining the overall classification logic, we relaxed the AG definition slightly by raising the allowable urban land threshold from $\leq$5\% to $\leq$7\%. With this adjustment, the basin distribution becomes 10.2\% AG, 11.2\% UR, 35.1\% UD, and 43.6\% MX (Fig.~\ref{fig:s1}B).

\subsection*{Multi-task deep learning models training and evaluation}
\vspace{-0.3cm}
To comprehensively evaluate the trustworthiness of deep learning for water quality prediction, we examined three different model paradigms: recurrent-based (LSTM), operator-based (DeepONet), and attention-based (Informer). Model architectures and hyperparameters are described below, and schematic overviews are provided in Fig.~\ref{fig:s2}.

\textbf{LSTM.} The Long Short-Term Memory (LSTM) model is a prominent member of Recurrent Neural Network (RNN) models designed to leverage sequential information for time series prediction~\cite{hochreiter1997long}. Unlike standard RNNs, which suffer from the vanishing gradient problem when capturing long-term dependencies~\cite{noh2021analysis}. LSTM incorporates a memory mechanism to address this limitation. This mechanism, involving “memory states” and “gates”, allows the model to regulate what information to retain or discard over time, enabling more effective learning of temporal patterns. In this work, we implemented a two-layer LSTM with 512 hidden units and a dropout rate of 0.3. The input sequence length was set to 365 days to capture seasonal and annual cycles~\cite{fang2024modeling}. Training was conducted using the AdamW optimizer with an initial learning rate of 0.001 and a decay rate of 0.5 applied every 100 epochs.

\textbf{DeepONet.} Deep Operator Networks (DeepONet)~\cite{lu2021learning,mao2023ppdonet} learn mappings between input functions and output functions. In this study, the input functions are spatiotemporal forcings combined with static basin attributes, and the output functions are stream water quality dynamics. Our implementation comprises a branch network that encodes dynamic forcings over a temporal window together with static basin attributes, and a trunk network that encodes spatial coordinates (longitude and latitude) into basis functions. Outputs of the two networks are fused via element-wise multiplication and passed to a D network for the final water quality prediction. Each branch and trunk uses seven MLP-BatchNorm-LeakyReLU (LReLU) blocks with hidden size 1024; the D network uses two MLP-BN-LReLU blocks with hidden sizes 512 and 256, followed by a final MLP. Training follows the same protocol as LSTM.

\textbf{Informer.} Informer~\cite{zhou2021informer} is a transformer-based time series forecasting model specifically designed to handle long sequences efficiently. It extends the standard Transformer architecture~\cite{vaswani2017attention} by introducing mechanisms that capture both long-range dependencies and local spatiotemporal patterns. In this work, we implemented Informer in a rolling, single-step setup to ensure a fair comparison with the other two models. In addition, we replaced the probability sparse attention with the full attention mechanism. The encoder takes 365-day historical dynamic forcings together with static attributes, processed through three attention blocks, two convolutional layers, and a layer normalization (LayerNorm) layer. The decoder takes two inputs: (1) recent dynamic and static forcings from a temporal window of 96 steps, and (2) dynamic and static forcings at the prediction time step. These inputs are concatenated with the encoder output and refined through two Attention Blocks and a LayerNorm layer. A final MLP produces the predictions of water quality variables at the prediction time step. The model was trained with 4 attention heads, hidden dimension 512, feed-forward dimension 2048, GeLU activation, and a cosine annealing learning rate schedule starting at 0.0001 with a minimum of $1\times10^{-6}$.

All three models were trained to simultaneously predict 20 water quality variables, enabling shared learning of inter-variable dependencies in the complex biogeochemical processes and improving computational efficiency~\cite{fang2024modeling} compared to training separate single-task models~\cite{zhi2023widespread,zhi2024increasing,yao2024interpretable,fang2024modeling}. Experiments were conducted in PyTorch on NVIDIA RTX 3090 GPUs, with a consistent training protocol of 300 epochs, minibatch size 512, and mean squared error (MSE) loss between predictions and normalized ground truths for optimization.

\textbf{Evaluation strategy and data normalization.} The model evaluation followed a robust temporal held-out strategy, ensuring the statistical representativeness of the training data while accounting for climate variability~\cite{mcgovern2022we}. Following the approach of \cite{fang2024modeling}, data from four out of every five years were used for training, with the remaining year in each five-year period used for testing. Specifically, observations from the years 1985, 1990, 1995, 2000, 2005, 2010, and 2015 were systematically withheld during training and used exclusively for testing.

To prepare the input data for the three models, we applied different normalization techniques based on the distribution of each variable (see details in Table~\ref{tab:wq_statistics} and Table~\ref{tab:inputs1}). In brief, for vegetation indices, water quality station coordinates, time-related variables, and the target water quality variables temperature, DO and pH, we used min-max normalization. This method effectively scales variables within a range of 0 to 1, preserving the relative differences between values. For other input features and target water quality variables, we employed a log-min-max normalization approach to handle skewed distributions. Normalization parameters (minimum and maximum values) were calculated exclusively from the training data. Then, the same parameters were applied to normalize the test data to avoid information leakage.

\textbf{Performance metric.} The Kling-Gupta Efficiency (KGE, Eq.~(\ref{eq:kge}))~\cite{gupta2009decomposition} was selected as the primary metric to evaluate model performance for each of the 482 basins.  KGE is widely used in hydrological modeling studies~\cite{fang2024modeling,hunt2022using,he2022prediction}. It ranges from $-\infty$ to 1, where a value of 1 indicates perfect agreement, and values below -0.41 denote poor performance, where predictions are worse than the mean of observations~\cite{knoben2019inherent}. KGE is mathematically defined as: 
\begin{equation}\label{eq:kge}
    \text{KGE}=1-\sqrt{(r-1)^2+(\beta-1)^2+(\gamma-1)^2},
\end{equation}
where $r$ represents the correlation coefficient between observations ($O$) and model predictions ($P$); $\beta=\mu_P/\mu_O$ is the bias ratio, defined as the ratio of the mean of predictions ($\mu_P$) to the mean of observations ($\mu_O$); and $\gamma=\sigma_P/\sigma_O$ is the variability ratio, defined as the ratio of the standard deviation of predictions ($\sigma_P$) to the standard deviation of observations ($\sigma_O$). Note that metrics were computed using the original data, with inverse normalization applied to model outputs. 

\section*{Trustworthiness evaluation framework}

\section*{Robustness}

\textbf{Outliers simulation.} Extreme high and low values in water quality datasets are not uncommon due to a combination of factors, including sensor malfunctions, instrument detection limits, episodic pollution events (e.g., storm-driven contaminant pulses), and anthropogenic influences such as industrial discharges or agricultural runoff. Additionally, natural processes like sediment resuspension, extreme weather conditions, and seasonal fluctuations can drive sudden shifts in the concentration of water quality variables, further contributing to the presence of outliers in observational datasets. To evaluate the model’s robustness against such anomalies, we introduced artificial outliers into the training data by modifying 10\%, 20\%, and 30\% of the training samples. These proportions were chosen to strike a balance between realism and analytical rigor: it represents scenarios where outliers could meaningfully impact model predictions while preserving the dataset’s overall structure and distribution. We created outliers by shifting raw values to the upper or lower extremes of their distributions. These perturbations were applied to input features and target water quality variables, respectively. 

\textbf{Random measurement noise simulation.} In environmental monitoring, measurement errors are inevitable due to sensor inaccuracies, environmental variability, and sample collection inconsistencies. To simulate these conditions, we introduced random perturbations into the training data, applied to 30\%, 40\%, and 50\% of the dataset. For input features, additive deviations were introduced to mimic common sources of error. For target variables, proportional random modifications were applied to reflect discrepancies in observed water quality values, which can occur due to sampling inconsistencies, laboratory measurement precision limits, or data logging errors. 

\textbf{Adversarial inputs generation.} Adversarial vulnerabilities are a well-documented issue in AI models~\cite{kurakin2018adversarial,finlayson2019adversarial} and can have significant consequences in environmental modeling as well~\cite{mcgovern2022we}. In water quality applications, adversarially perturbed inputs could arise from systematic errors in sensor readings, cyber-physical security threats in IoT-based monitoring networks, or targeted manipulation of data used in regulatory decision-making~\cite{wu2020robust}. Unlike random noise or outliers, which are typically random or extreme deviations from the data distribution, adversarial perturbations are carefully crafted to exploit specific vulnerabilities in the model, often targeting the decisions that the model has learned. We generated adversarial inputs using the Projected Gradient Descent (PGD) method~\cite{madry2017towards}. Perturbations were applied to 10\%, 20\%, and 30\% of the dataset, targeting input features only, with an attack budget of 0.1 and a step size of one-quarter of the attack budget per iteration. 

To quantify the impact of each type of data corruption, we computed the average KGE across all scenarios and evaluated performance degradation by calculating the percentage difference relative to the baseline performance. To assess the sensitivity of models to increasing data corruption, for each corruption level, we calculated the median percent change in KGE across all station-variable pairs relative to the baseline and quantified the trend in degradation using a Theil-Sen slope estimator~\cite{theil1950rank,sen1968estimates}, expressed as the percent change in KGE per 0.1 increase of data corruption.

\section*{Uncertainty} 

We quantified the uncertainty in water quality predictions using two complementary methods, distinguishing the effects of noisy inputs from uncertainty in model parameters. Test-time augmentation (TTA)~\cite{wang2019aleatoric} introduces variability during inference by applying multiple perturbations to the test inputs, therefore estimating aleatoric uncertainty (data-driven variability). In our case, Gaussian noise with a standard deviation of 0.1, was added to the runoff input, corresponding to the typical around 10\% measurement error in streamflow~\cite{harmel2006cumulative}. Since TTA does not rely on model architecture, it was applied to all three models.

In contrast, Monte Carlo (MC) dropout~\cite{gal2016dropout} captures epistemic uncertainty (model-driven variability) by sampling different subnetworks during inference. A dropout probability of 0.3 was applied during testing. Because dropout layers were only implemented in the LSTM, MC dropout was applied exclusively to this model. For both methods, the process was repeated 50 times, and prediction uncertainty was quantified as the standard deviation (SD) of KGE across the ensemble predictions.

\section*{Interpretability} 

We evaluate the contribution of five input categories: meteorological forcing (M), runoff (Q), rainfall chemistry (RC), vegetation indices (V), and static basin attributes (BA), using two \emph{performance-based} attribution methods (ablation and traverse analysis) and one \emph{explanatory} approach (Integrated Gradients, IG). 

\textbf{Ablation.} Ablation isolates the incremental value of an input group by measuring the loss in performance when that group is removed from the full feature set. Here, the importance of each feature group is quantified by the percent reduction in model performance (e.g., KGE) when that group is removed from the full model. This approach is widely used to in hydrology and water quality modeling~\cite{zhi2023widespread,li2021development,dvorett2016mapping}. 

\textbf{Traverse analysis.} To account for interactions among groups, we evaluate all $2^5=32$ combinations of the five groups. Spatiotemporal covariates (latitude, longitude, datenum, sinT, cosT) were consistently included across all experiments to provide spatial/seasonal context and to satisfy architectural constraints (e.g., nonempty branch/trunk for DeepONet). For each group, we averaged the percent KGE difference between subsets that include a given group and those that exclude it. This uniform averaging over all contexts approximates a Shapley-style marginal contribution by accounting for interactions among groups~\cite{shapley1953value,lundberg2017unified}. As with Ablation, models were retrained for each subset using the same training protocol.

\textbf{Integrated Gradients (IG).} IG attributes a trained model’s prediction to its inputs by integrating the gradient of the output along a straight-line path from a baseline to the observed input~\cite{sundararajan2017axiomatic}. It explains which inputs the model relies on locally, but it does not estimate how model performance would change if a group were removed. Therefore, IG complements the performance-based analyses. To facilitate interpretation and maintain consistency with previous two methods, we first computed individual feature importance as the mean absolute IG over test samples and then aggregated to groups by averaging within each group.

\section*{Generalizability} 

To test the model's generalizability, we employed a spatial held-out strategy in which, within each land-use type of basins, 80\% were randomly selected for training and the remaining 20\% for testing. This ensures that the test basins are completely unseen during training, and training and testing sets include similar land-use distributions. For a fair comparison, the same training and testing basins were used across all three models. 

In summary, the training set consists of data from 385 basins, while the test set covers 97 basins. Training/testing basins contain 98/28 headwater basins (26\%/29\%) with 1st to 3rd stream orders, 232/48 medium-size basins (60\%/49\%) with 4th to 6th stream orders, and 55/21 larger basins (14\%/22\%) with the 7th stream order and higher. The mean~(median) drainage areas of the training/testing set are 94.50~(53.70)/69.91~(57.30) $\text{km}^2$ for headwater basins, 3171.89~(1228.40)/3476.28~(1692.50) $\text{km}^2$ for medium basins, and 20640.54~(17944.10)/20205.72~(15724.90) $\text{km}^2$ for larger basins. For hydroclimatic conditions, the mean annual precipitation of training/testing basins ranges from 213.5 to 2647.3/301.6 to 2748.2 mm/year, with an overall mean~(median) of 966.7~(983.7)/1017.0~(1059.0) mm/year. Mean annual temperatures of training and testing basins range from 0.53 to 22.9/-1.3 to 22.7 °C, with a mean (median) of 10.4~(10.0)/10.9~(10.3) °C. Mean annual runoff for training and testing sets ranges from 0.8 to 796.2/0.6 to 773.4 mm/year, with a mean (median) of 124.2~(118.7)/139.2~(109.4) mm/year. Furthermore, land-use distributions of the training and testing sets were comparable due to the stratified held-out procedure: 3.1\% agricultural (AG), 11.2\% urban (UR), 35.1\% undeveloped (UD), and 50.6\% mixed (MX).

\section*{Data sources}
Streamflow and water quality data were extracted from the U.S. Geological Survey (USGS) National Water Information System (NWIS) database (\url{https://waterdata.usgs.gov/nwis}). Meteorological variables were extracted from the gridMET dataset (\url{https://www.climatologylab.org/gridmet.html}). Vegetation indices were acquired from the Global Land Surface Satellite (GLASS) dataset (\url{http://www.glass.umd.edu/Download.html}). Rainfall chemistry data were retrieved from the National Atmospheric Deposition Program/National Trends Network (NADP/NTN) (\url{https://nadp.slh.wisc.edu/networks/national-trends-network/}). Basin attributes were obtained from the Geospatial Attributes of Gages for Evaluating Streamflow, version II (GAGES-II) database (\url{https://www.sciencebase.gov/catalog/item/631405bbd34e36012efa304a}). Processed data for the 482 basins used in this study are publicly available at \url{https://figshare.com/s/e0151c12b6e6482bae83}.

\section*{Code and data availability}
Python scripts for downloading water quality data are available at: \url{https://github.com/fkwai/geolearn/tree/master/hydroDL/data}. The codes for the three deep learning models and their trustworthiness evaluation developed in this study are available at: \url{https://github.com/xiaoboxia/TrustEval-DeepWQ}. Python codes for statistical analysis and visualization are available from the authors upon request.

\section*{Acknowledgements}
The authors sincerely thank the editor and anonymous reviewers for their constructive and insightful comments, which greatly improved the quality of this manuscript. X.X. was supported by MoE Key Laboratory of Brain-inspired Intelligent Perception and Cognition, University of Science and Technology of China (Grant No. 2421002). X.L. was supported by Schmidt Sciences. T.L. was partially supported by the following Australian Research Council projects: FT220100318, DP220102121, LP220100527, LP220200949, and IC190100031. 

\section*{Author contributions}
Research conceptualization: X.X., X.L., and T.L.; experiment design: X.X. and X.L.; experiment implementation: X.X. and J.L.; data preprocessing: K.F., X.X., and X.L.; data analysis: X.X. and X.L.; results visualization and interpretation: X.L. and X.X.; writing - original draft: X.X. and X.L.; writing - review and editing: X.X., X.L., J.L., K.F., L.L., S.O., W.S., and T.L.; supervision: T.L.

\section*{Declaration of interests}
The authors declare no competing interests.

%\bibliography{reference}

%\bibliographystyle{naturemag}
\printbibliography

@article{lu2021learning,
  title={Learning nonlinear operators via DeepONet based on the universal approximation theorem of operators},
  author={Lu, Lu and Jin, Pengzhan and Pang, Guofei and Zhang, Zhongqiang and Karniadakis, George Em},
  journal={Nature Machine Intelligence},
  volume={3},
  number={3},
  pages={218--229},
  year={2021}
}

@inproceedings{zhou2021informer,
  title={Informer: Beyond efficient transformer for long sequence time-series forecasting},
  author={Zhou, Haoyi and Zhang, Shanghang and Peng, Jieqi and Zhang, Shuai and Li, Jianxin and Xiong, Hui and Zhang, Wancai},
  booktitle={Proceedings of the AAAI Conference on Artificial Intelligence},
  volume={35},
  number={12},
  pages={11106--11115},
  year={2021}
}

@article{mao2023ppdonet,
  title={Ppdonet: Deep operator networks for fast prediction of steady-state solutions in disk--planet systems},
  author={Mao, Shunyuan and Dong, Ruobing and Lu, Lu and Yi, Kwang Moo and Wang, Sifan and Perdikaris, Paris},
  journal={The Astrophysical Journal Letters},
  volume={950},
  number={2},
  pages={L12},
  year={2023}
}

@article{li2024research,
  title={Research progress in water quality prediction based on deep learning technology: a review},
  author={Li, Wenhao and Zhao, Yin and Zhu, Yining and Dong, Zhongtian and Wang, Fenghe and Huang, Fengliang},
  journal={Environmental Science and Pollution Research},
  volume={31},
  number={18},
  pages={26415--26431},
  year={2024}
}

@article{zhou2024comprehensive,
  title={A comprehensive survey on pretrained foundation models: A history from bert to chatgpt},
  author={Zhou, Ce and Li, Qian and Li, Chen and Yu, Jun and Liu, Yixin and Wang, Guangjing and Zhang, Kai and Ji, Cheng and Yan, Qiben and He, Lifang and others},
  journal={International Journal of Machine Learning and Cybernetics},
  pages={1--65},
  year={2024}
}

@article{liu2024probing,
  title={Probing the limit of hydrologic predictability with the Transformer network},
  author={Liu, Jiangtao and Bian, Yuchen and Lawson, Kathryn and Shen, Chaopeng},
  journal={Journal of Hydrology},
  volume={637},
  pages={131389},
  year={2024}
}

@incollection{daw2022physics,
  title={Physics-guided neural networks (pgnn): An application in lake temperature modeling},
  author={Daw, Arka and Karpatne, Anuj and Watkins, William D and Read, Jordan S and Kumar, Vipin},
  booktitle={Knowledge Guided Machine Learning},
  pages={353--372},
  year={2022}
}

@article{bhasme2022enhancing,
  title={Enhancing predictive skills in physically-consistent way: Physics informed machine learning for hydrological processes},
  author={Bhasme, Pravin and Vagadiya, Jenil and Bhatia, Udit},
  journal={Journal of Hydrology},
  volume={615},
  pages={128618},
  year={2022}
}

@article{daneker2024transfer,
  title={Transfer learning on physics-informed neural networks for tracking the hemodynamics in the evolving false lumen of dissected aorta},
  author={Daneker, Mitchell and Cai, Shengze and Qian, Ying and Myzelev, Eric and Kumbhat, Arsh and Li, He and Lu, Lu},
  journal={Nexus},
  volume={1},
  number={2},
  year={2024}
}

@article{wang2021learning,
  title={Learning the solution operator of parametric partial differential equations with physics-informed DeepONets},
  author={Wang, Sifan and Wang, Hanwen and Perdikaris, Paris},
  journal={Science Advances},
  volume={7},
  number={40},
  pages={eabi8605},
  year={2021}
}

@article{gupta2009decomposition,
  title={Decomposition of the mean squared error and NSE performance criteria: Implications for improving hydrological modelling},
  author={Gupta, Hoshin V and Kling, Harald and Yilmaz, Koray K and Martinez, Guillermo F},
  journal={Journal of Hydrology},
  volume={377},
  number={1-2},
  pages={80--91},
  year={2009}
}

@article{knoben2019inherent,
  title={Inherent benchmark or not? Comparing Nash--Sutcliffe and Kling--Gupta efficiency scores},
  author={Knoben, Wouter JM and Freer, Jim E and Woods, Ross A},
  journal={Hydrology and Earth System Sciences},
  volume={23},
  number={10},
  pages={4323--4331},
  year={2019}
}

@article{vaswani2017attention,
  title={Attention is all you need},
  author={Vaswani, Ashish and Shazeer, Noam and Parmar, Niki and Uszkoreit, Jakob and Jones, Llion and Gomez, Aidan N and Kaiser, {\L}ukasz and Polosukhin, Illia},
  journal={Advances in Neural Information Processing Systems},
  volume={30},
  year={2017}
}

@article{hunt2022using,
  title={Using a long short-term memory (LSTM) neural network to boost river streamflow forecasts over the western United States},
  author={Hunt, Kieran MR and Matthews, Gwyneth R and Pappenberger, Florian and Prudhomme, Christel},
  journal={Hydrology and Earth System Sciences},
  volume={26},
  number={21},
  pages={5449--5472},
  year={2022}
}

@article{fang2024modeling,
  title={Modeling continental US stream water quality using long-short term memory and weighted regressions on time, discharge, and season},
  author={Fang, Kuai and Caers, Jef and Maher, Kate},
  journal={Frontiers in Water},
  volume={6},
  pages={1456647},
  year={2024},
  publisher={Frontiers Media SA}
}

@article{he2022prediction,
  title={Prediction of total nitrogen and phosphorus in surface water by deep learning methods based on multi-scale feature extraction},
  author={He, Miao and Wu, Shaofei and Huang, Binbin and Kang, Chuanxiong and Gui, Faliang},
  journal={Water},
  volume={14},
  number={10},
  pages={1643},
  year={2022}
}

@article{mcgovern2022we,
  title={Why we need to focus on developing ethical, responsible, and trustworthy artificial intelligence approaches for environmental science},
  author={McGovern, Amy and Ebert-Uphoff, Imme and Gagne II, David John and Bostrom, Ann},
  journal={Environmental Data Science},
  volume={1},
  pages={e6},
  year={2022}
}

@article{zhi2024increasing,
  title={Increasing phosphorus loss despite widespread concentration decline in US rivers},
  author={Zhi, Wei and Baniecki, Hubert and Liu, Jiangtao and Boyer, Elizabeth and Shen, Chaopeng and Shenk, Gary and Liu, Xiaofeng and Li, Li},
  journal={Proceedings of the National Academy of Sciences},
  volume={121},
  number={48},
  pages={e2402028121},
  year={2024}
}

@article{hochreiter1997long,
  title={Long short-term memory},
  author={Hochreiter, Sepp and Schmidhuber, J{\"u}rgen},
  journal={Neural computation},
  volume={9},
  number={8},
  pages={1735--1780},
  year={1997}
}

@inproceedings{sundararajan2017axiomatic,
  title={Axiomatic attribution for deep networks},
  author={Sundararajan, Mukund and Taly, Ankur and Yan, Qiqi},
  booktitle={International Conference on Machine Learning},
  pages={3319--3328},
  year={2017},
  organization={PMLR}
}

@article{zhi2023widespread,
  title={Widespread deoxygenation in warming rivers},
  author={Zhi, Wei and Klingler, Christoph and Liu, Jiangtao and Li, Li},
  journal={Nature Climate Change},
  volume={13},
  number={10},
  pages={1105--1113},
  year={2023}
}

@article{noh2021analysis,
  title={Analysis of gradient vanishing of RNNs and performance comparison},
  author={Noh, Seol-Hyun},
  journal={Information},
  volume={12},
  number={11},
  pages={442},
  year={2021}
}

@article{yao2024interpretable,
  title={Interpretable CEEMDAN-FE-LSTM-transformer hybrid model for predicting total phosphorus concentrations in surface water},
  author={Yao, Jiefu and Chen, Shuai and Ruan, Xiaohong},
  journal={Journal of Hydrology},
  volume={629},
  pages={130609},
  year={2024}
}

@article{calvin2023ipcc,
  title={IPCC, 2023: Climate Change 2023: Synthesis Report, Summary for Policymakers. Contribution of Working Groups I, II and III to the Sixth Assessment Report of the Intergovernmental Panel on Climate Change [Core Writing Team, H. Lee and J. Romero (eds.)]. IPCC, Geneva, Switzerland.},
  author={Calvin, Katherine and Dasgupta, Dipak and Krinner, Gerhard and Mukherji, Aditi and Thorne, Peter W and Trisos, Christopher and Romero, Jos{\'e} and Aldunce, Paulina and Barret, Ko and Blanco, Gabriel and others},
  journal={IPCC, 2023: Climate Change 2023: Synthesis Report. Contribution of Working Groups I, II and III to the Sixth Assessment Report of the Intergovernmental Panel on Climate Change [Core Writing Team, H. Lee and J. Romero (eds.)]. IPCC, Geneva, Switzerland.},
  pages={1--34},
  year={2023}
}

@article{agrawal2025improving,
  title={Improving stream solute predictions with a modified LSTM model incorporating solute interdependences and hysteresis patterns},
  author={Agrawal, Tarun and Goodwell, Allison and Kumar, Praveen},
  journal={Journal of Geophysical Research: Machine Learning and Computation},
  volume={2},
  number={1},
  pages={e2024JH000383},
  year={2025}
}

@inproceedings{gal2016dropout,
  title={Dropout as a bayesian approximation: Representing model uncertainty in deep learning},
  author={Gal, Yarin and Ghahramani, Zoubin},
  booktitle={International Conference on Machine Learning},
  pages={1050--1059},
  year={2016}
}

@article{li2021development,
  title={Development of a Wilks feature importance method with improved variable rankings for supporting hydrological inference and modelling},
  author={Li, Kailong and Huang, Guohe and Baetz, Brian},
  journal={Hydrology and Earth System Sciences},
  volume={25},
  number={9},
  pages={4947--4966},
  year={2021}
}

@article{dvorett2016mapping,
  title={Mapping and hydrologic attribution of temporary wetlands using recurrent Landsat imagery},
  author={Dvorett, Daniel and Davis, Craig and Pape{\c{s}}, Monica},
  journal={Wetlands},
  volume={36},
  pages={431--443},
  year={2016}
}

@misc{USGS_WaterData,
  author = {{U.S. Geological Survey}},
  title = {{National Water Information System}},
  year = {2025},
  note = {Available at: \url{https://waterdata.usgs.gov/nwis} (Last accessed: March 5, 2025).}
}

@article{falcone2011gages,
  title={GAGES-II: Geospatial attributes of gages for evaluating streamflow},
  author={Falcone, James A},
  journal={USGS Report},
  pages={41},
  year={2011}
}

@misc{NADP_2005,
  author = {{National Atmospheric Deposition Program (NADP)}},
  year = {2005},
  title = {{National Trends Network}},
  institution = {{Illinois State Water Survey}},
  address = {Champaign, IL, U.S.A.},
  howpublished = {\url{http://nadp.slh.wisc.edu/networks/national-trends-network/}},
  note = {Last accessed: January 16, 2025}
}

@techreport{spahr2010nitrate,
  title={Nitrate loads and concentrations in surface-water base flow and shallow groundwater for selected basins in the United States, water years 1990-2006},
  author={Spahr, Norman E and Dubrovsky, Neil M and Gronberg, JoAnn M and Franke, O Lehn and Wolock, David M},
  year={2010},
  institution={US Geological Survey}
}

@article{dudgeon2006freshwater,
  title={Freshwater biodiversity: importance, threats, status and conservation challenges},
  author={Dudgeon, David and Arthington, Angela H and Gessner, Mark O and Kawabata, Zen-Ichiro and Knowler, Duncan J and L{\'e}v{\^e}que, Christian and Naiman, Robert J and Prieur-Richard, Anne-H{\'e}l{\`e}ne and Soto, Doris and Stiassny, Melanie LJ and others},
  journal={Biological Reviews},
  volume={81},
  number={2},
  pages={163--182},
  year={2006}
}

@book{mckeown2015impact,
  title={Impact of water pollution on human health and environmental sustainability},
  author={McKeown, A Elaine},
  year={2015},
  publisher={IGI Global}
}

@article{du2022persistent,
  title={Persistent degradation: Global water quality challenges and required actions},
  author={du Plessis, Anja},
  journal={One Earth},
  volume={5},
  number={2},
  pages={129--131},
  year={2022}
}

@article{keiser2019low,
  title={The low but uncertain measured benefits of US water quality policy},
  author={Keiser, David A and Kling, Catherine L and Shapiro, Joseph S},
  journal={Proceedings of the National Academy of Sciences},
  volume={116},
  number={12},
  pages={5262--5269},
  year={2019},
  publisher={National Academy of Sciences}
}

@article{babatunde2024study,
  title={A study on traditional water quality assessment methods},
  author={Babatunde, Abiodun},
  journal={Risk Assessment and Management Decisions},
  volume={1},
  number={1},
  pages={41--52},
  year={2024}
}

@article{hooker2019benchmark,
  title={A benchmark for interpretability methods in deep neural networks},
  author={Hooker, Sara and Erhan, Dumitru and Kindermans, Pieter-Jan and Kim, Been},
  journal={Advances in Neural Information Processing Systems},
  volume={32},
  year={2019}
}

@inproceedings{catav2021marginal,
  title={Marginal contribution feature importance-an axiomatic approach for explaining data},
  author={Catav, Amnon and Fu, Boyang and Zoabi, Yazeed and Meilik, Ahuva Libi Weiss and Shomron, Noam and Ernst, Jason and Sankararaman, Sriram and Gilad-Bachrach, Ran},
  booktitle={International Conference on Machine Learning},
  pages={1324--1335},
  year={2021},
}

@article{lundberg2017unified,
  title={A unified approach to interpreting model predictions},
  author={Lundberg, Scott M and Lee, Su-In},
  journal={Advances in Neural Information Processing Systems},
  volume={30},
  year={2017}
}

@article{kratzert2019toward,
  title={Toward improved predictions in ungauged basins: Exploiting the power of machine learning},
  author={Kratzert, Frederik and Klotz, Daniel and Herrnegger, Mathew and Sampson, Alden K and Hochreiter, Sepp and Nearing, Grey S},
  journal={Water Resources Research},
  volume={55},
  number={12},
  pages={11344--11354},
  year={2019}
}

@article{bai2021recent,
  title={Recent advances in adversarial training for adversarial robustness},
  author={Bai, Tao and Luo, Jinqi and Zhao, Jun and Wen, Bihan and Wang, Qian},
  journal={arXiv preprint arXiv:2102.01356},
  year={2021}
}

@article{lin2022distributionally,
  title={Distributionally robust optimization: A review on theory and applications},
  author={Lin, Fengming and Fang, Xiaolei and Gao, Zheming},
  journal={Numerical Algebra, Control and Optimization},
  volume={12},
  number={1},
  pages={159--212},
  year={2022}
}

@inproceedings{hu2021uncertainty,
  title={Uncertainty-aware reliable text classification},
  author={Hu, Yibo and Khan, Latifur},
  booktitle={Proceedings of the 27th ACM SIGKDD Conference on Knowledge Discovery \& Data Mining},
  pages={628--636},
  year={2021}
}

@article{einbinder2022training,
  title={Training uncertainty-aware classifiers with conformalized deep learning},
  author={Einbinder, Bat-Sheva and Romano, Yaniv and Sesia, Matteo and Zhou, Yanfei},
  journal={Advances in Neural Information Processing Systems},
  volume={35},
  pages={22380--22395},
  year={2022}
}

@article{liang2013long,
  title={A long-term Global LAnd Surface Satellite (GLASS) data-set for environmental studies},
  author={Liang, Shunlin and Zhao, Xiang and Liu, Suhong and Yuan, Wenping and Cheng, Xiao and Xiao, Zhiqiang and Zhang, Xiaotong and Liu, Qiang and Cheng, Jie and Tang, Hairong and others},
  journal={International Journal of Digital Earth},
  volume={6},
  number={sup1},
  pages={5--33},
  year={2013}
}

@inproceedings{li2021learning,
  title={Learning probabilistic ordinal embeddings for uncertainty-aware regression},
  author={Li, Wanhua and Huang, Xiaoke and Lu, Jiwen and Feng, Jianjiang and Zhou, Jie},
  booktitle={Proceedings of the IEEE/CVF Conference on Computer Vision and Pattern Recognition},
  pages={13896--13905},
  year={2021}
}

@article{chua2023tackling,
  title={Tackling prediction uncertainty in machine learning for healthcare},
  author={Chua, Michelle and Kim, Doyun and Choi, Jongmun and Lee, Nahyoung G and Deshpande, Vikram and Schwab, Joseph and Lev, Michael H and Gonzalez, Ramon G and Gee, Michael S and Do, Synho},
  journal={Nature Biomedical Engineering},
  volume={7},
  number={6},
  pages={711--718},
  year={2023}
}

@article{abatzoglou2013development,
  title={Development of gridded surface meteorological data for ecological applications and modelling},
  author={Abatzoglou, John T},
  journal={International Journal of Climatology},
  volume={33},
  number={1},
  pages={121--131},
  year={2013}
}

@article{madry2017towards,
  title={Towards deep learning models resistant to adversarial attacks},
  author={Madry, Aleksander and Makelov, Aleksandar and Schmidt, Ludwig and Tsipras, Dimitris and Vladu, Adrian},
  journal={arXiv preprint arXiv:1706.06083},
  year={2017}
}

@incollection{kurakin2018adversarial,
  title={Adversarial examples in the physical world},
  author={Kurakin, Alexey and Goodfellow, Ian J and Bengio, Samy},
  booktitle={Artificial Intelligence Safety and Security},
  pages={99--112},
  year={2018}
}

@article{finlayson2019adversarial,
  title={Adversarial attacks on medical machine learning},
  author={Finlayson, Samuel G and Bowers, John D and Ito, Joichi and Zittrain, Jonathan L and Beam, Andrew L and Kohane, Isaac S},
  journal={Science},
  volume={363},
  number={6433},
  pages={1287--1289},
  year={2019}
}

@article{wu2020robust,
  title={Robust learning-enabled intelligence for the internet of things: A survey from the perspectives of noisy data and adversarial examples},
  author={Wu, Yulei},
  journal={IEEE Internet of Things Journal},
  volume={8},
  number={12},
  pages={9568--9579},
  year={2020}
}

@article{li2023trustworthy,
  title={Trustworthy AI: From principles to practices},
  author={Li, Bo and Qi, Peng and Liu, Bo and Di, Shuai and Liu, Jingen and Pei, Jiquan and Yi, Jinfeng and Zhou, Bowen},
  journal={ACM Computing Surveys},
  volume={55},
  number={9},
  pages={1--46},
  year={2023}
}

@article{gundersen2018reproducible,
  title={On reproducible AI: Towards reproducible research, open science, and digital scholarship in AI publications},
  author={Gundersen, Odd Erik and Gil, Yolanda and Aha, David W},
  journal={AI Magazine},
  volume={39},
  number={3},
  pages={56--68},
  year={2018}
}

@article{brown2020language,
  title={Language models are few-shot learners},
  author={Brown, Tom and Mann, Benjamin and Ryder, Nick and Subbiah, Melanie and Kaplan, Jared D and Dhariwal, Prafulla and Neelakantan, Arvind and Shyam, Pranav and Sastry, Girish and Askell, Amanda and others},
  journal={Advances in Neural Information Processing Systems},
  volume={33},
  pages={1877--1901},
  year={2020}
}

@article{liu2021chlorophyll,
  title={Chlorophyll a estimation in lakes using multi-parameter sonde data},
  author={Liu, Xiaofeng and Georgakakos, Aris P},
  journal={Water Research},
  volume={205},
  pages={117661},
  year={2021},
  publisher={Elsevier}
}

@article{wang2019aleatoric,
  title={Aleatoric uncertainty estimation with test-time augmentation for medical image segmentation with convolutional neural networks},
  author={Wang, Guotai and Li, Wenqi and Aertsen, Michael and Deprest, Jan and Ourselin, S{\'e}bastien and Vercauteren, Tom},
  journal={Neurocomputing},
  volume={338},
  pages={34--45},
  year={2019},
  publisher={Elsevier}
}

@article{shen2018transdisciplinary,
  title={A transdisciplinary review of deep learning research and its relevance for water resources scientists},
  author={Shen, Chaopeng},
  journal={Water Resources Research},
  volume={54},
  number={11},
  pages={8558--8593},
  year={2018}
}

@article{liang2022advances,
  title={Advances, challenges and opportunities in creating data for trustworthy AI},
  author={Liang, Weixin and Tadesse, Girmaw Abebe and Ho, Daniel and Fei-Fei, Li and Zaharia, Matei and Zhang, Ce and Zou, James},
  journal={Nature Machine Intelligence},
  volume={4},
  number={8},
  pages={669--677},
  year={2022}
}

@article{eshete2021making,
  title={Making machine learning trustworthy},
  author={Eshete, Birhanu},
  journal={Science},
  volume={373},
  number={6556},
  pages={743--744},
  year={2021}
}

@article{tomsett2020rapid,
  title={Rapid trust calibration through interpretable and uncertainty-aware AI},
  author={Tomsett, Richard and Preece, Alun and Braines, Dave and Cerutti, Federico and Chakraborty, Supriyo and Srivastava, Mani and Pearson, Gavin and Kaplan, Lance},
  journal={Patterns},
  volume={1},
  number={4},
  year={2020}
}

@article{zhu2023reliable,
  title={Reliable extrapolation of deep neural operators informed by physics or sparse observations},
  author={Zhu, Min and Zhang, Handi and Jiao, Anran and Karniadakis, George Em and Lu, Lu},
  journal={Computer Methods in Applied Mechanics and Engineering},
  volume={412},
  pages={116064},
  year={2023}
}

@article{markus2021role,
  title={The role of explainability in creating trustworthy artificial intelligence for health care: a comprehensive survey of the terminology, design choices, and evaluation strategies},
  author={Markus, Aniek F and Kors, Jan A and Rijnbeek, Peter R},
  journal={Journal of Biomedical Informatics},
  volume={113},
  pages={103655},
  year={2021}
}

@article{fernandez2021trustworthy,
  title={Trustworthy autonomous vehicles},
  author={Fern{\'a}ndez Llorca, David and G{\'o}mez, Emilia},
  journal={Publications Office of the European Union, Luxembourg,, EUR},
  volume={30942},
  year={2021}
}

@article{wang2024sentiment,
  title={Sentiment analysis via trustworthy label enhancement for consumer electronics applications},
  author={Wang, Xin and Yi, Bo and Felemban, Bassem F and Aly, Ayman A and Li, Wenjuan and Liu, Jinlei},
  journal={IEEE Transactions on Consumer Electronics},
  year={2024}
}

@article{mcgovern2024value,
  title={The value of convergence research for developing trustworthy AI for weather, climate, and ocean hazards},
  author={McGovern, Amy and Demuth, Julie and Bostrom, Ann and Wirz, Christopher D and Tissot, Philippe E and Cains, Mariana G and Musgrave, Kate D},
  journal={npj Natural Hazards},
  volume={1},
  number={1},
  pages={13},
  year={2024}
}

@article{zhi2024deep,
  title={Deep learning for water quality},
  author={Zhi, Wei and Appling, Alison P and Golden, Heather E and Podgorski, Joel and Li, Li},
  journal={Nature water},
  volume={2},
  number={3},
  pages={228--241},
  year={2024}
}

@article{liuself,
  title={Self-Imputation and Cross-Variable Learning Improve Water Quality Prediction with Sparse Data},
  author={Liu, Xiaofeng and Xia, Xiaobo and Zhang, Xuechen and Chakraborty, Mohna and Chang, Xiyuan and Fang, Kuai and Currie, William S and Oymak, Samet},
  journal={1st ICML Workshop on Foundation Models for Structured Data},
  year={2025}
}

@article{theil1950rank,
  title={A rank-invariant method of linear and polynomial regression analysis},
  author={Theil, Henri},
  journal={Nederl. Akad. Wetensch., Proc.},
  volume={53},
  pages={386--392},
  year={1950}
}

@article{sen1968estimates,
  title={Estimates of the regression coefficient based on Kendall’s tau},
  author={Sen, Pranab K.},
  journal={Journal of the American Statistical Association},
  volume={63},
  number={324},
  pages={1379--1389},
  year={1968},
  publisher={Taylor \& Francis}
}

@article{harmel2006cumulative,
  title={Cumulative uncertainty in measured streamflow and water quality data for small watersheds},
  author={Harmel, RD and Cooper, RJ and Slade, RM and Haney, RL and Arnold, JG},
  journal={Transactions of the ASABE},
  volume={49},
  number={3},
  pages={689--701},
  year={2006},
  publisher={American Society of Agricultural and Biological Engineers}
}

@article{shapley1953value,
  title={A value for n-person games},
  author={Shapley, Lloyd S and others},
  year={1953},
  publisher={Princeton University Press Princeton}
}

@article{karpatne2017theory,
  title={Theory-guided data science: A new paradigm for scientific discovery from data},
  author={Karpatne, Anuj and Atluri, Gowtham and Faghmous, James H and Steinbach, Michael and Banerjee, Arindam and Ganguly, Auroop and Shekhar, Shashi and Samatova, Nagiza and Kumar, Vipin},
  journal={IEEE Transactions on knowledge and data engineering},
  volume={29},
  number={10},
  pages={2318--2331},
  year={2017},
  publisher={IEEE}
}

@article{mcgraw1992common,
  title={A common language effect size statistic.},
  author={McGraw, Kenneth O and Wong, Seok P},
  journal={Psychological bulletin},
  volume={111},
  number={2},
  pages={361},
  year={1992},
  publisher={American Psychological Association}
}

@article{liu2025rnns,
  title={From RNNs to Transformers: benchmarking deep learning architectures for hydrologic prediction},
  author={Liu, Jiangtao and Shen, Chaopeng and O'Donncha, Fearghal and Song, Yalan and Zhi, Wei and Beck, Hylke E and Bindas, Tadd and Kraabel, Nicholas and Lawson, Kathryn},
  journal={EGUsphere},
  volume={2025},
  pages={1--21},
  year={2025},
  publisher={Copernicus Publications G{\"o}ttingen, Germany}
}

@article{sun2024bridging,
  title={Bridging hydrological ensemble simulation and learning using deep neural operators},
  author={Sun, Alexander Y and Jiang, Peishi and Shuai, Pin and Chen, Xingyuan},
  journal={Water Resources Research},
  volume={60},
  number={10},
  pages={e2024WR037555},
  year={2024},
  publisher={Wiley Online Library}
}

@article{frame2022deep,
  title={Deep learning rainfall--runoff predictions of extreme events},
  author={Frame, Jonathan M and Kratzert, Frederik and Klotz, Daniel and Gauch, Martin and Shalev, Guy and Gilon, Oren and Qualls, Logan M and Gupta, Hoshin V and Nearing, Grey S},
  journal={Hydrology and Earth System Sciences},
  volume={26},
  number={13},
  pages={3377--3392},
  year={2022},
  publisher={Copernicus Publications G{\"o}ttingen, Germany}
}

@inproceedings{wang2024self,
  title={Self-adaptive extreme penalized loss for imbalanced time series prediction},
  author={Wang, Yiyang and Han, Yuchen and Guo, Yuhan},
  booktitle={Proceedings of the Thirty-Third International Joint Conference on Artificial Intelligence},
  pages={5135--5143},
  year={2024}
}

@inproceedings{ding2019modeling,
  title={Modeling extreme events in time series prediction},
  author={Ding, Daizong and Zhang, Mi and Pan, Xudong and Yang, Min and He, Xiangnan},
  booktitle={Proceedings of the 25th ACM SIGKDD international conference on knowledge discovery \& data mining},
  pages={1114--1122},
  year={2019}
}

@article{radfar2025integrating,
  title={Integrating Newton's Laws with deep learning for enhanced physics-informed compound flood modelling},
  author={Radfar, Soheil and Maghsoodifar, Faezeh and Moftakhari, Hamed and Moradkhani, Hamid},
  journal={arXiv preprint arXiv:2507.15021},
  year={2025}
}

@article{song2025high,
  title={High-resolution national-scale water modeling is enhanced by multiscale differentiable physics-informed machine learning},
  author={Song, Yalan and Bindas, Tadd and Shen, Chaopeng and Ji, Haoyu and Knoben, Wouter JM and Lonzarich, Leo and Clark, Martyn P and Liu, Jiangtao and van Werkhoven, Katie and Lamont, Sam and others},
  journal={Water Resources Research},
  volume={61},
  number={4},
  pages={e2024WR038928},
  year={2025},
  publisher={Wiley Online Library}
}
\newpage
\appendix
\appendix
\renewcommand{\thetable}{S\arabic{table}} 
\renewcommand{\thefigure}{S\arabic{figure}} 
\setcounter{table}{0}
\setcounter{figure}{0}

%\section*{Supplementary Information}
\begin{center}
    \textbf{\Large Table of content}
\end{center}

\begin{figure}[!t]
    \centering
    \includegraphics[width=0.6\linewidth]{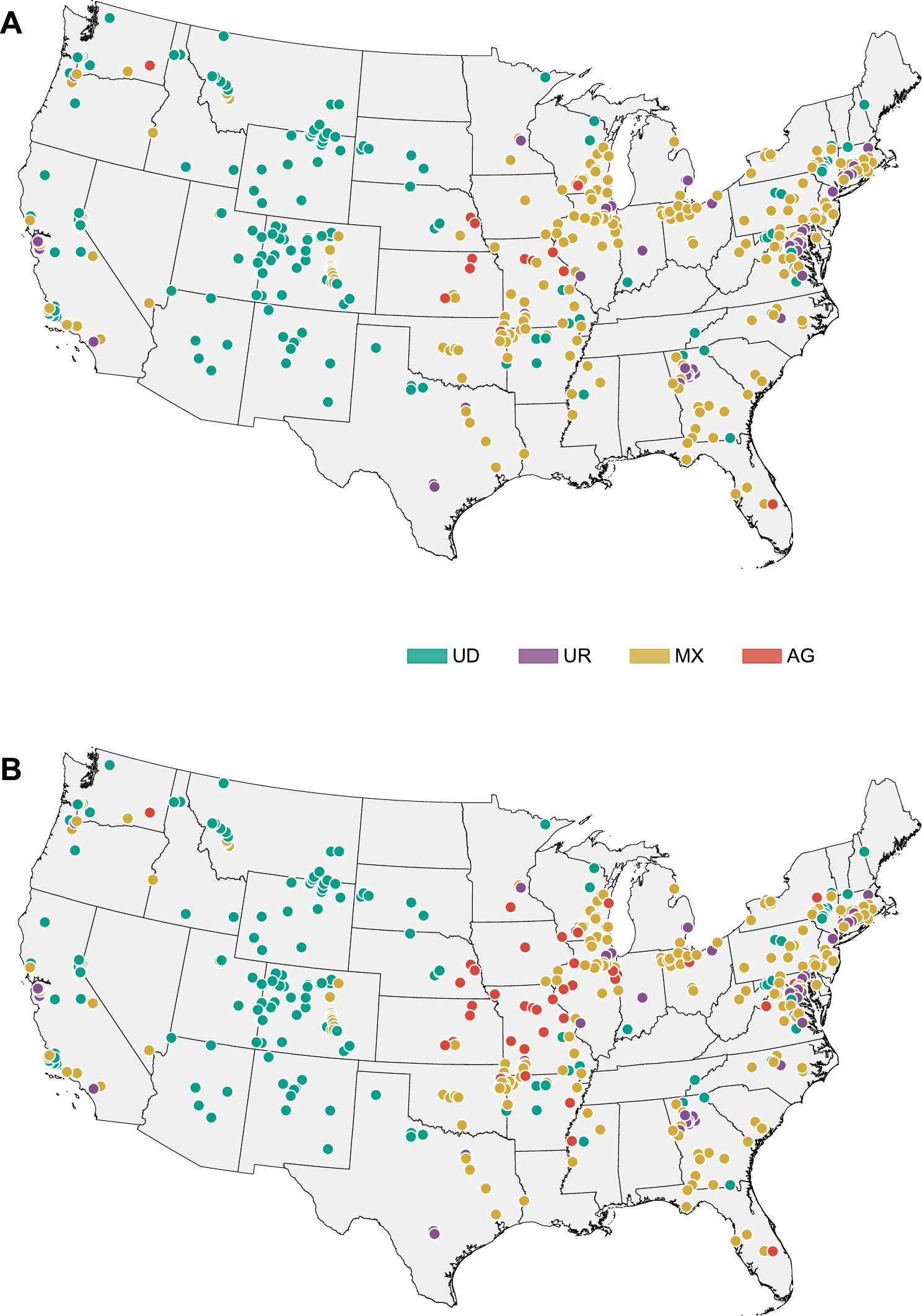}
    \caption{\setlength{\baselineskip}{1.5\baselineskip}Spatial distribution of studied basins classified by land uses. (A) Basin types following the USGS classification criteria~\cite{spahr2010nitrate}, agricultural basins (AG, red) are defined as having more than 50\% agricultural land (PLANTNLCD06 in the GAGES‐II database) and at most 5\% urban land (DEVNLCD06). Undeveloped basins (UD, green) have at most 5\% urban land and at most 25\% agricultural land. Urban basins (UR, purple) are defined as having more than 25\% urban land and at most 25\% agricultural land, while mixed basins (MX, yellow) include all other combinations of urban, agricultural, and undeveloped land. Based on these thresholds, 3.1\% were classified as AG, 11.2\% as UR, 35.1\% as UD, and 50.6\% as MX. (B) To provide a more balanced representation in the subsequent analysis while maintaining classification logic, the AG definition was relaxed to allow up to 7\% urban land. Under this adjustment, the distribution shifted to 10.2\% AG, 11.2\% UR, 35.1\% UD, and 43.6\% MX.}
    \label{fig:s1}
\end{figure}

\begin{figure}[!t]
    \centering
   % \vspace{cm}
    \includegraphics[width=0.8\linewidth]{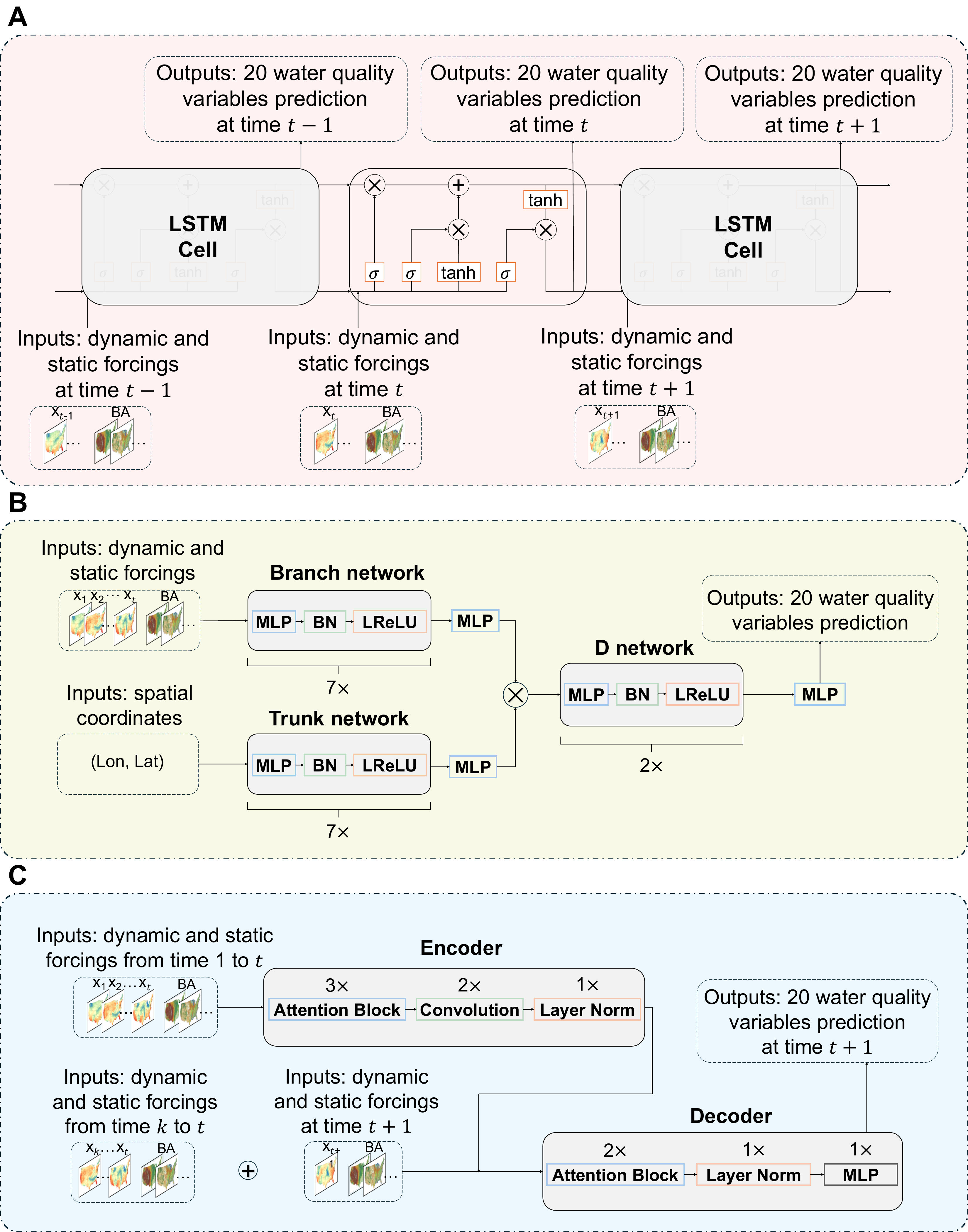}
    \caption{\setlength{\baselineskip}{1.5\baselineskip}\textcolor{black}{Schematic overview of the multi-task LSTM model (A), DeepONet (B), and Informer (C) to predict 20 water quality variables simultaneously by leveraging time-series hydroclimate forcings and static basin attributes as inputs. }}
    \label{fig:s2}
\end{figure}

\begin{figure}[!t]
    \centering
    \includegraphics[width=1\linewidth]{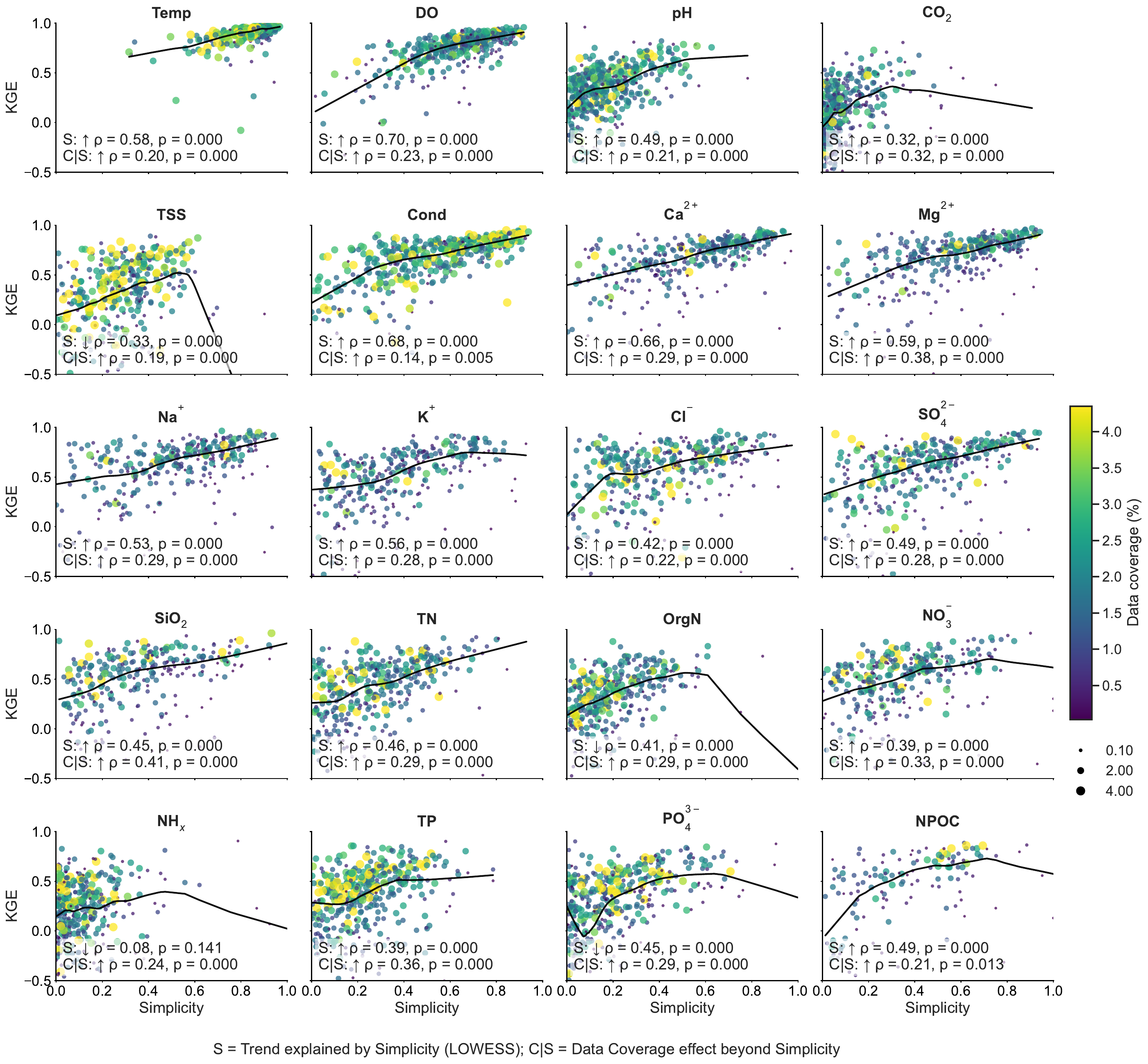}
    \caption{\setlength{\baselineskip}{1.5\baselineskip} \textcolor{black}{Relationships between model performance (DeepONet), process simplicity, and data coverage across basins. For each water quality variable (panel), each dot represents a basin and both the dot's color and size encode data coverage (darker and larger dots indicate higher coverage). A locally weighted scatterplot smoothing (LOWESS) curve summarizes the relationship between model performance (KGE) and simplicity (station-derived). The arrow marks the LOWESS slope at the highest simplicity, indicating whether performance tends to increase or decrease with simplicity. Each panel reports Spearman's correlation coefficient ($\rho$) and p-value for: (1) KGE vs. simplicity, and (2) data coverage vs. LOWESS residuals (i.e., the data coverage effect conditional on simplicity), where the residual is computed as the observed KGE minus the LOWESS predicted KGE at the same simplicity. A positive value indicates that, at fixed simplicity, higher data coverage is associated with higher-than-expected performance (KGE).}}
    \label{fig:s3}
\end{figure}

\begin{figure}[!t]
    \centering
    \includegraphics[width=1\linewidth]{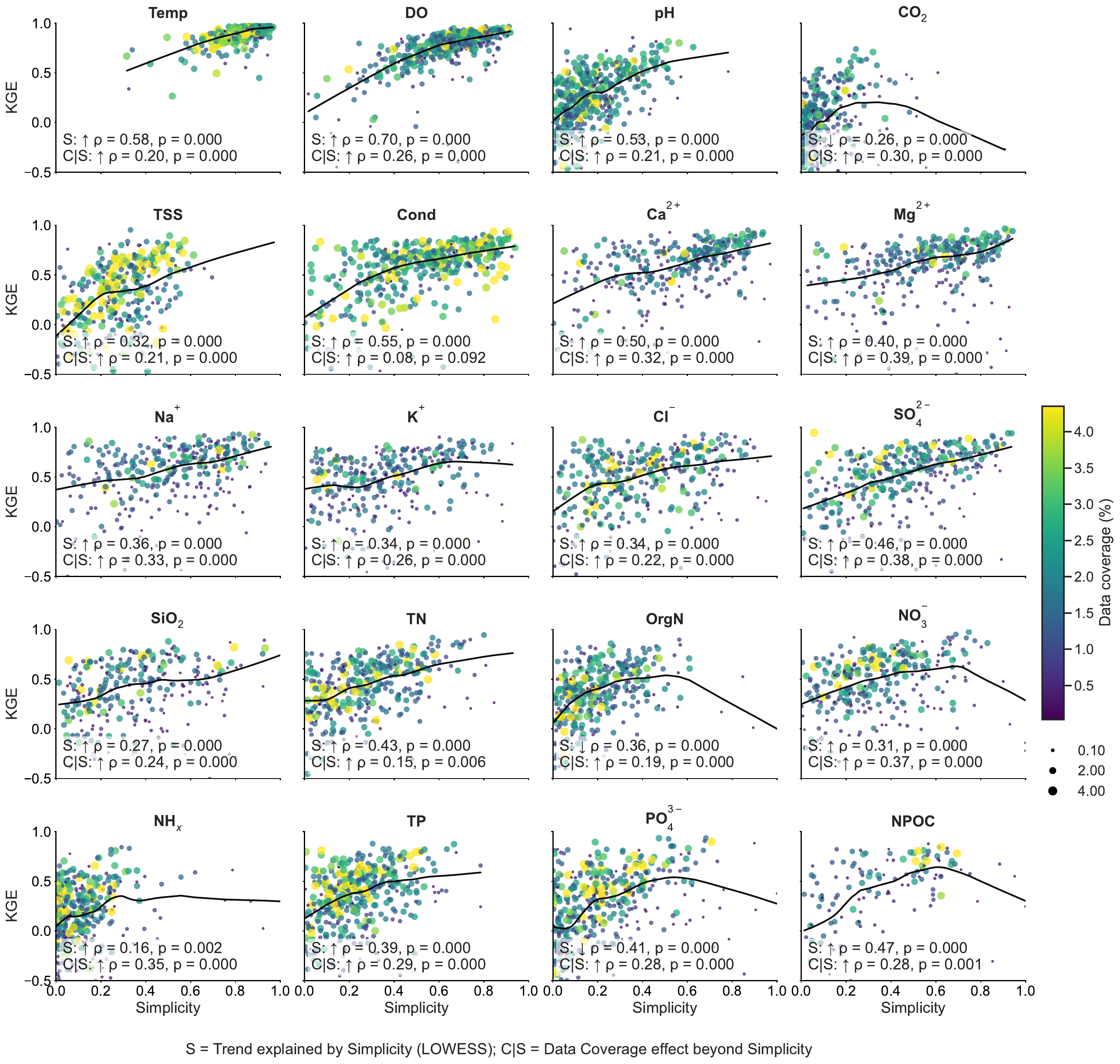}
    \caption{\setlength{\baselineskip}{1.5\baselineskip} \textcolor{black}{Relationships between model performance (Informer), process simplicity, and data coverage across basins. For each water quality variable (panel), each dot represents a basin and both the dot's color and size encode data coverage (darker and larger dots indicate higher coverage). A locally weighted scatterplot smoothing (LOWESS) curve summarizes the relationship between model performance (KGE) and simplicity (station-derived). The arrow marks the LOWESS slope at the highest simplicity, indicating whether performance tends to increase or decrease with simplicity. Each panel reports Spearman's correlation coefficient ($\rho$) and p-value for: (1) KGE vs. simplicity, and (2) data coverage vs. LOWESS residuals (i.e., the data coverage effect conditional on simplicity), where the residual is computed as the observed KGE minus the LOWESS predicted KGE at the same simplicity. A positive value indicates that, at fixed simplicity, higher data coverage is associated with higher-than-expected performance (KGE).}}
    \label{fig:s4}
\end{figure}

\begin{figure}
    \centering
    \includegraphics[width=1.0\linewidth]{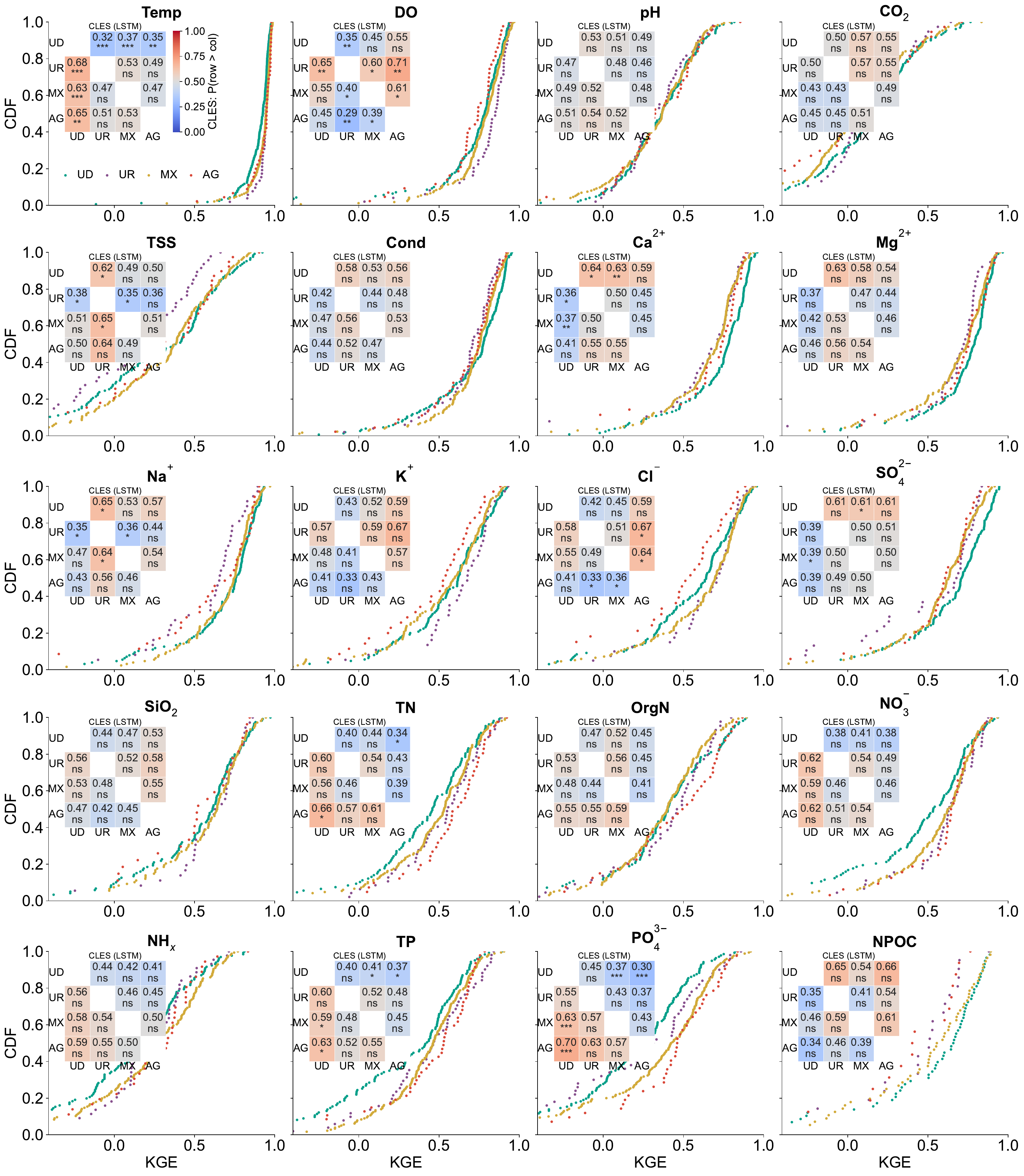}
    \caption{\setlength{\baselineskip}{1.5\baselineskip}\textcolor{black}{Multi-task LSTM model performance across basin types. CDFs of Kling-Gupta Efficiency (KGE) for undeveloped (UD), urban (UR), mixed (MX), and agricultural (AG) basins. A curve below others indicates better performance. Upper left: pairwise Common Language Effect Size (CLES) matrix~\cite{mcgraw1992common}, where each cell is $P(KGE_{row} > KGE_{col})$ and $> 0.5$ means the row group tends to have higher KGE than the column group. Two-sided Mann-Whitney U p-values ($^{***}p < 0.001$, $^{**}p < 0.01$, $^{*}p < 0.05$, and ``ns'' $p\geq 0.05$) are adjusted for multiple tests using Benjamini-Hochberg false discovery rate (FDR).}}
    \label{fig:s5}
\end{figure}

\begin{figure}
    \centering
    \includegraphics[width=1.0\linewidth]{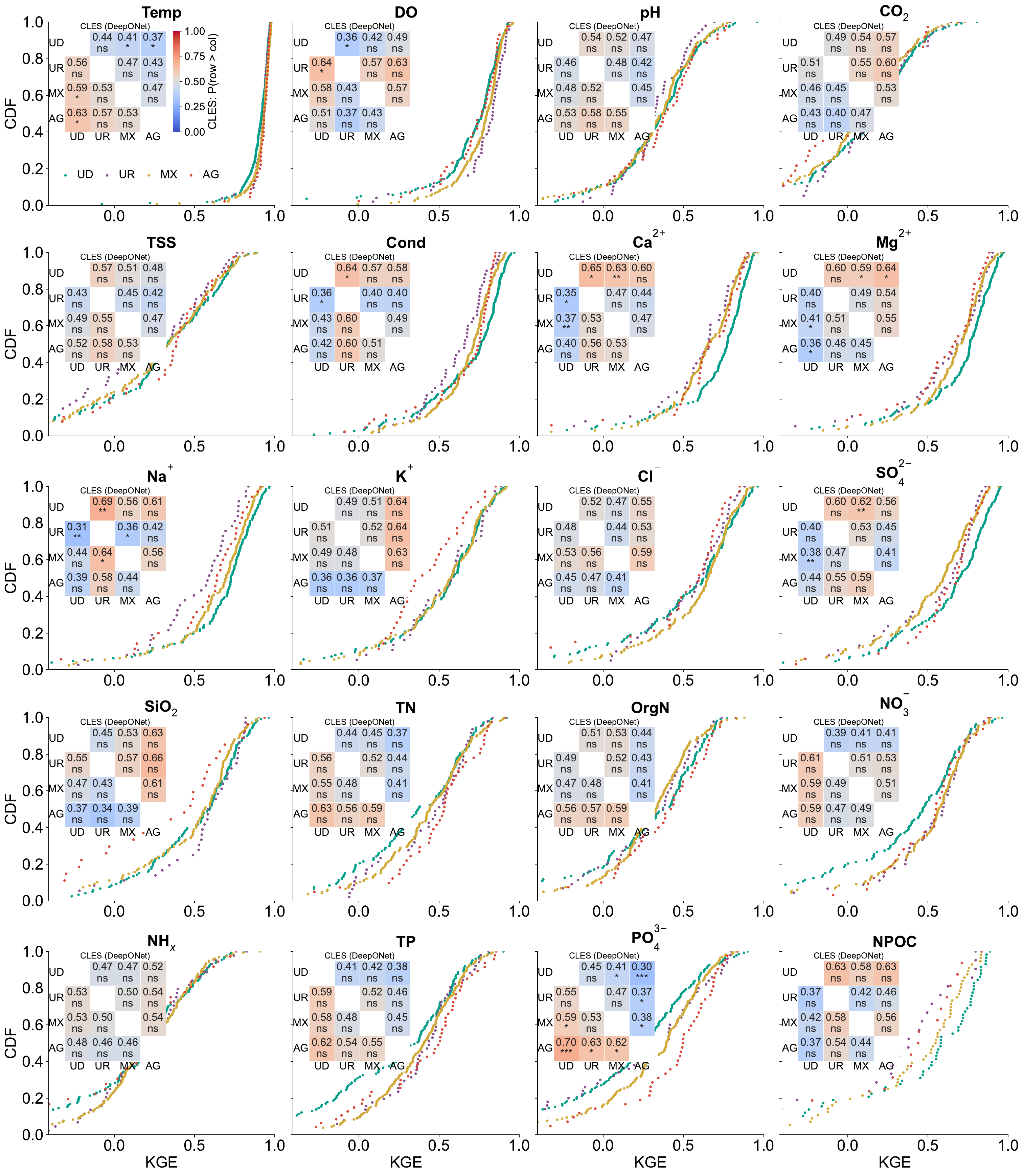}
    \caption{\setlength{\baselineskip}{1.5\baselineskip}\textcolor{black}{Multi-task DeepONet model performance across basin types. CDFs of Kling-Gupta Efficiency (KGE) for undeveloped (UD), urban (UR), mixed (MX), and agricultural (AG) basins. A curve below others indicates better performance. Upper left: pairwise Common Language Effect Size (CLES) matrix~\cite{mcgraw1992common}, where each cell is $P(KGE_{row} > KGE_{col})$ and $> 0.5$ means the row group tends to have higher KGE than the column group. Two-sided Mann-Whitney U p-values ($^{***}p < 0.001$, $^{**}p < 0.01$, $^{*}p < 0.05$, and ``ns'' $p\geq 0.05$) are adjusted for multiple tests using Benjamini-Hochberg false discovery rate (FDR).}}
    \label{fig:s6}
\end{figure}

\begin{figure}
    \centering
    \includegraphics[width=1.0\linewidth]{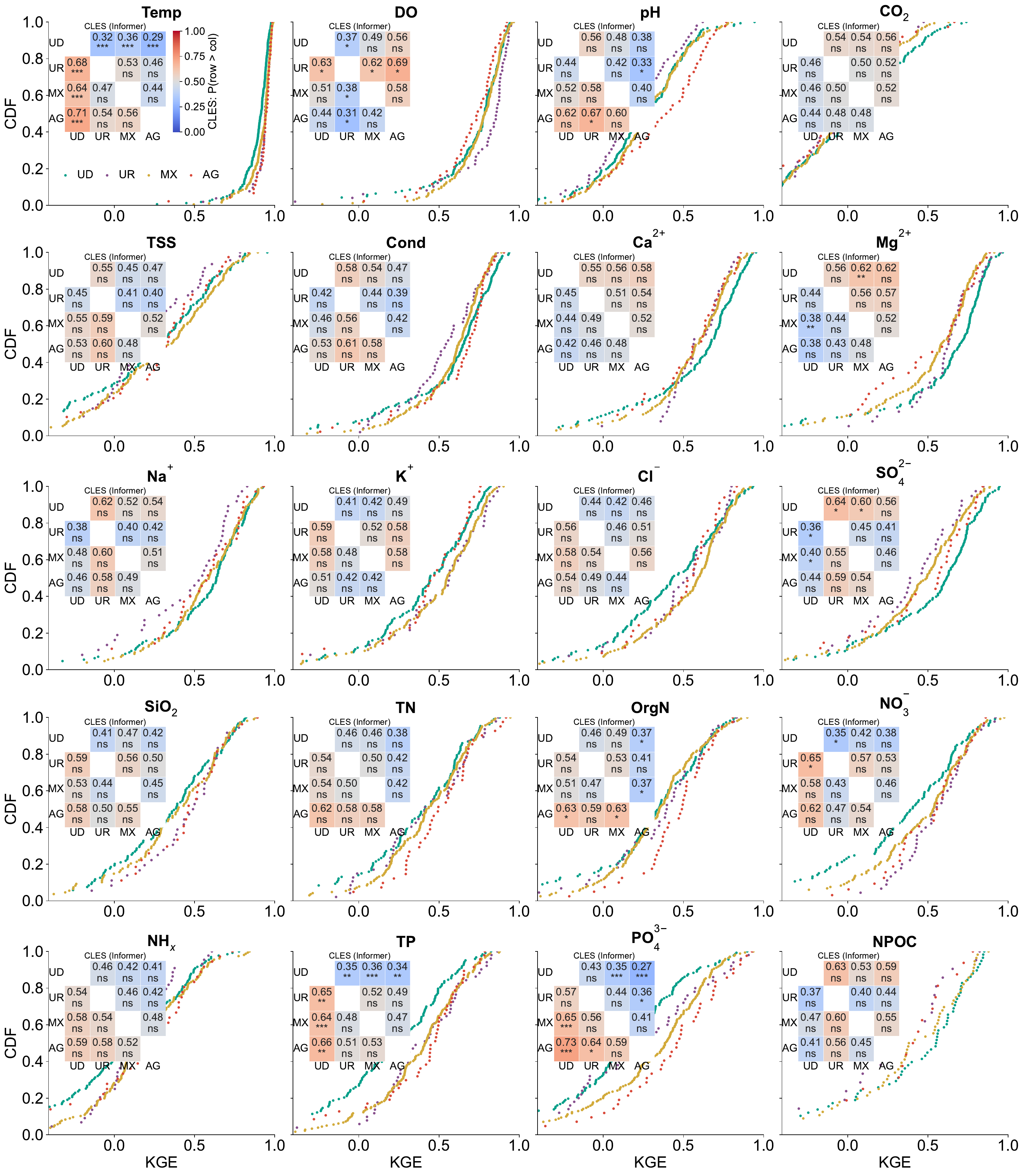}
    \caption{\setlength{\baselineskip}{1.5\baselineskip}\textcolor{black}{Multi-task Informer model performance across basin types. CDFs of Kling-Gupta Efficiency (KGE) for undeveloped (UD), urban (UR), mixed (MX), and agricultural (AG) basins. A curve below others indicates better performance. Upper left: pairwise Common Language Effect Size (CLES) matrix~\cite{mcgraw1992common}, where each cell is $P(KGE_{row} > KGE_{col})$ and $> 0.5$ means the row group tends to have higher KGE than the column group. Two-sided Mann-Whitney U p-values ($^{***}p < 0.001$, $^{**}p < 0.01$, $^{*}p < 0.05$, and ``ns'' $p\geq 0.05$) are adjusted for multiple tests using Benjamini-Hochberg false discovery rate (FDR).}}
    \label{fig:s7}
\end{figure}

\begin{figure}
    \centering
    \includegraphics[width=1.0\linewidth]{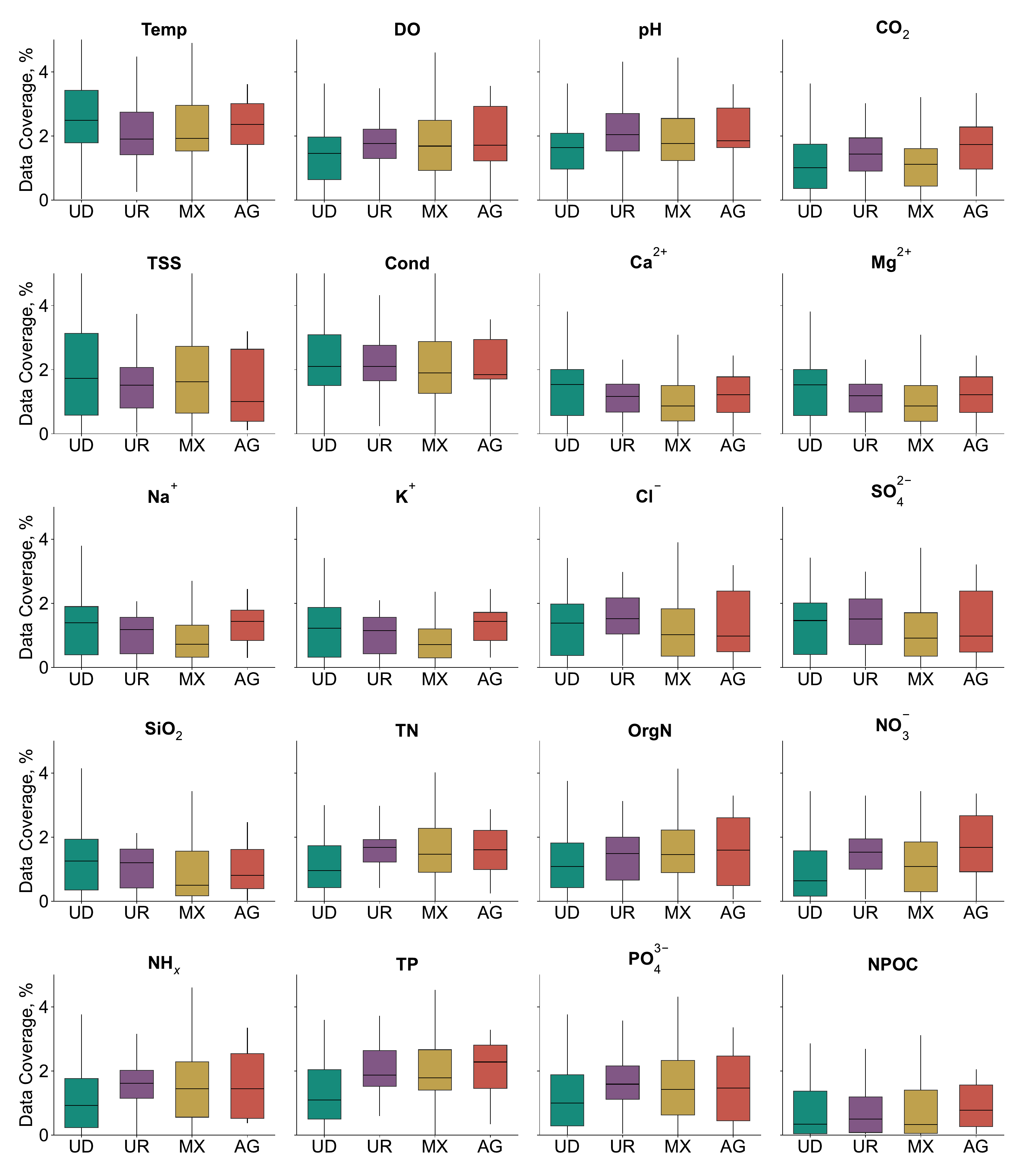}
    \caption{\setlength{\baselineskip}{1.5\baselineskip}Water quality data coverage (\%) across basins of different land use types, computed as the ratio of days monitored to the total number of days between 01/01/1982 and 12/31/2018. A coverage of 100\% indicates that water quality measurements were available for the entire study period and 0\% indicates no measurements were available. The boxplots display the median (central line), interquartile range (IQR, represented by the boxes spanning the first (Q1) to the third quartile (Q3)), and whiskers extending to $\text{Q1}-1.5\times\text{IQR}$ and $\text{Q3}+1.5\times\text{IQR}$.}
    \label{fig:s8}
\end{figure}

\textbf{Supplementary figures}
\begin{itemize}
    \item \textbf{Fig.~\ref{fig:s1}.} Spatial distribution of studied basins classified by land uses. (A) Basin types following the USGS classification criteria~\cite{spahr2010nitrate}, agricultural basins (AG, red) are defined as having more than 50\% agricultural land (PLANTNLCD06 in the GAGES‐II database) and at most 5\% urban land (DEVNLCD06). Undeveloped basins (UD, green) have at most 5\% urban land and at most 25\% agricultural land. Urban basins (UR, purple) are defined as having more than 25\% urban land and at most 25\% agricultural land, while mixed basins (MX, yellow) include all other combinations of urban, agricultural, and undeveloped land. Based on these thresholds, 3.1\% were classified as AG, 11.2\% as UR, 35.1\% as UD, and 50.6\% as MX. (B) To provide a more balanced representation in the subsequent analysis while maintaining classification logic, the AG definition was relaxed to allow up to 7\% urban land. Under this adjustment, the distribution shifted to 10.2\% AG, 11.2\% UR, 35.1\% UD, and 43.6\% MX.
    \item \textbf{Fig.~\ref{fig:s2}.} Schematic overview of the multi-task LSTM model (A), DeepONet (B), and Informer (C) to predict 20 water quality variables simultaneously by leveraging time-series hydroclimate forcings and static basin attributes as inputs.
    \item \textbf{Fig.~\ref{fig:s3}.} Relationships between model performance (DeepONet), process simplicity, and data coverage across basins. For each water quality variable (panel), each dot represents a basin and both the dot's color and size encode data coverage (darker and larger dots indicate higher coverage). A locally weighted scatterplot smoothing (LOWESS) curve summarizes the relationship between model performance (KGE) and simplicity (station-derived). The arrow marks the LOWESS slope at the highest simplicity, indicating whether performance tends to increase or decrease with simplicity. Each panel reports Spearman's correlation coefficient ($\rho$) and p-value for: (1) KGE vs. simplicity, and (2) data coverage vs. LOWESS residuals (i.e., the data coverage effect conditional on simplicity), where the residual is computed as the observed KGE minus the LOWESS predicted KGE at the same simplicity. A positive value indicates that, at fixed simplicity, higher data coverage is associated with higher-than-expected performance (KGE).
    \item \textbf{Fig.~\ref{fig:s4}.} Relationships between model performance (Informer), process simplicity, and data coverage across basins. For each water quality variable (panel), each dot represents a basin and both the dot's color and size encode data coverage (darker and larger dots indicate higher coverage). A locally weighted scatterplot smoothing (LOWESS) curve summarizes the relationship between model performance (KGE) and simplicity (station-derived). The arrow marks the LOWESS slope at the highest simplicity, indicating whether performance tends to increase or decrease with simplicity. Each panel reports Spearman's correlation coefficient ($\rho$) and p-value for: (1) KGE vs. simplicity, and (2) data coverage vs. LOWESS residuals (i.e., the data coverage effect conditional on simplicity), where the residual is computed as the observed KGE minus the LOWESS predicted KGE at the same simplicity. A positive value indicates that, at fixed simplicity, higher data coverage is associated with higher-than-expected performance (KGE).
    \item \textbf{Fig.~\ref{fig:s5}.} Multi-task LSTM model performance across basin types. CDFs of Kling-Gupta Efficiency (KGE) for undeveloped (UD), urban (UR), mixed (MX), and agricultural (AG) basins. A curve below others indicates better performance. Upper left: pairwise Common Language Effect Size (CLES) matrix~\cite{mcgraw1992common}, where each cell is $P(KGE_{row} > KGE_{col})$ and $> 0.5$ means the row group tends to have higher KGE than the column group. Two-sided Mann-Whitney U p-values ($^{***}p < 0.001$, $^{**}p < 0.01$, $^{*}p < 0.05$, and ``ns'' $p\geq 0.05$) are adjusted for multiple tests using Benjamini-Hochberg false discovery rate (FDR).
    \item \textbf{Fig.~\ref{fig:s6}.} Multi-task DeepONet model performance across basin types. CDFs of Kling-Gupta Efficiency (KGE) for undeveloped (UD), urban (UR), mixed (MX), and agricultural (AG) basins. A curve below others indicates better performance. Upper left: pairwise Common Language Effect Size (CLES) matrix~\cite{mcgraw1992common}, where each cell is $P(KGE_{row} > KGE_{col})$ and $> 0.5$ means the row group tends to have higher KGE than the column group. Two-sided Mann-Whitney U p-values ($^{***}p < 0.001$, $^{**}p < 0.01$, $^{*}p < 0.05$, and ``ns'' $p\geq 0.05$) are adjusted for multiple tests using Benjamini-Hochberg false discovery rate (FDR).
    \item \textbf{Fig.~\ref{fig:s7}.} Multi-task Informer model performance across basin types. CDFs of Kling-Gupta Efficiency (KGE) for undeveloped (UD), urban (UR), mixed (MX), and agricultural (AG) basins. A curve below others indicates better performance. Upper left: pairwise Common Language Effect Size (CLES) matrix~\cite{mcgraw1992common}, where each cell is $P(KGE_{row} > KGE_{col})$ and $> 0.5$ means the row group tends to have higher KGE than the column group. Two-sided Mann-Whitney U p-values ($^{***}p < 0.001$, $^{**}p < 0.01$, $^{*}p < 0.05$, and ``ns'' $p\geq 0.05$) are adjusted for multiple tests using Benjamini-Hochberg false discovery rate (FDR).
    \item \textbf{Fig.~\ref{fig:s8}.} Water quality data coverage (\%) across basins of different land use types, computed as the ratio of days monitored to the total number of days between 01/01/1982 and 12/31/2018. A coverage of 100\% indicates that water quality measurements were available for the entire study period and 0\% indicates no measurements were available. The boxplots display the median (central line), interquartile range (IQR, represented by the boxes spanning the first (Q1) to the third quartile (Q3)), and whiskers extending to $\text{Q1}-1.5\times\text{IQR}$ and $\text{Q3}+1.5\times\text{IQR}$.
    \item \textbf{Fig.~\ref{fig:s9}.} Simplicity index distributions across undeveloped (UD), urban (UR), mixed (MX), and agricultural (AG) basins. The simplicity index (adapted from~\cite{fang2024modeling}) quantifies the proportion of variance in water quality dynamics explained by linear relationships with runoff and annual cycles. Lower CDF (cumulative distribution function) curves indicate higher simplicity. Upper left: pairwise Common Language Effect Size (CLES) matrix~\cite{mcgraw1992common}, where each cell is $P(Simplicity_{row} > Simplicity_{col})$ and $> 0.5$ means the row group tends to have higher simplicity than the column group. Two-sided Mann-Whitney U p-values ($^{***}p < 0.001$, $^{**}p < 0.01$, $^{*}p < 0.05$, and ``ns'' $p\geq 0.05$) are adjusted for multiple tests using Benjamini-Hochberg false discovery rate (FDR).
    \item \textbf{Fig.~\ref{fig:s10}.} Comparison of predictive uncertainty in LSTM under two test-time augmentation (TTA) settings: adding Gaussian noise with a standard deviation of 0.1 only to the runoff input versus applying it to all dynamic features. Uncertainty is quantified as the standard deviation (SD) of Kling-Gupta Efficiency (KGE) across 50 TTA runs (see Methods). Boxplots show the median (central line), interquartile range (IQR; Q1-Q3), and whiskers extending to $\text{Q1} - 1.5 \times \text{IQR}$ and $\text{Q3} + 1.5 \times \text{IQR}$.
    \item \textbf{Fig.~\ref{fig:s11}.} Relationship between predictive performance and uncertainty in LSTM with Monte Carlo dropout. (\textbf{A}) The uncertainty of model predictions across different water quality variables, quantified as the standard deviation (SD) of the Kling-Gupta Efficiency (KGE) obtained from Monte Carlo dropout across 50 simulations (see Methods). The boxplots show the median (central line), interquartile range (IQR, represented by the boxes spanning the first (Q1) to the third quartile (Q3)), and whiskers extending to $\text{Q1}-1.5\times\text{IQR}$ and $\text{Q3} + 1.5\times\text{IQR}$. (\textbf{B}) A strong negative correlation ($r = -0.62$, $p < 0.001$) between the baseline median KGE across 482 basins and the median uncertainty (SD of KGE), indicating that water quality variables with lower predictive performance tend to exhibit higher uncertainty. The shaded region around the regression line represents the 95\% confidence interval.
    \item \textbf{Fig.~\ref{fig:s12}.} Group-level Integrated Gradients (IG). (A) Total phosphorus (TP: USGS 00665) across four LSTM modeling configurations. (B) 20 water quality variables for the LSTM with missing-value filling set to 0 (-1 in this work and the previous study~\cite{fang2024modeling}). Five feature groups are: meteorological forcings (M), runoff/discharge (Q), rainfall chemistry (RC), vegetation indices (V), and basin attributes (BA) (full group definitions in Methods). IG values are computed for each sample; the feature importance is the mean absolute IG over samples, and the group importance is the mean of feature-level $|$IG$|$ within that group. The ribbon width is normalized for each model configuration so widths to all groups sum to 1. These results indicate that IG-based attributions are sensitive to modeling setup: task formulation (single vs multi-task), input representation (raw discharge vs area-normalized), or features missing-value handling (0 or -1).
    \item \textbf{Fig.~\ref{fig:s13}.} Performance-based feature importance comparison across five groups in LSTM: meteorological forcings (M), runoff (Q), rainfall chemistry (RC), vegetation indices (V), and basin attributes (BA). Feature group details are provided in Methods. In the Ablation approach (light red boxes), feature importance is quantified by the reduction in Kling-Gupta Efficiency (KGE) when that group is removed from the full model. In the Traverse approach (dark red boxes), the feature importance of each group is calculated as the average KGE reduction across all possible feature group combinations with and without the target group (see Methods). The boxplots show the median (central line), interquartile range (IQR, represented by the boxes spanning the first (Q1) to the third quartile (Q3)), and whiskers extending to $\text{Q1}-1.5\times\text{IQR}$ and $\text{Q3} + 1.5\times\text{IQR}$. For both methods, Wilcoxon signed-rank tests were performed to assess whether median KGE reductions across 482 basins significantly exceeded zero (black stars; $^{***}p < 0.001$, $^{**}p < 0.01$, $^{*}p < 0.05$, and ``ns'' for $p\geq0.05$). Numbers above each box indicate the relative importance ranking of that group.
    \item \textbf{Fig.~\ref{fig:s14}.} Performance-based feature importance comparison across five groups in Informer: meteorological forcings (M), runoff (Q), rainfall chemistry (RC), vegetation indices (V), and basin attributes (BA). Feature group details are provided in Methods. In the Ablation approach (light red boxes), feature importance is quantified by the reduction in Kling-Gupta Efficiency (KGE) when that group is removed from the full model. In the Traverse approach (dark red boxes), the feature importance of each group is calculated as the average KGE reduction across all possible feature group combinations with and without the target group (see Methods). The boxplots show the median (central line), interquartile range (IQR, represented by the boxes spanning the first (Q1) to the third quartile (Q3)), and whiskers extending to $\text{Q1}-1.5\times\text{IQR}$ and $\text{Q3} + 1.5\times\text{IQR}$. For both methods, Wilcoxon signed-rank tests were performed to assess whether median KGE reductions across 482 basins significantly exceeded zero (black stars; $^{***}p < 0.001$, $^{**}p < 0.01$, $^{*}p < 0.05$, and ``ns'' for $p\geq0.05$). Numbers above each box indicate the relative importance ranking of that group.
    \item \textbf{Fig.~\ref{fig:s15}.} Performance-based feature importance comparison across five groups in DeepONet: meteorological forcings (M), runoff (Q), rainfall chemistry (RC), vegetation indices (V), and basin attributes (BA). Feature group details are provided in Methods. In the Ablation approach (light red boxes), feature importance is quantified by the reduction in Kling-Gupta Efficiency (KGE) when that group is removed from the full model. In the Traverse approach (dark red boxes), the feature importance of each group is calculated as the average KGE reduction across all possible feature group combinations with and without the target group (see Methods). The boxplots show the median (central line), interquartile range (IQR, represented by the boxes spanning the first (Q1) to the third quartile (Q3)), and whiskers extending to $\text{Q1}-1.5\times\text{IQR}$ and $\text{Q3} + 1.5\times\text{IQR}$. For both methods, Wilcoxon signed-rank tests were performed to assess whether median KGE reductions across 482 basins significantly exceeded zero (black stars; $^{***}p < 0.001$, $^{**}p < 0.01$, $^{*}p < 0.05$, and ``ns'' for $p\geq0.05$). Numbers above each box indicate the relative importance ranking of that group.
    \item \textbf{Fig.~\ref{fig:s16}.} Context-dependent feature importance (KGE reduction) of meteorological variables (M) and runoff (Q) derived via the Traverse method for LSTM. Dark blue boxplots represent KGE reduction from excluding Q when M is already excluded, whereas light blue boxplots represent excluding Q when M is included. Similarly, dark red boxplots show the KGE reduction from excluding M when Q is absent, whereas light red boxplots represent excluding M when Q is included. Wilcoxon signed-rank tests were conducted to assess whether median KGE reductions from subsets lacking Q or M were significantly greater than those from subsets where Q or M were present ($^{***}p < 0.001$). The results indicate that meteorological variables become largely redundant when runoff is included.
    \item \textbf{Fig.~\ref{fig:s17}.} Context-dependent feature importance (KGE reduction) of meteorological variables (M) and runoff (Q) derived via the Traverse method \textcolor{black}{for Informer}. Dark blue boxplots represent KGE reduction from excluding Q when M is already excluded, whereas light blue boxplots represent excluding Q when M is included. Similarly, dark red boxplots show the KGE reduction from excluding M when Q is absent, whereas light red boxplots represent excluding M when Q is included. Wilcoxon signed-rank tests were conducted to assess whether median KGE reductions from subsets lacking Q or M were significantly greater than those from subsets where Q or M were present ($^{***}p < 0.001$). The results indicate that meteorological variables become largely redundant when runoff is included.
    \item \textbf{Fig.~\ref{fig:s18}.} Context-dependent feature importance (KGE reduction) of meteorological variables (M) and runoff (Q) derived via the Traverse method \textcolor{black}{for DeepONet}. Dark blue boxplots represent KGE reduction from excluding Q when M is already excluded, whereas light blue boxplots represent excluding Q when M is included. Similarly, dark red boxplots show the KGE reduction from excluding M when Q is absent, whereas light red boxplots represent excluding M when Q is included. Wilcoxon signed-rank tests were conducted to assess whether median KGE reductions from subsets lacking Q or M were significantly greater than those from subsets where Q or M were present ($^{***}p < 0.001$). The results indicate that meteorological variables become largely redundant when runoff is included.
    \item \textbf{Fig.~\ref{fig:s19}.} Boxplot of Kling-Gupta Efficiency (KGE) values for the testing basins \textcolor{black}{predicted by three different deep learning models} under the spatial training-testing split, evaluated across 20 predicted water quality variables associated with physical/chemical properties, geochemical weathering processes, and nutrient cycling, respectively. Each boxplot shows the median (central line), interquartile range (IQR, represented by the boxes spanning the first (Q1) to the third quartile (Q3)), and whiskers extending to $\text{Q1}-1.5\times\text{IQR}$ and $\text{Q3} + 1.5\times\text{IQR}$. The number labeled on the box indicates the median. Wilcoxon signed-rank tests with False Discovery Rate-Benjamini-Hochberg (FDR-BH) correction indicate no significant performance differences among the three models.
\end{itemize}
\newpage
\textbf{Supplementary tables}
\begin{itemize}
    \item \textbf{Table~\ref{tab:wq_statistics}.} Summary of the studied water quality variables and the average number of observations per basin, based on 482 U.S. rivers between 01/01/1982 and 12/31/2018.
    \item \textbf{Table~\ref{tab:inputs1}.} Model input features, consisting of 25 time series variables and 49 static basin attributes (sourced from the GAGES-II database).
\end{itemize}
\clearpage

\clearpage

\clearpage

\clearpage

\clearpage

\clearpage

\clearpage

\clearpage

\clearpage

\begin{figure}
    \centering
    \includegraphics[width=1.0\linewidth]{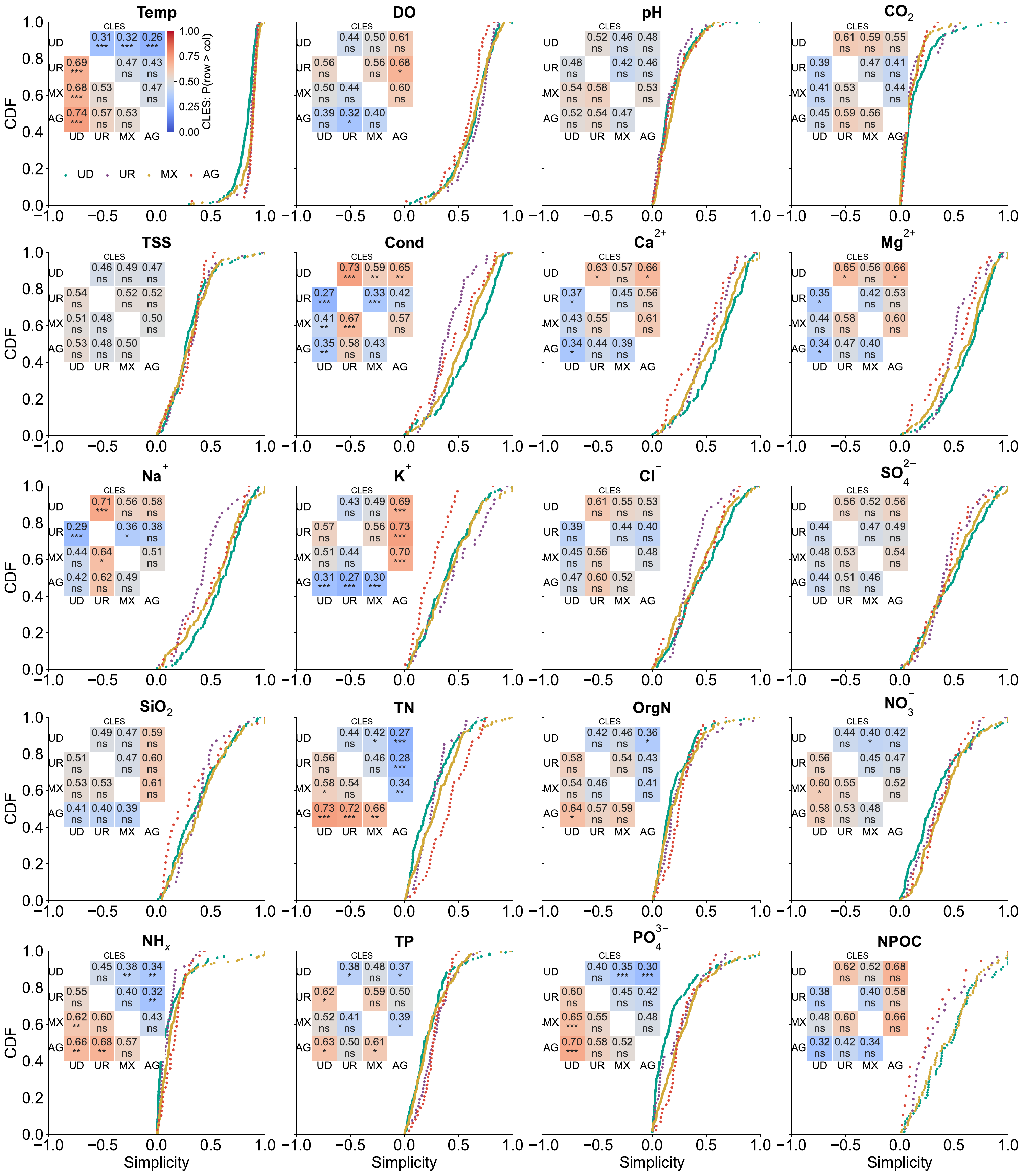}
    \caption{\setlength{\baselineskip}{1.5\baselineskip}Simplicity index distributions across undeveloped (UD), urban (UR), mixed (MX), and agricultural (AG) basins. The simplicity index (adapted from~\cite{fang2024modeling}) quantifies the proportion of variance in water quality dynamics explained by linear relationships with runoff and annual cycles. Lower CDF (cumulative distribution function) curves indicate higher simplicity. Upper left: pairwise Common Language Effect Size (CLES) matrix~\cite{mcgraw1992common}, where each cell is $P(Simplicity_{row} > Simplicity_{col})$ and $> 0.5$ means the row group tends to have higher simplicity than the column group. Two-sided Mann-Whitney U p-values ($^{***}p < 0.001$, $^{**}p < 0.01$, $^{*}p < 0.05$, and ``ns'' $p\geq 0.05$) are adjusted for multiple tests using Benjamini-Hochberg false discovery rate (FDR).}
    \label{fig:s9}
\end{figure}
\clearpage

\clearpage
\begin{figure}
    \centering
    \vspace{-9cm}
    \includegraphics[width=0.9\linewidth]{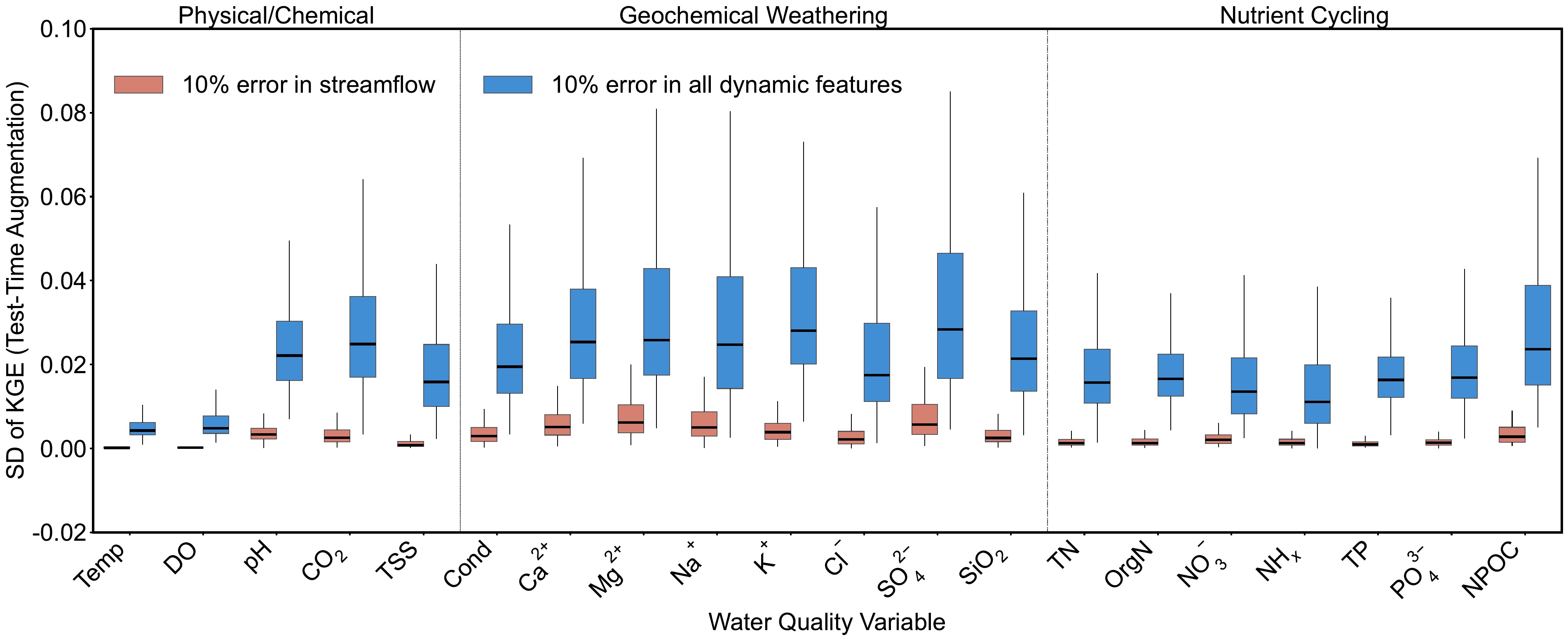}\caption{\setlength{\baselineskip}{1.5\baselineskip} \textcolor{black}{Comparison of predictive uncertainty in LSTM under two test-time augmentation (TTA) settings: adding Gaussian noise with a standard deviation of 0.1 only to the runoff input versus applying it to all dynamic features. Uncertainty is quantified as the standard deviation (SD) of Kling-Gupta Efficiency (KGE) across 50 TTA runs (see Methods). Boxplots show the median (central line), interquartile range (IQR; Q1-Q3), and whiskers extending to $\text{Q1} - 1.5 \times \text{IQR}$ and $\text{Q3} + 1.5 \times \text{IQR}$.}}
    \label{fig:s10}
\end{figure}

\clearpage
\begin{figure}
    \centering
    \vspace{-9cm}
    \includegraphics[width=0.9\linewidth]{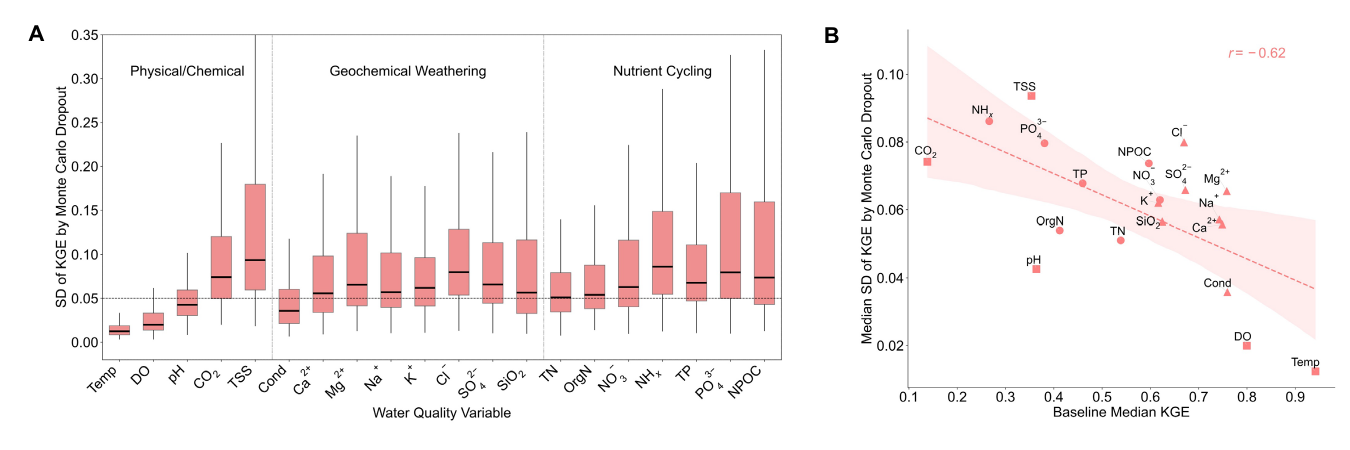}\caption{\setlength{\baselineskip}{1.5\baselineskip} \textcolor{black}{Relationship between predictive performance and uncertainty in LSTM with Monte Carlo dropout. (\textbf{A}) The uncertainty of model predictions across different water quality variables, quantified as the standard deviation (SD) of the Kling-Gupta Efficiency (KGE) obtained from Monte Carlo dropout across 50 simulations (see Methods). The boxplots show the median (central line), interquartile range (IQR, represented by the boxes spanning the first (Q1) to the third quartile (Q3)), and whiskers extending to $\text{Q1}-1.5\times\text{IQR}$ and $\text{Q3} + 1.5\times\text{IQR}$. (\textbf{B}) A strong negative correlation ($r = -0.62$, $p < 0.001$) between the baseline median KGE across 482 basins and the median uncertainty (SD of KGE), indicating that water quality variables with lower predictive performance tend to exhibit higher uncertainty. The shaded region around the regression line represents the 95\% confidence interval.}}
    \label{fig:s11}
\end{figure}

\clearpage
\begin{figure}
    \centering
    \vspace{-9cm}
    \includegraphics[width=0.9\linewidth]{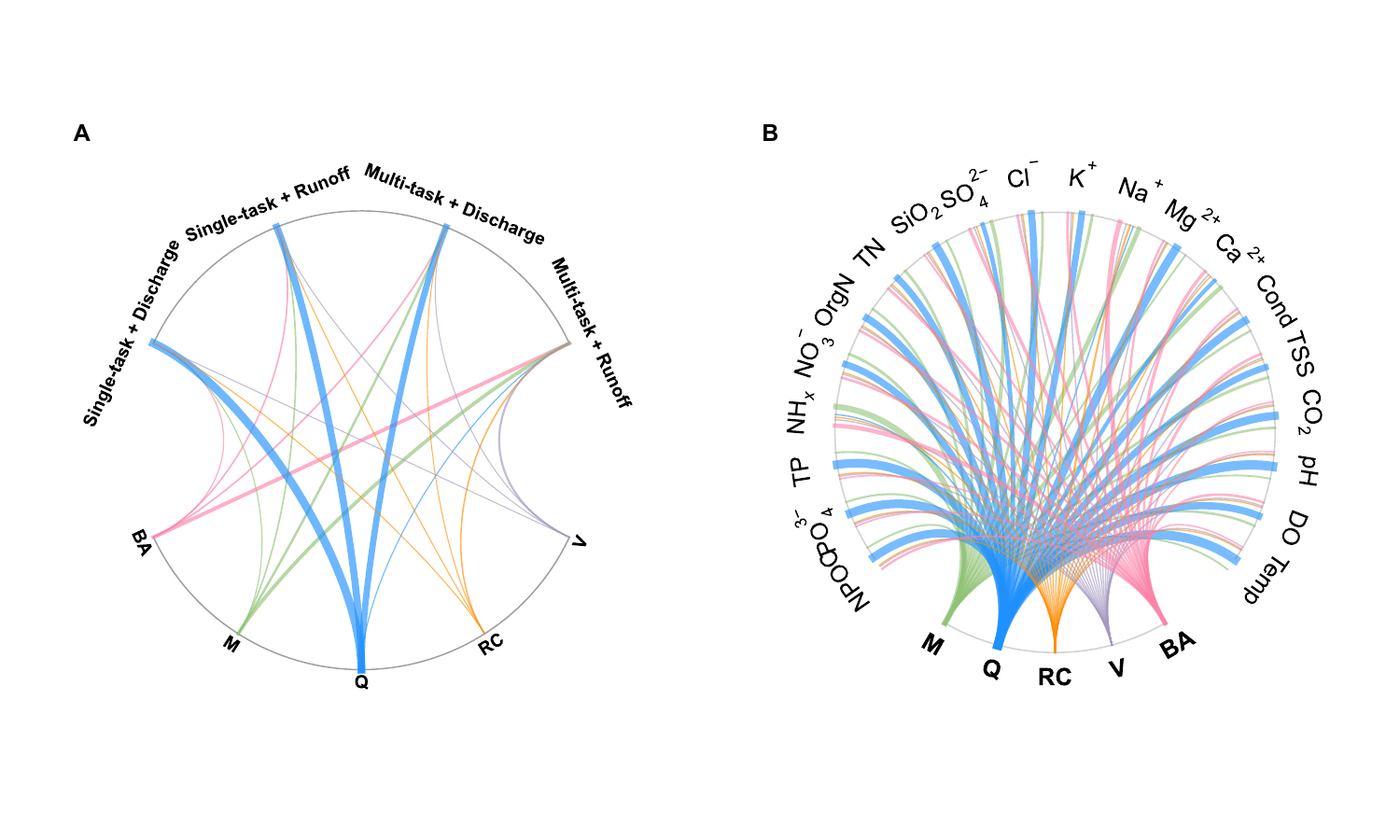}\caption{\setlength{\baselineskip}{1.5\baselineskip} \textcolor{black}{Group-level Integrated Gradients (IG). (A) Total phosphorus (TP: USGS 00665) across four LSTM modeling configurations. (B) 20 water quality variables for the LSTM with missing-value filling set to 0 (-1 in this work and the previous study~\cite{fang2024modeling}). Five feature groups are: meteorological forcings (M), runoff/discharge (Q), rainfall chemistry (RC), vegetation indices (V), and basin attributes (BA) (full group definitions in Methods). IG values are computed for each sample; the feature importance is the mean absolute IG over samples, and the group importance is the mean of feature-level $|$IG$|$ within that group. The ribbon width is normalized for each model configuration so widths to all groups sum to 1. These results indicate that IG-based attributions are sensitive to modeling setup: task formulation (single vs multi-task), input representation (raw discharge vs area-normalized), or features missing-value handling (0 or -1).}}
    \label{fig:s12}
\end{figure}

\clearpage
\begin{figure}
    \centering
    \includegraphics[width=0.9\linewidth]{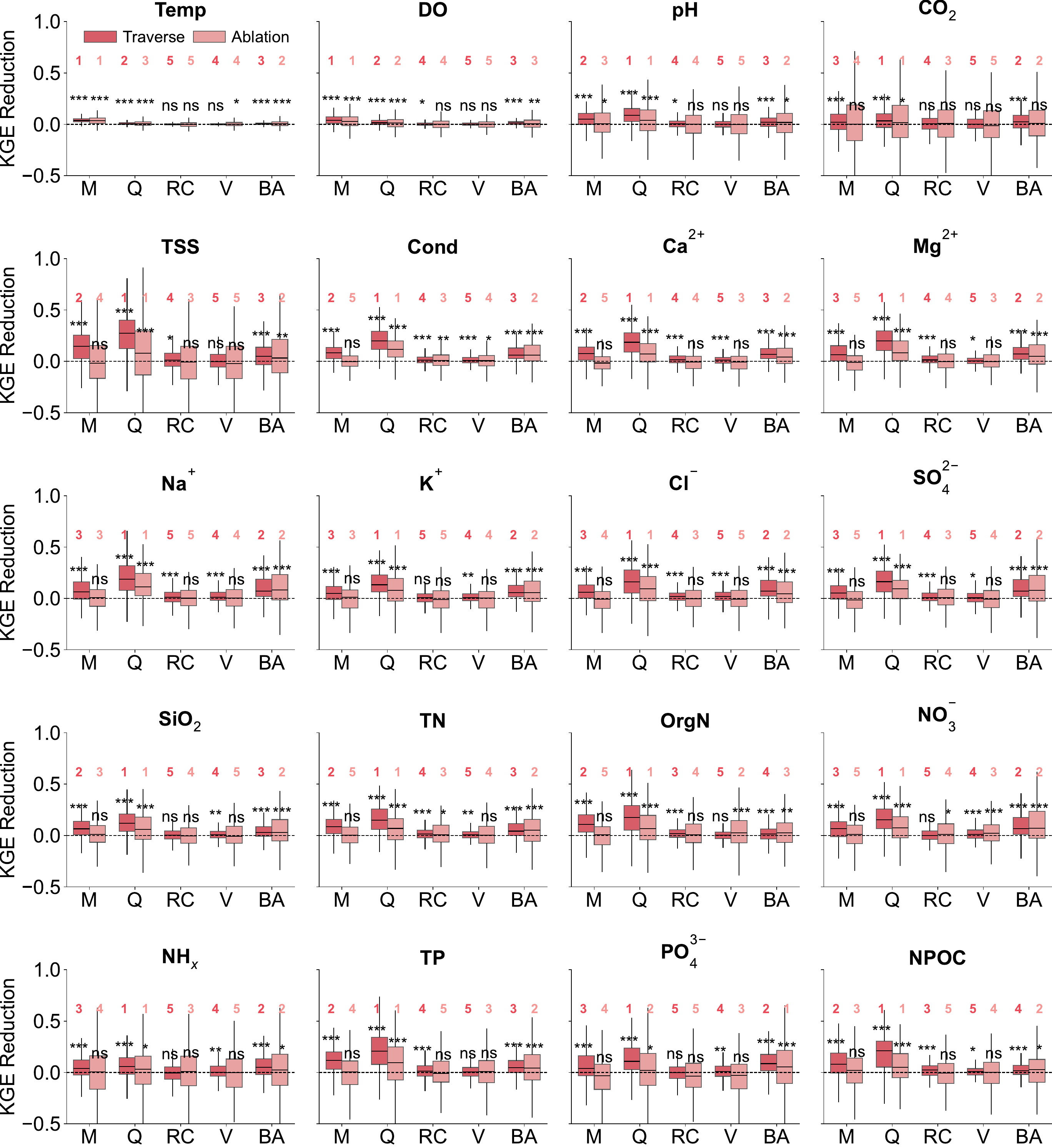}
    \caption{\setlength{\baselineskip}{1.5\baselineskip}\textcolor{black}{Performance-based} feature importance comparison across five groups \textcolor{black}{in LSTM}: meteorological forcings (M), runoff (Q), rainfall chemistry (RC), vegetation indices (V), and basin attributes (BA). Feature group details are provided in Methods. In the Ablation approach (light red boxes), feature importance is quantified by the reduction in Kling-Gupta Efficiency (KGE) when that group is removed from the full model. In the Traverse approach (dark red boxes), the feature importance of each group is calculated as the average KGE reduction across all possible feature group combinations with and without the target group (see Methods). The boxplots show the median (central line), interquartile range (IQR, represented by the boxes spanning the first (Q1) to the third quartile (Q3)), and whiskers extending to $\text{Q1}-1.5\times\text{IQR}$ and $\text{Q3} + 1.5\times\text{IQR}$. For both methods, Wilcoxon signed-rank tests were performed to assess whether median KGE reductions across 482 basins significantly exceeded zero (black stars; $^{***}p < 0.001$, $^{**}p < 0.01$, $^{*}p < 0.05$, and ``ns'' for $p\geq0.05$). Numbers above each box indicate the relative importance ranking of that group.}
    \label{fig:s13}
\end{figure}

\clearpage
\begin{figure}
    \centering
    \includegraphics[width=0.9\linewidth]{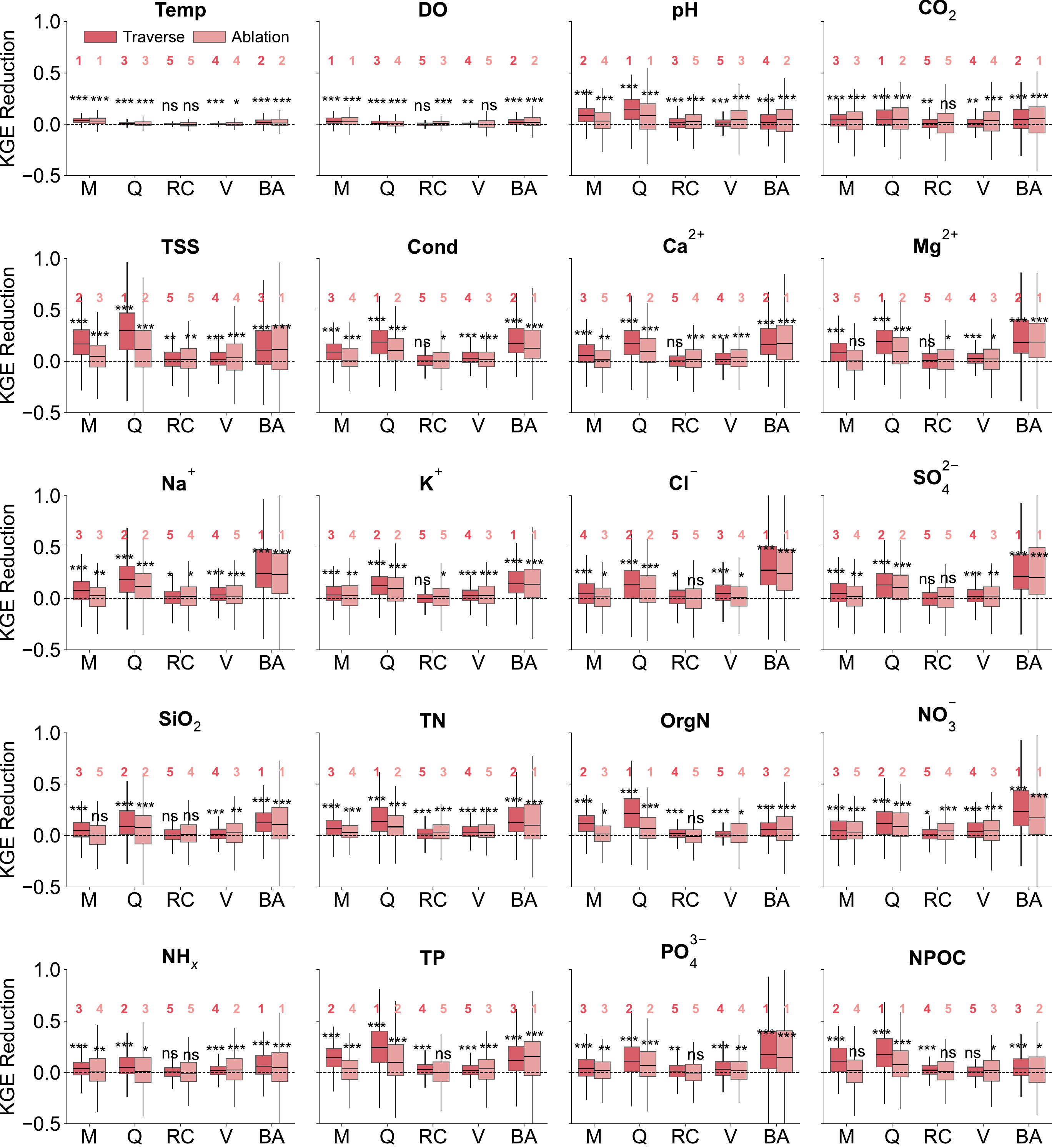}
    \caption{\setlength{\baselineskip}{1.5\baselineskip}\textcolor{black}{Performance-based feature importance comparison across five groups in Informer: meteorological forcings (M), runoff (Q), rainfall chemistry (RC), vegetation indices (V), and basin attributes (BA). Feature group details are provided in Methods. In the Ablation approach (light red boxes), feature importance is quantified by the reduction in Kling-Gupta Efficiency (KGE) when that group is removed from the full model. In the Traverse approach (dark red boxes), the feature importance of each group is calculated as the average KGE reduction across all possible feature group combinations with and without the target group (see Methods). The boxplots show the median (central line), interquartile range (IQR, represented by the boxes spanning the first (Q1) to the third quartile (Q3)), and whiskers extending to $\text{Q1}-1.5\times\text{IQR}$ and $\text{Q3} + 1.5\times\text{IQR}$. For both methods, Wilcoxon signed-rank tests were performed to assess whether median KGE reductions across 482 basins significantly exceeded zero (black stars; $^{***}p < 0.001$, $^{**}p < 0.01$, $^{*}p < 0.05$, and ``ns'' for $p\geq0.05$). Numbers above each box indicate the relative importance ranking of that group.}}
    \label{fig:s14}
\end{figure}

\clearpage
\begin{figure}
    \centering
    \includegraphics[width=0.9\linewidth]{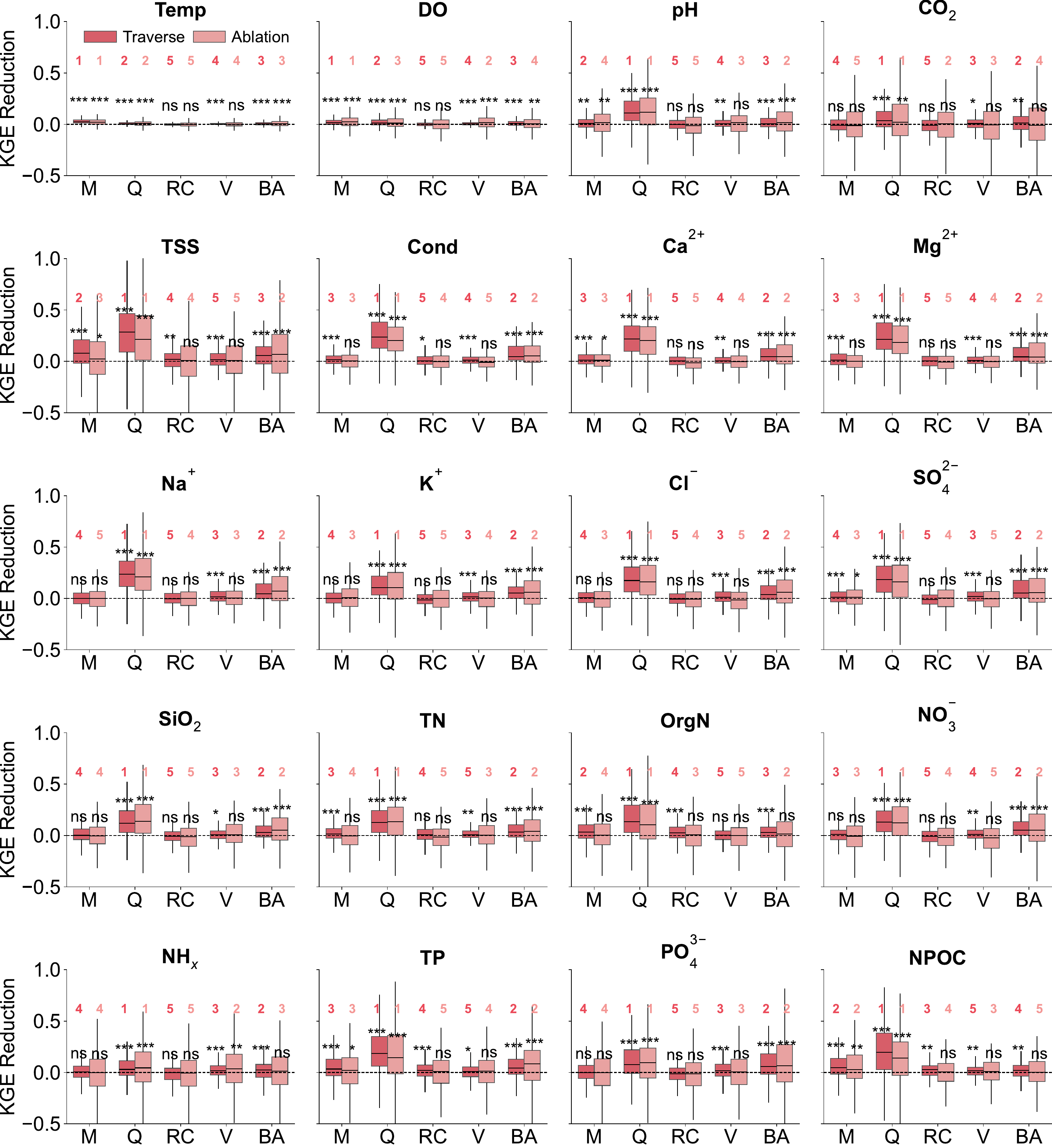}
    \caption{\setlength{\baselineskip}{1.5\baselineskip}\textcolor{black}{Performance-based feature importance comparison across five groups in DeepONet: meteorological forcings (M), runoff (Q), rainfall chemistry (RC), vegetation indices (V), and basin attributes (BA). Feature group details are provided in Methods. In the Ablation approach (light red boxes), feature importance is quantified by the reduction in Kling-Gupta Efficiency (KGE) when that group is removed from the full model. In the Traverse approach (dark red boxes), the feature importance of each group is calculated as the average KGE reduction across all possible feature group combinations with and without the target group (see Methods). The boxplots show the median (central line), interquartile range (IQR, represented by the boxes spanning the first (Q1) to the third quartile (Q3)), and whiskers extending to $\text{Q1}-1.5\times\text{IQR}$ and $\text{Q3} + 1.5\times\text{IQR}$. For both methods, Wilcoxon signed-rank tests were performed to assess whether median KGE reductions across 482 basins significantly exceeded zero (black stars; $^{***}p < 0.001$, $^{**}p < 0.01$, $^{*}p < 0.05$, and ``ns'' for $p\geq0.05$). Numbers above each box indicate the relative importance ranking of that group.}}
    \label{fig:s15}
\end{figure}

\clearpage
\begin{figure}
    \centering
    \includegraphics[width=0.9\linewidth]{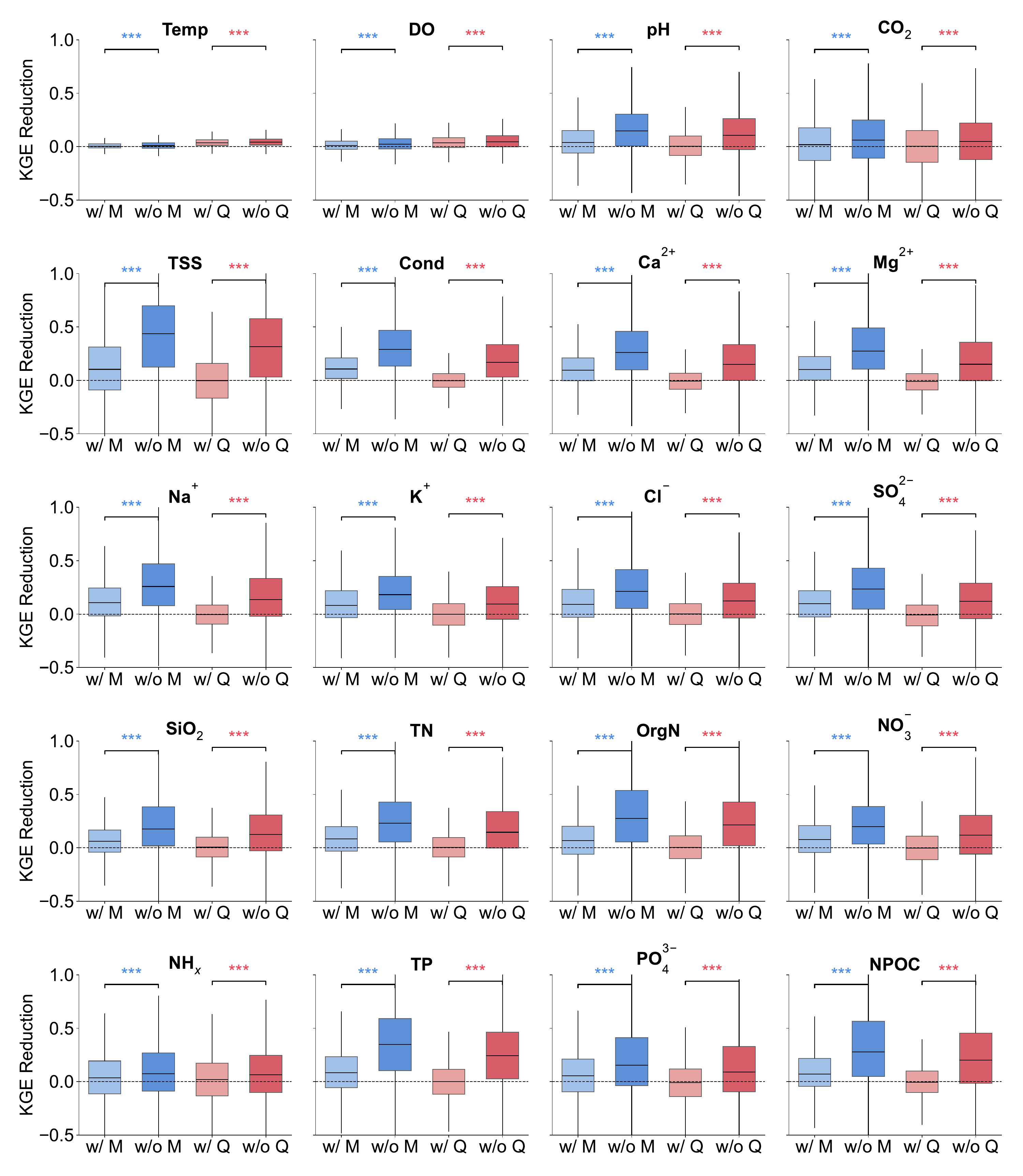}    \caption{\setlength{\baselineskip}{1.5\baselineskip} Context-dependent feature importance (KGE reduction) of meteorological variables (M) and runoff (Q) derived via the Traverse method for LSTM. Dark blue boxplots represent KGE reduction from excluding Q when M is already excluded, whereas light blue boxplots represent excluding Q when M is included. Similarly, dark red boxplots show the KGE reduction from excluding M when Q is absent, whereas light red boxplots represent excluding M when Q is included. Wilcoxon signed-rank tests were conducted to assess whether median KGE reductions from subsets lacking Q or M were significantly greater than those from subsets where Q or M were present ($^{***}p < 0.001$). The results indicate that meteorological variables become largely redundant when runoff is included.}
    \label{fig:s16}
\end{figure}

\clearpage
\begin{figure}
    \centering
    \includegraphics[width=0.9\linewidth]{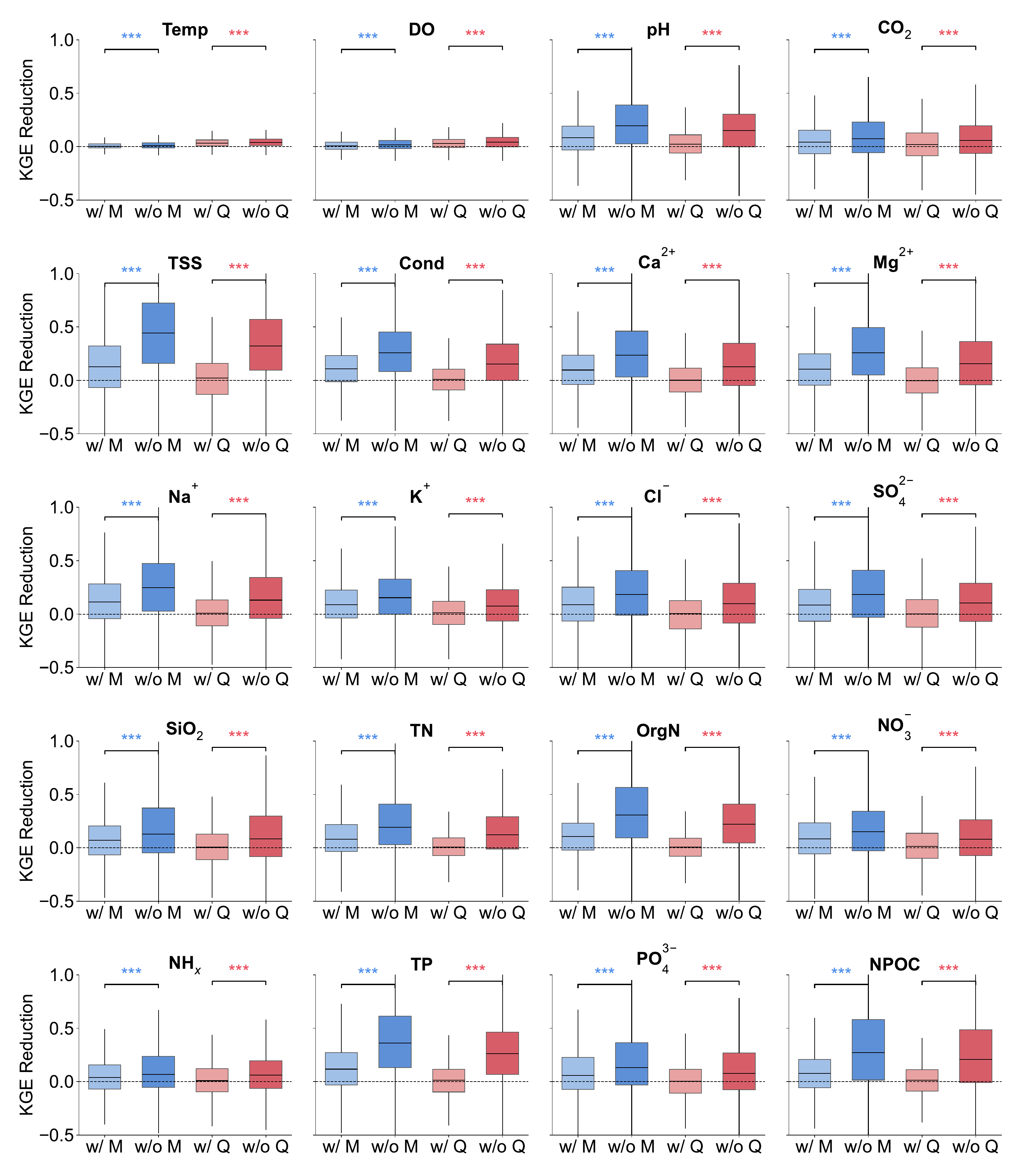}    \caption{\setlength{\baselineskip}{1.5\baselineskip} \textcolor{black}{Context-dependent feature importance (KGE reduction) of meteorological variables (M) and runoff (Q) derived via the Traverse method \textcolor{black}{for Informer}. Dark blue boxplots represent KGE reduction from excluding Q when M is already excluded, whereas light blue boxplots represent excluding Q when M is included. Similarly, dark red boxplots show the KGE reduction from excluding M when Q is absent, whereas light red boxplots represent excluding M when Q is included. Wilcoxon signed-rank tests were conducted to assess whether median KGE reductions from subsets lacking Q or M were significantly greater than those from subsets where Q or M were present ($^{***}p < 0.001$). The results indicate that meteorological variables become largely redundant when runoff is included.}}
    \label{fig:s17}
\end{figure}

\clearpage
\begin{figure}
    \centering
    \includegraphics[width=0.9\linewidth]{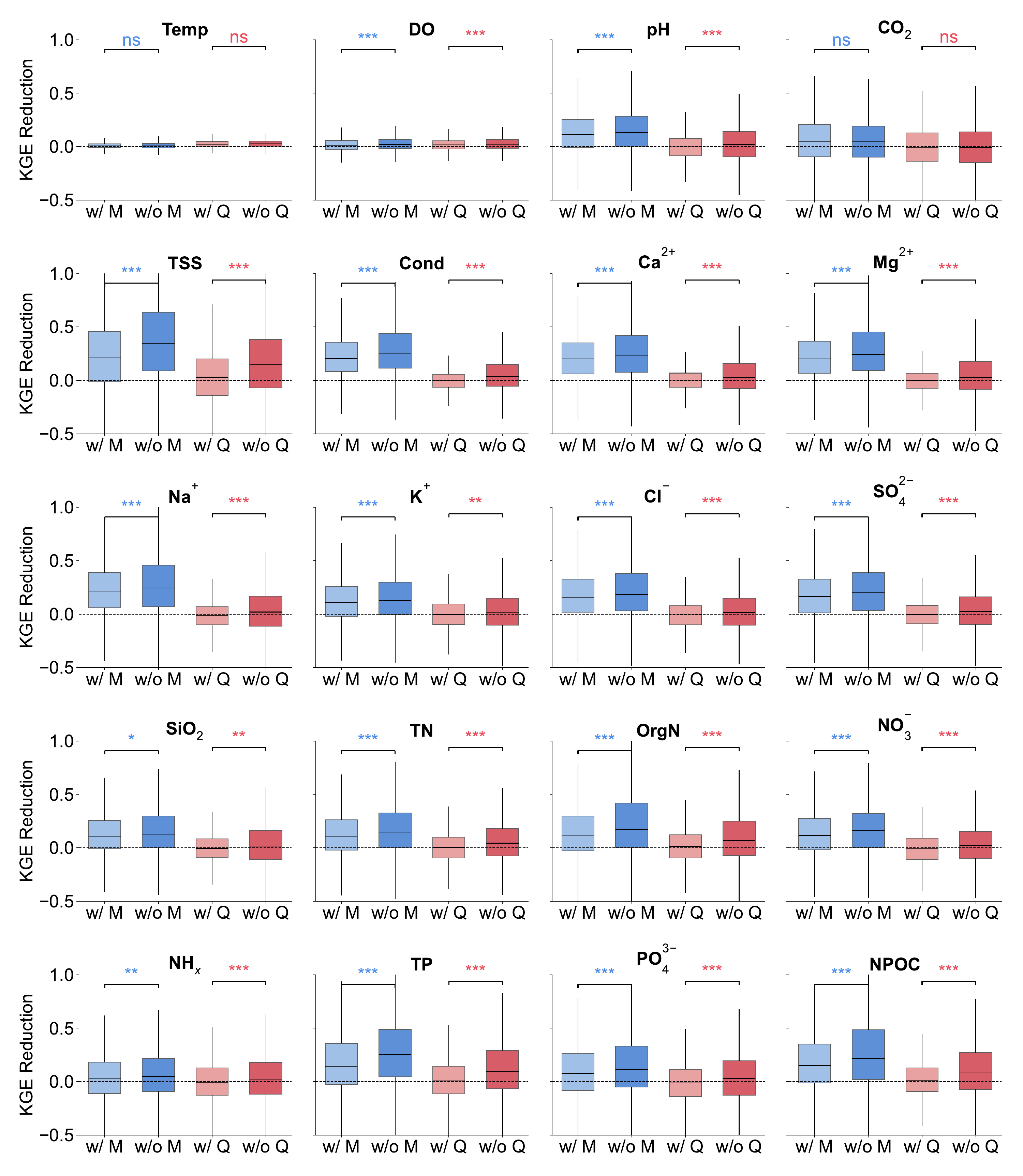}    \caption{\setlength{\baselineskip}{1.5\baselineskip} \textcolor{black}{Context-dependent feature importance (KGE reduction) of meteorological variables (M) and runoff (Q) derived via the Traverse method \textcolor{black}{for DeepONet}. Dark blue boxplots represent KGE reduction from excluding Q when M is already excluded, whereas light blue boxplots represent excluding Q when M is included. Similarly, dark red boxplots show the KGE reduction from excluding M when Q is absent, whereas light red boxplots represent excluding M when Q is included. Wilcoxon signed-rank tests were conducted to assess whether median KGE reductions from subsets lacking Q or M were significantly greater than those from subsets where Q or M were present ($^{***}p < 0.001$). The results indicate that meteorological variables become largely redundant when runoff is included.}}
    \label{fig:s18}
\end{figure}

\clearpage
\begin{figure}
    \centering
    \vspace{-9cm}
    \includegraphics[width=1.0\linewidth]{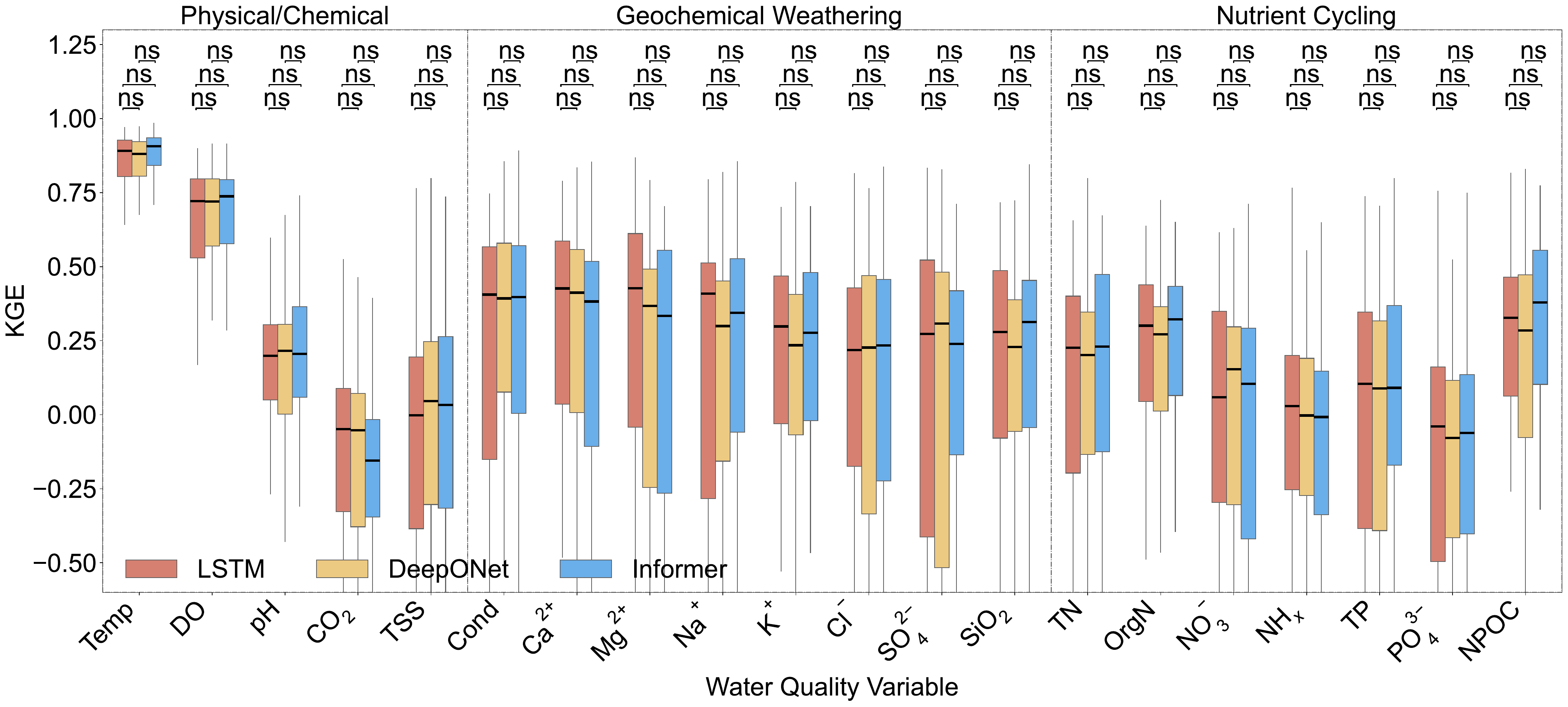}    \caption{\setlength{\baselineskip}{1.5\baselineskip} Boxplot of Kling-Gupta Efficiency (KGE) values for the testing basins \textcolor{black}{predicted by three different deep learning models} under the spatial training-testing split, evaluated across 20 predicted water quality variables associated with physical/chemical properties, geochemical weathering processes, and nutrient cycling, respectively. Each boxplot shows the median (central line), interquartile range (IQR, represented by the boxes spanning the first (Q1) to the third quartile (Q3)), and whiskers extending to $\text{Q1}-1.5\times\text{IQR}$ and $\text{Q3} + 1.5\times\text{IQR}$. The number labeled on the box indicates the median. \textcolor{black}{Wilcoxon signed-rank tests with False Discovery Rate-Benjamini-Hochberg (FDR-BH) correction indicate no significant performance differences among the three models.}}
    \label{fig:s19}
\end{figure}

\clearpage
\begin{table}[!t]
    \centering
    \vspace{-9cm}
    \caption{\setlength{\baselineskip}{1.5\baselineskip}Summary of the studied water quality variables and the average number of observations per basin, based on 482 U.S. rivers between 01/01/1982 and 12/31/2018.}
    \renewcommand{\arraystretch}{1.5}
    \makebox[\textwidth][c]{
    \begin{tabular}{|c|c|c|c|c|c|}
    \hline
    \textbf{USGS code} & \textbf{Description} & \textbf{Abbreviation} & \textbf{Unit} &  \textbf{\makecell{\# Observations\\ per basin}}&  \textcolor{black}{\textbf{\makecell{Normalization\\ method}}}\\\hline
    00010 & Water temperature & Temp & °C & 330.5 & \textcolor{black}{min-max}\\\hline
    00095 & Specific conductance & Cond & uS/cm at 25°C & 285.6 & \textcolor{black}{log-min-max}\\\hline
    00300 & Oxygen & DO & mg/L & 197.8 & \textcolor{black}{min-max} \\\hline
    00400 & pH & pH & - & 224.9 & \textcolor{black}{min-max} \\\hline
    00405 & Carbon dioxide & $\text{CO}_2$ & mg/L & 129.2 & \textcolor{black}{log-min-max} \\\hline
    00600 & Total nitrogen & TN & mg/L & 193.3 & \textcolor{black}{log-min-max} \\\hline
    00605 & Organic nitrogen & OrgN & mg/L & 171.7 & \textcolor{black}{log-min-max}\\\hline
    00618 & Nitrate & $\text{NO}_3^-$ & mg/L as N & 138.3 & \textcolor{black}{log-min-max} \\\hline
    00660 & Orthophosphate & $\text{PO}_4^{3-}$ & mg/L as $\text{PO}_4^{3-}$  & 204.9 & \textcolor{black}{log-min-max} \\\hline
    00665 & Total phosphorus & TP & mg/L as P & 266.9 & \textcolor{black}{log-min-max} \\\hline
    00681 & Organic carbon & NPOC & mg/L & 60.3 & \textcolor{black}{log-min-max} \\\hline
    00915 & Calcium & $\text{Ca}^{2+}$ & mg/L & 131.7 & \textcolor{black}{log-min-max}\\\hline
    00925 & Magnesium & $\text{Mg}^{2+}$ & mg/L & 131.8 & \textcolor{black}{log-min-max}\\\hline
    00930 & Sodium & $\text{Na}^{+}$ & mg/L & 117.3 & \textcolor{black}{log-min-max}\\\hline
    00935 & Potassium & $\text{K}^{+}$ & mg/L & 114.8 & \textcolor{black}{log-min-max}\\\hline
    00940 & Chloride & $\text{Cl}^{-}$ & mg/L & 184.1 & \textcolor{black}{log-min-max}\\\hline
    00945 & Sulfate & $\text{SO}_4^{2-}$ & mg/L & 154.3 & \textcolor{black}{log-min-max}\\\hline
    00955 & Silica & $\text{SiO}_2$ & mg/L & 116.1 & \textcolor{black}{log-min-max}\\\hline
    71846 & Ammonia and ammonium & $\text{NH}_\text{x}~(\text{NH}_\text{3}~\text{and}~\text{NH}_\text{4}^+)$ & mg/L as $\text{NH}_\text{4}^+$ & 184.1 & \textcolor{black}{log-min-max}\\\hline
    80154 & Suspended sediment concentration & TSS & mg/L & 305.4 & \textcolor{black}{log-min-max}\\\hline
    \end{tabular}}
    \label{tab:wq_statistics}
\end{table}
\clearpage

\begin{table}[!t]
    \centering
    \scriptsize
    \caption{\setlength{\baselineskip}{1.5\baselineskip}Model input features, consisting of 25 time series variables and 49 static basin attributes (sourced from the GAGES-II database).}
     \renewcommand{\arraystretch}{1.5}
     \makebox[\textwidth][c]{
    \begin{tabular}{|c|l|l|l|l|c|}
    \hline
    \textbf{Group} & \textbf{Name} & \textbf{Type} & \textbf{Description} & \textbf{Unit} & \textcolor{black}{\textbf{\makecell{Normalization\\ method}}}\\\hline
    Runoff     & runoff & time-varying & Area normalized streamflow from USGS & m/y & \textcolor{black}{log-min-max}\\\hline
    \multirow{7}{*}{\makecell{Meteorological\\ forcings}} & pr & time-varying & Daily total precipitation & mm/day & \textcolor{black}{log-min-max}\\\cline{2-6}
    & sph & time-varying & Specific humidity &unitless& \textcolor{black}{log-min-max} \\\cline{2-6}
    & srad & time-varying & Surface downwelling solar radiation &$\text{W/m}^2$& \textcolor{black}{log-min-max}\\\cline{2-6}
    & tmmn & time-varying & Daily minimum 2-meter air temperature & F& \textcolor{black}{log-min-max}\\\cline{2-6}
    & tmmx & time-varying & Daily maximum 2-meter air temperature & F& \textcolor{black}{log-min-max}\\\cline{2-6}
    & pet & time-varying & Reference grass evapotranspiration & mm/day& \textcolor{black}{log-min-max}\\\cline{2-6}
    & etr & time-varying & Reference alfalfa evapotranspiration & mm/day& \textcolor{black}{log-min-max}\\\hline
    \multirow{11}{*}{\makecell{Rainfall\\ chemistry}} & pH & time-varying & Logarithm of the H ion activity & unitless & \textcolor{black}{log-min-max} \\\cline{2-6}
    & Cond & time-varying & Electrical conductivity of water & $\mu\text{S/cm}$ & \textcolor{black}{log-min-max} \\\cline{2-6} 
    & $\text{Ca}^{2+}$ & time-varying & Ca ion concentration & mg/L & \textcolor{black}{log-min-max}\\\cline{2-6} 
    & $\text{Mg}^{2+}$ & time-varying & Mg ion concentration & mg/L & \textcolor{black}{log-min-max}\\\cline{2-6} 
    & $\text{K}^{+}$ & time-varying & K ion concentration & mg/L & \textcolor{black}{log-min-max}\\\cline{2-6} 
    & $\text{Na}^{+}$ & time-varying & Na ion concentration & mg/L & \textcolor{black}{log-min-max}\\\cline{2-6} 
    & $\text{NH}_4$ & time-varying & $\text{NH}_4$ concentration & mg/L & \textcolor{black}{log-min-max}\\\cline{2-6} 
    & $\text{NO}_3$ & time-varying & $\text{NO}_3$ concentration & mg/L & \textcolor{black}{log-min-max}\\\cline{2-6} 
    & $\text{Cl}^{-}$ & time-varying & Cl ion concentration & mg/L & \textcolor{black}{log-min-max}\\\cline{2-6}
    & $\text{SO}_4$ & time-varying & $\text{SO}_4$ concentration & mg/L & \textcolor{black}{log-min-max}\\\cline{2-6} 
    & distNTN & time-varying & The distance to the nearest NTN sampling site & km & \textcolor{black}{log-min-max}\\\hline 
    \multirow{3}{*}{\makecell{Vegetation\\ indices}} & LAI &  time-varying & Leaf area index of vegetation & $\text{m}^2/\text{m}^2$ & \textcolor{black}{min-max}\\\cline{2-6}
    & FAPAR & time-varying & Fraction of absorbed photosynthetically active radiation & unitless & \textcolor{black}{min-max}\\\cline{2-6}
    & NPP & time-varying & Net primary production & $\text{gC}/\text{m}^2/\text{day}$& \textcolor{black}{min-max}\\\hline
    \multirow{3}{*}{\makecell{Time\\ variables}} & datenum &  time-varying & The number of days relative to January 1, 2000 & unitless & \textcolor{black}{min-max}\\\cline{2-6}
    & sinT & time-varying & Sine of datenum & unitless & \textcolor{black}{min-max}\\\cline{2-6}
    & cosT & time-varying & Cosine of datenum & unitless& \textcolor{black}{min-max}\\\hline
    \multirow{3}{*}{\makecell{Basic\\ characteristics}} & HYDRO\_DISTURB\_INDX & static & \makecell[l]{Hydrologic ``disturbance index'' score, based on 7 variables:\\ 1) MAJ\_DDENS\_2009, 2) WATER\_WITHDR,\\ 3) change in dam storage 1950-2009, 4) CANALS\_PCT,\\ 5) RAW\_DIS\_NEAREST\_MAJ\_NPDES, 6) ROADS\_KM\_SQ\_KM,\\ and 7) FRAGUN\_BASIN} & unitless& \textcolor{black}{log-min-max}\\\cline{2-6}
    & BAS\_COMPACTNESS & static & \makecell[l]{Watershed compactness ratio, = area/$\text{perimeter}^2$ * 100;\\ higher number = more compact shape} & unitless& \textcolor{black}{log-min-max}\\\cline{2-6}
    & DRAIN\_SQKM & static & \makecell[l]{Watershed drainage area, sq km, as delineated in our basin boundary} & $\text{km}^2$ & \textcolor{black}{log-min-max}\\\hline
     \multirow{2}{*}{Geology} & GEOL\_REEDBUSH\_DOM & static & \makecell[l]{Dominant (highest percent of area) geology, \\derived from a simplified version of Reed \& Bush (2001) - \\Generalized Geologic Map of the Conterminous United States} & unitless& \textcolor{black}{log-min-max}\\\cline{2-6}
     & GEOL\_REEDBUSH\_DOM\_PCT & static & Percentage of the watershed covered by the dominant geology type & percentage & \textcolor{black}{log-min-max}\\\hline
     \multirow{11}{*}{\makecell{Hydrologic\\ characteristics}} & STREAMS\_K\_S\_KM & static & Stream density, km of streams per watershed sq km, from NHD 100k streams & $\text{km}/\text{km}^2$& \textcolor{black}{log-min-max}\\\cline{2-6}
     & STRAHLER\_MAX & static & Maximum Strahler stream order in the watershed, from NHDPlus & unitless& \textcolor{black}{log-min-max}\\\cline{2-6}
     & MAINSTEM\_SINUOUSITY & static & \makecell[l]{Sinuosity of mainstem stream line, from our delineation of mainstem\\ stream lines. Defined as curvilinear length of the mainstem stream line\\ dividedby the straight-line distance between the end points of the line.} & unitless& \textcolor{black}{log-min-max}\\\cline{2-6}
     & BFI\_AVE & static & \makecell[l]{Base Flow Index (BFI). The BFI is a ratio of base flow to total streamflow,\\ expressed as a percentage and ranging from 0 to 100.\\ Base flow is the sustained, slowly varying component of streamflow,\\ usually attributed to ground-water discharge to a stream.} & percentage& \textcolor{black}{log-min-max}\\\cline{2-6}
     & CONTACT & static & Subsurface flow contact time index & days& \textcolor{black}{log-min-max}\\\cline{2-6}
     & PCT\_1ST\_ORDER & static & \makecell[l]{Percent of stream lengths in the watershed which are first-order\\ streams (Strahler order); from NHDPlus \& percentage} & percentage & \textcolor{black}{log-min-max}\\\cline{2-6}
     & PCT\_2ND\_ORDER & static &  \makecell[l]{Percent of stream lengths in the watershed which are second-order\\ streams (Strahler order); from NHDPlus \& percentage} & percentage & \textcolor{black}{log-min-max}\\\cline{2-6}
     & PCT\_3RD\_ORDER & static &  \makecell[l]{Percent of stream lengths in the watershed which are third-order\\ streams (Strahler order); from NHDPlus \& percentage} & percentage & \textcolor{black}{log-min-max}\\\cline{2-6}
     & PCT\_4TH\_ORDER & static &  \makecell[l]{Percent of stream lengths in the watershed which are fourth-order\\ streams (Strahler order); from NHDPlus \& percentage} & percentage & \textcolor{black}{log-min-max}\\\cline{2-6}
     & PCT\_5TH\_ORDER & static &  \makecell[l]{Percent of stream lengths in the watershed which are fifth-order\\ streams (Strahler order); from NHDPlus \& percentage} & percentage & \textcolor{black}{log-min-max}\\\cline{2-6}
     & PCT\_6TH\_ORDER\_OR\_MORE & static &  \makecell[l]{Percent of stream lengths in the watershed which are sixth or greater-order\\ streams (Strahler order); from NHDPlus \& percentage} & percentage & \textcolor{black}{log-min-max}\\\hline
   
    \end{tabular}}
    \label{tab:inputs1}
\end{table}

\begin{table}[!t]
    \centering
    \scriptsize
    \renewcommand{\arraystretch}{1.5}
    \makebox[\textwidth][c]{
    \begin{tabular}{|c|l|l|l|l|c|}
    \hline
    \textbf{Group} & \textbf{Name} & \textbf{Type} & \textbf{Description} & \textbf{Unit} & \textcolor{black}{\textbf{\makecell{Normalization\\ method}}}\\\hline
    \multirow{2}{*}{\makecell{Historical and\\
    current dams\\ information}} & DDENS\_2009 & static & Dam density; number per 100 km sq & \makecell[l]{number of\\ dams/100 $\text{km}^2$} & \textcolor{black}{log-min-max}\\\cline{2-6}
    & STOR\_NOR\_2009 & static & \makecell[l]{Dam storage in watershed (``NORMAL\_STORAGE'');\\ megaliters total storage per sq km\\  (1 megalitres = 1,000,000 liters = 1,000 cubic meters)}& megaliters/$\text{km}^2$ & \textcolor{black}{log-min-max}\\\hline
    NPDES & NPDES\_MAJ\_DENS & static & \makecell[l]{Density of NPDES (National Pollutant Discharge Elimination System)\\``major'' point locations in the watershed; number per 100 km sq.  \\Major locations are defined by an EPA-assigned major flag.\\ From the download of NPDES national database summer 2006.} & \makecell[l]{number of \\sites/100$\text{km}^2$} & \textcolor{black}{log-min-max}\\\hline
     \multirow{6}{*}{\makecell{Percentages of\\
    land cover 2006 \\in the watershed\\ and lanscape}} & DEVNLCD06 & static & \makecell[l]{Watershed percent ``developed'' (urban), 2006 era (2001 for AK-HI-PR).\\  Sum of classes 21, 22, 23, and 24.} & percentage & \textcolor{black}{log-min-max}\\\cline{2-6}
     & FORESTNLCD06 & static &  \makecell[l]{Watershed percent ``forest'', 2006 era (2001 for AK-HI-PR).\\ Sum of classes 41, 42, and 43.} & percentage & \textcolor{black}{log-min-max}\\\cline{2-6}
     & PLANTNLCD06 & static & \makecell[l]{Watershed percent ``planted/cultivated'' (agriculture),\\ 2006 era (2001 for AK-HI-PR). Sum of classes 81 and 82.} & percentage & \textcolor{black}{log-min-max} \\\cline{2-6}
     & WATERNLCD06 & static & \makecell[l]{Watershed percent Open Water (class 11)} & percentage & \textcolor{black}{log-min-max}\\\cline{2-6}
     & WOODYWETNLCD06 & static & \makecell[l]{Watershed percent Woody Wetlands (class 90)} & percentage & \textcolor{black}{log-min-max}\\\cline{2-6}
     & EMERGWETNLCD06 & static & \makecell[l]{Watershed percent Emergent Herbaceous Wetlands (class 95)} & percentage & \textcolor{black}{log-min-max} \\\hline
    \multirow{2}{*}{\makecell{Nitrogen and phosphorus\\ application rate\\ in the watershed}} & NITR\_APP\_KG\_SQKM & static & \makecell[l]{Estimate of nitrogen from fertilizer and manure, from Census\\ of Ag 1997, based on county-wide sales and percent\\ agricultural land cover in the watershed.} & kg/$\text{km}^2$ & \textcolor{black}{log-min-max}\\\cline{2-6}
     & PHOS\_APP\_KG\_SQKM & static & \makecell[l]{Estimate of nitrogen from fertilizer and manure, from Census\\ of Ag 1997, based on county-wide sales and percent\\ agricultural land cover in the watershed.} & kg/$\text{km}^2$ & \textcolor{black}{log-min-max}\\\hline
     Pesticide & PESTAPP\_KG\_SQKM & static & \makecell[l]{Estimate of agricultural pesticide application (219 types),\\ kg/sq km, from Census of Ag 1997, based on county-wide sales\\ and percent agricultural land cover in the watershed} & kg/$\text{km}^2$ & \textcolor{black}{log-min-max}\\\hline
     \multirow{5}{*}{\makecell{Regions}} & ECO2\_BAS\_DOM & static & \makecell[l]{Dominant (highest \% of the area) Level II ecoregion within the watershed.\\ See X\_Region\_Names sheet for crosswalk to name.} & unitless & \textcolor{black}{log-min-max}\\\cline{2-6}
     & ECO3\_BAS\_DOM & static & \makecell[l]{Dominant (highest \% of the area) Level III ecoregion within the watershed.\\ See X\_Region\_Names sheet for crosswalk to name.} & \makecell[l]{Level III\\ ecoregion (1-84)} & \textcolor{black}{log-min-max}\\\cline{2-6}
     & NUTR\_BAS\_DOM & static & \makecell[l]{Dominant (highest \% of the area) nutrient ecoregion\\ within the watershed. See X\_Region\_Names sheet for crosswalk to name.} & \makecell[l]{Nutrient\\ ecoregion (1-14)} & \textcolor{black}{log-min-max} \\\cline{2-6}
     & HLR\_BAS\_DOM\_100M & static & \makecell[l]{Dominant (highest \% of the area) Hydrologic Landscape Region\\ within the watershed. See X\_Region\_Names sheet for crosswalk to name.} & \makecell[l]{HLR region (1-20)} & \textcolor{black}{log-min-max}\\\cline{2-6}
     & PNV\_BAS\_DOM & static & \makecell[l]{Dominant (highest \% of the area) Potential Natural Vegetation (PNV)\\ within the watershed. See X\_Region\_Names sheet for crosswalk to name.} & \makecell[l]{PNV type (1-63)} & \textcolor{black}{log-min-max}\\\hline
     \multirow{10}{*}{\makecell{Soil}} & AWCAVE & static & \makecell[l]{Average value for the range of available water capacity for\\ the soil layer or horizon (inches of water per inch of soil depth)} & unitless & \textcolor{black}{log-min-max}\\\cline{2-6}
     & PERMAVE & static & \makecell[l]{Average permeability (inches/hour)} & inches/hour & \textcolor{black}{log-min-max} \\\cline{2-6}
     & BDAVE & static & \makecell[l]{Average value of bulk density (grams per cubic centimeter)} & \makecell[l]{grams per\\ cubic centimeter} & \textcolor{black}{log-min-max}\\\cline{2-6}
     & OMAVE & static & \makecell[l]{Average value of organic matter content (percent by weight)} & percentage & \textcolor{black}{log-min-max}\\\cline{2-6}
     & WTDEPAVE & static & \makecell[l]{Average value of depth to seasonally high water table (feet)} & feet & \textcolor{black}{log-min-max}\\\cline{2-6}
     & ROCKDEPAVE & static & \makecell[l]{Average value of total soil thickness examined (inches)} & \makecell[l]{inches} & \textcolor{black}{log-min-max}\\\cline{2-6}
     & CLAYAVE & static & \makecell[l]{Average value of clay content (percentage)} & percentage & \textcolor{black}{log-min-max}\\\cline{2-6}
      & SILTAVE & static & \makecell[l]{Average value of silt content (percentage)} & percentage & \textcolor{black}{log-min-max}\\\cline{2-6}
     & KFACT\_UP & static & \makecell[l]{Average K-factor value for the uppermost soil horizon\\ in each soil component. K-factor is an erodibility factor\\ which quantifies the susceptibility of soil particles to\\ detachment and movement by water. The K-factor is used in\\ the Universal Soil Loss Equation (USLE) to estimate soil loss by water.\\ Higher values of the K-factor indicate greater potential for erosion} & \makecell[l]{unitless} & \textcolor{black}{log-min-max}\\\cline{2-6}
     & RFACT  & static & \makecell[l]{Rainfall and Runoff factor (``R factor'' of Universal Soil Loss Equation);\\ average annual value for the period 1971-2000.} & \makecell[l]{100s ft-tonf\\ in/h/ac/yr} & \textcolor{black}{log-min-max}\\\hline
     \multirow{3}{*}{\makecell{Topographic\\
    characteristics}} & ELEV\_MEAN\_M\_BASIN & static & \makecell[l]{Mean watershed elevation (meters) from 100m National Elevation Dataset} & m & \textcolor{black}{log-min-max}\\\cline{2-6}
    & SLOPE\_PCT & static & \makecell[l]{Mean watershed slope, percent.\\ Derived from 100m resolution National Elevation Dataset,\\ so slope values may differ from those calculated from data\\ of other resolutions.} & percentage & \textcolor{black}{log-min-max}\\\cline{2-6}
     & ASPECT\_DEGREES & static & \makecell[l]{Mean watershed aspect, degrees (degrees of the compass, 0-360).\\ Derived from 100m resolution National Elevation Data.\\  0 and 360 point to north, because of the national Albers projection\\ actual aspect may vary.} & degrees (0-360)& \textcolor{black}{log-min-max}\\\hline
     \multirow{2}{*}{\makecell{Latitude\\ and Longitude}} & LAT\_GAGE & static & \makecell[l]{Latitude at gage, decimal degrees} & \makecell[l]{decimal degrees,\\ datum NAD83} & \textcolor{black}{min-max}\\\cline{2-6}
     & LNG\_GAGE & static & \makecell[l]{Longitude at gage, decimal degrees} & \makecell[l]{decimal degrees,\\ datum NAD83} & \textcolor{black}{min-max}\\\hline
     Snow & SNOW\_PCT\_PRECIP & static & \makecell[l]{Snow percent of total precipitation estimate,\\ mean for period 1901-2000.  From McCabe and Wolock\\ (submitted, 2008), 1km grid.} & percentage & \textcolor{black}{log-min-max}\\\hline
    \end{tabular}}
    \label{tab:inputs2}
\end{table}

\end{document}